\pdfoutput=1
\newif\ifaclstyle
\aclstylefalse
\newif\ifanonymous
\anonymousfalse
\ifaclstyle
  \documentclass[11pt]{article}
  \usepackage[review]{acl}
  \usepackage{times}
  \anonymoustrue
\else
  \documentclass[10pt]{article}
  \usepackage[letterpaper,top=0.70in,bottom=0.78in,left=0.68in,right=0.68in]{geometry}
\fi
\usepackage[T1]{fontenc}
\usepackage[utf8]{inputenc}
\ifaclstyle\else\usepackage{lmodern}\fi
\usepackage{microtype}
\usepackage{amsmath,amssymb,mathtools}
\usepackage{booktabs,longtable}
\usepackage{balance}
\usepackage{afterpage}
\usepackage{etoolbox}
\makeatletter
\patchcmd{\@BAlancecol}{\setbox\@leftcolumn}
  {\advance\@tempdima by 2pt\setbox\@leftcolumn}{}{}
\makeatother
\usepackage{multirow}
\usepackage{array}
\newcolumntype{P}[1]{>{\RaggedRight\arraybackslash}p{#1}}
\newcolumntype{Y}{>{\RaggedRight\arraybackslash}X}
\usepackage{graphicx}
\usepackage{xcolor}
\usepackage{url}
\usepackage{hyperref}
\usepackage{enumitem}
\usepackage{caption}
\usepackage{subcaption}
\usepackage{siunitx}
\usepackage{natbib}
\usepackage{placeins}
\usepackage{float}
\usepackage{needspace}
\usepackage{ragged2e,multicol}
\ifaclstyle\else
\usepackage{titlesec}
\titleformat{\section}{\normalfont\fontsize{12}{14}\selectfont\bfseries\raggedright}{\thesection}{.65em}{}
\titleformat{\subsection}{\normalfont\fontsize{10}{12}\selectfont\bfseries\raggedright}{\thesubsection}{.65em}{}
\titlespacing*{\section}{0pt}{9pt plus 2pt minus 2pt}{4pt plus 1pt}
\titlespacing*{\subsection}{0pt}{7pt plus 2pt minus 1pt}{3pt plus 1pt}
\titlespacing*{\paragraph}{0pt}{5pt plus 1pt minus 1pt}{.65em}
\fi

\usepackage{tikz}
\usetikzlibrary{arrows.meta,positioning,fit,calc}
\usepackage{pgfplots}
\usepgfplotslibrary{groupplots}
\pgfplotsset{compat=1.14,
  every axis/.append style={font=\rmfamily\fontsize{8}{9.5}\selectfont},
  every axis/.append style={
    /pgf/number format/1000 sep={},
    every node near coord/.append style={/pgf/number format/fixed,
      /pgf/number format/precision=2},
    legend cell align=left,
    legend image post style={line width=.65pt}
  }}
\usepackage{tabularx}
\IfFileExists{xurl.sty}{\usepackage{xurl}}{%
  \appto\UrlBreaks{\do\a\do\b\do\c\do\d\do\e\do\f\do\g\do\h\do\i\do\j\do\k\do\l\do\m\do\n\do\o\do\p\do\q\do\r\do\s\do\t\do\u\do\v\do\w\do\x\do\y\do\z\do\0\do\1\do\2\do\3\do\4\do\5\do\6\do\7\do\8\do\9}
}

\hypersetup{
  pdftitle={AROMA: Adaptive Representation-Oriented Multimodal Actuation for Vision--Language Model Control},
  pdfsubject={Sensitivity geometry, finite reachability, and target selection for vision-language model control},
  bookmarksnumbered=true,
  pdfdisplaydoctitle=true,
  colorlinks=true,
  linkcolor=aromaInk,
  citecolor=aromaInk,
  urlcolor=aromaInk
}
\ifanonymous\else\hypersetup{pdfauthor={Zhaoning Zhang}}\fi

\ifaclstyle\else
\fi
\ifaclstyle\else
\setlist[itemize]{leftmargin=*,topsep=2pt,itemsep=1pt,parsep=0pt}
\setlist[enumerate]{leftmargin=*,topsep=2pt,itemsep=1pt,parsep=0pt}
\fi
\ifaclstyle
  \DeclareCaptionFont{aromacaption}{\fontsize{10}{12}\selectfont}
\else
  \DeclareCaptionFont{aromacaption}{\fontsize{9}{11}\selectfont}
\fi
\renewcommand{\arraystretch}{1.16}
\ifdefined\XeTeXversion
  \microtypesetup{protrusion=true,expansion=false}
\else
  \microtypesetup{protrusion=true,expansion=true}
\fi
\ifaclstyle\else
\fi
\ifaclstyle\else
\fi

\newcommand{\method}{\textsc{AROMA}}

\newcommand{\mncsa}{\textsc{MN-CSA}}

\newcommand{\ugr}{\textsc{UGR}}
\newcommand{\mea}{\textsc{MEA}}
\newcommand{\headname}{L18H13}
\newcommand{\pp}{\,pp}
\newcommand{\noop}{\textsc{NoOp}}
\newcommand{\tally}{\textsc{TallyQA}}

\newcommand{\proc}{\textsc{Proc-Count-Causal}}
\newcommand{\modelname}{\textsc{Llama-3.2-11B-Vision-Instruct}}
\newcommand{\idefics}{\textsc{IDEFICS-9B-\allowbreak Instruct}}
\newcommand{\R}{\mathbb{R}}
\newcommand{\ind}{\mathbb{I}}

\ifaclstyle
\title{AROMA: Adaptive Representation-Oriented Multimodal Actuation\\
for Vision--Language Model Control}
\else
\title{
\fontsize{15}{17.5}\selectfont
\textbf{
AROMA: Adaptive Representation-Oriented Multimodal Actuation\\
for Vision--Language Model Control
}
}
\fi

\ifanonymous
  \author{Anonymous Author(s)}
\else
  \author{Zhaoning Zhang\\
  University of Toronto\\
  \texttt{zhaoning.zhang@mail.utoronto.ca}}
\fi

\date{}


\ifaclstyle\else
\makeatletter
\patchcmd{\@maketitle}{\vskip 2em}{\vskip .6em}{}{}
\patchcmd{\@maketitle}{\vskip 1.5em}{\vskip .8em}{}{}
\patchcmd{\@maketitle}{\vskip 1em}{\vskip .4em}{}{}
\patchcmd{\@maketitle}{\vskip 1.5em}{\vskip .6em}{}{}
\makeatother
\fi

\definecolor{aromaInk}{HTML}{263340}
\definecolor{aromaRule}{HTML}{9BA5AE}
\definecolor{aromaMechanism}{HTML}{F0EDF6}
\definecolor{aromaGeometry}{HTML}{EAF2F7}
\definecolor{aromaPolicy}{HTML}{EEF4ED}
\definecolor{aromaDecision}{HTML}{F8F2E9}
\definecolor{aromaPlain}{HTML}{F7F8F9}

\definecolor{aromaBlue}{HTML}{356C9B}
\definecolor{aromaGreen}{HTML}{39836B}
\definecolor{aromaOrange}{HTML}{C98239}
\definecolor{aromaPurple}{HTML}{806397}
\definecolor{aromaGray}{HTML}{A0AAB4}
\definecolor{aromaGrayMid}{HTML}{89939D}
\definecolor{aromaGrayDark}{HTML}{808B96}
\definecolor{aromaGrayLight}{HTML}{BFC6CD}
\definecolor{aromaGrid}{HTML}{E3E7EB}
\pgfplotsset{every axis/.append style={
  scaled ticks=false,
  tick label style={font=\fontsize{7.5}{9}\selectfont},
  label style={font=\fontsize{8}{9.5}\selectfont},
  legend style={font=\fontsize{7.5}{9}\selectfont},
  legend cell align=left,
  every node near coord/.append style={/pgf/number format/fixed,
    /pgf/number format/precision=2},
}}

\ifaclstyle
  \newcommand{\aromaPairPad}{1.70cm}\newcommand{\aromaTripleW}{.240\textwidth}\newcommand{\aromaGroupSep}{1.10cm}
  
  \newcommand{\aromaHa}{2.25cm}\newcommand{\aromaHb}{2.40cm}\newcommand{\aromaHc}{2.60cm}
  \newcommand{\aromaFigY}{0.84cm}\newcommand{\aromaBoxH}{1.12cm}\newcommand{\aromaLaneH}{.84cm}\newcommand{\aromaSlimH}{.52cm}

\else
  \newcommand{\aromaPairPad}{1.55cm}\newcommand{\aromaTripleW}{.255\textwidth}\newcommand{\aromaGroupSep}{1.35cm}
  
  \newcommand{\aromaHa}{3.0cm}\newcommand{\aromaHb}{3.25cm}\newcommand{\aromaHc}{3.5cm}
  \newcommand{\aromaFigY}{1cm}\newcommand{\aromaBoxH}{1.30cm}\newcommand{\aromaLaneH}{.94cm}\newcommand{\aromaSlimH}{.58cm}
\fi

\begin{document}
\twocolumn
\maketitle
\begin{abstract}
Mechanistic interpretability can localize where a vision--language model (VLM) computes a behavior, but localization does not by itself yield control. We introduce \method{} (\textbf{A}daptive \textbf{R}epresentation-\textbf{O}riented \textbf{M}ultimodal \textbf{A}ctuation), a framework that turns a causally validated component of a frozen VLM into a controller, and use visual counting to identify and close three gaps between localization and useful control. \textbf{Sensitivity is not energy:} in a cross-attention head of \modelname{}, four of 128 dimensions carry 89.8\% of cardinality sensitivity but 9.5\% of activation energy; a closed-form equal-norm calibration corrects the resulting under-actuation without gain search ($+8.0\pp$ on newly generated data) and outperforms mean-difference and PCA directions at matched displacement. \textbf{Local is not finite:} first-order predictions are near-exact for small edits but not at scale; Minimal-Effort Actuation (\mea{}), which re-linearizes the target margin at every step, reaches 81\% of ground-truth targets versus 15\% for a fixed direction, makes the model generate the correct count on 80 of 112 errors, and replicates in a second architecture. \textbf{Reachable is not useful:} once targets are reachable, the target decides deployment utility. A ground-truth-free runner-up target under a selective gate improves the reference policy by $4.3\pp$ on the 630-question Stage-2 branch of a prespecified cohort of 7{,}992 new natural questions while editing 0.8\% of inputs, and object-existence hallucination, the error the gate isolates, is already readable from the model's own numeral distribution.
\end{abstract}
\section{Introduction}
\label{sec:intro}

A child who can count to ten has already mastered a skill that today's vision--language models (VLMs) still find difficult. Even frontier models behave as ``4-knowers'' that enumerate reliably only within the subitizing range \citep{testolin2025enumeration}, and real-world counting remains hard for modern multimodal systems \citep{tamarapalli2025countqa,qharabagh2026lvlmcount,rong2026unicbench}. Counting requires individuating objects, keeping track of partial visual evidence, and aggregating exactly, which makes it a clean probe of whether a VLM's answer is grounded in its internal visual computation.

Mechanistic work shows that the relevant computation is often present before the answer is produced. Counting can be traced to identifiable attention heads \citep{che2026counting}, count information can be followed through model internals with causal mediation \citep{hasani2026countscope}, and the correct count can remain decodable even when the verbalized answer is wrong \citep{elshangiti2026count}. The next question is how to turn such localization into reliable control.

Causal localization identifies components implicated in a behavior, and activation steering specifies an intervention through a direction, a magnitude, or a condition for applying it \citep{meng2022rome,turner2023acteng,zou2023repe,lee2025cast}. Turning a localized component into a controller requires three further properties: the component must move at finite scale, it must be steerable to a chosen output, and the chosen output must be the right one. Each property is a separate empirical question, and \method{} answers each with a dedicated held-out experiment.

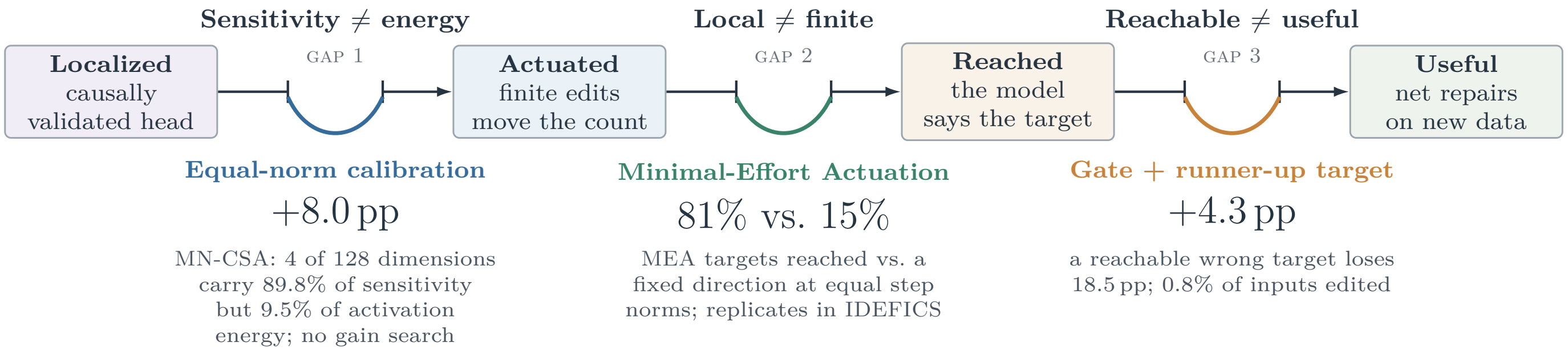
\begin{figure*}[!t]
\centering
\begin{tikzpicture}[
  state/.style={draw=aromaRule,rounded corners=2pt,line width=.55pt,align=center,inner sep=3pt,
    minimum height=1.0cm,text width=2.25cm,execute at begin node={\hyphenpenalty=10000\exhyphenpenalty=10000},
    font=\rmfamily\fontsize{8}{9.4}\selectfont,text=aromaInk},
  track/.style={draw=aromaInk,line width=.8pt},
  edge/.style={draw=aromaInk,line width=.8pt},
  bridge/.style={line width=1.3pt},
  gaplab/.style={font=\rmfamily\fontsize{7}{8.5}\selectfont,text=aromaInk!70,align=center},
  claim/.style={font=\rmfamily\bfseries\fontsize{8.5}{10}\selectfont,text=aromaInk,align=center},
  rname/.style={font=\rmfamily\bfseries\fontsize{8}{9.5}\selectfont,align=center},
  big/.style={font=\rmfamily\fontsize{13}{15}\selectfont,text=aromaInk,align=center},
  small/.style={font=\rmfamily\fontsize{7}{8.4}\selectfont,text=aromaInk!85,align=center,text width=4.1cm,
    execute at begin node={\hyphenpenalty=10000\exhyphenpenalty=10000}}
]
\node[state,fill=aromaMechanism] (s1) at (0,0)    {\textbf{Localized}\\causally validated head};
\node[state,fill=aromaGeometry]  (s2) at (5.2,0)  {\textbf{Actuated}\\finite edits move the count};
\node[state,fill=aromaDecision]  (s3) at (10.4,0) {\textbf{Reached}\\the model says the target};
\node[state,fill=aromaPolicy]    (s4) at (15.6,0) {\textbf{Useful}\\net repairs on new data};
\foreach \a/\b/\m/\c/\n/\cl/\rn/\bg/\sm in {
  s1/s2/2.6/aromaBlue/1/{Sensitivity $\neq$ energy}/{Equal-norm calibration}/{$+8.0$\,pp}/{\mncsa{}: 4 of 128 dimensions carry 89.8\% of sensitivity but 9.5\% of activation energy; no gain search},
  s2/s3/7.8/aromaGreen/2/{Local $\neq$ finite}/{Minimal-Effort Actuation}/{81\% vs.\ 15\%}/{\mea{} targets reached vs.\ a fixed direction at equal step norms; replicates in \textsc{IDEFICS}},
  s3/s4/13.0/aromaOrange/3/{Reachable $\neq$ useful}/{Gate + runner-up target}/{$+4.3$\,pp}/{a reachable wrong target loses 18.5\,pp; 0.8\% of inputs edited}}{
  \draw[track] (\a.east) -- (\m-0.55,0);
  \draw[track,-{Latex[length=1.8mm,width=1.2mm]}] (\m+0.55,0) -- (\b.west);
  \draw[edge] (\m-0.55,-0.13)--(\m-0.55,0.13);
  \draw[edge] (\m+0.55,-0.13)--(\m+0.55,0.13);
  \node[gaplab] at (\m,0.42) {\textsc{gap \n}};
  \node[claim] at (\m,0.82) {\cl};
  \draw[bridge,draw=\c] (\m-0.55,-0.06) .. controls (\m-0.3,-0.62) and (\m+0.3,-0.62) .. (\m+0.55,-0.06);
  \node[rname,text=\c] at (\m,-0.92) {\rn};
  \node[big] at (\m,-1.42) {\bg};
  \node[small,anchor=north] at (\m,-1.72) {\sm};
}
\end{tikzpicture}
\caption{\textbf{Localized is not controllable.} A causally localized component becomes a useful controller only after three gaps are closed. Each gap is a separate empirical question with its own held-out test, and \method{} supplies one remedy per gap (colored arcs): equal-norm calibration of the sensitivity subspace (Gap~1, \S\ref{sec:res_mncsa}), Minimal-Effort Actuation (Gap~2, \S\ref{sec:res_mea}), and a learned gate with a ground-truth-free runner-up target (Gap~3, \S\ref{sec:deployable_mea}). Numbers are from \modelname{} on held-out or prospectively specified cohorts unless noted.}
\label{fig:three_gaps}
\end{figure*}

\method{} organizes this program around one thesis, \emph{localized is not controllable}, and three gaps (Fig.~\ref{fig:three_gaps}). \textbf{Gap~1, sensitivity is not energy:} the directions that move the count carry little of the head's activation, so naive low-rank edits under-actuate; an equal-norm calibration closes the gap. \textbf{Gap~2, local is not finite:} a first-order response law is exact for small edits and drifts at finite scale; Minimal-Effort Actuation (\mea{}) closes the gap by re-linearizing the target margin at every step. \textbf{Gap~3, reachable is not useful:} once the actuator can reach a target, utility is decided by which target is pursued and when; a ground-truth-free runner-up target and a learned gate close the gap. The same stack separates five decisions, \emph{where}, \emph{which geometry}, \emph{how}, \emph{which target}, and \emph{when} (Fig.~\ref{fig:aroma_stack}). Because counting has a discrete outcome space, every edit is scored as a repair, a break, or no change, with abstention tracked separately, and every deployment claim is confirmed on data generated after the method, comparators, and endpoints were fixed. The full stack is developed on one model and actuator; actuator behavior and its low-rank geometry replicate across candidate heads and in a second architecture, where \mea{} also replicates on newly generated data.

\noindent Our contributions are:
\begin{itemize}
  \item \textbf{A factorized mechanism-to-control framework.} \method{} separates \emph{where}, \emph{which geometry}, \emph{how}, \emph{which target}, and \emph{when} into stages that are each confirmed on held-out data, in place of a single steering heuristic (\S\ref{sec:method}).
  \item \textbf{Gap 1, sensitivity is not energy: low-rank causal actuation.} Inside a validated actuator, four of 128 head dimensions carry 89.8\% of cardinality sensitivity and 9.5\% of activation energy. This sensitivity--energy mismatch predicts how far naive projection under-actuates (median $\rho=0.31$, a $3\times$ displacement deficit). Magnitude-normalized causal subspace actuation (\mncsa{}) corrects it with a closed-form equal-norm calibration that needs no gain search and gives $+8.00\pp$ on a cohort generated after the method was fixed. On that cohort and at equal displacement, the sensitivity subspace beats a mean-difference count direction by $2.80\pp$, activation PCA by $5.50\pp$, and Bottom-4 and random rank-4 controls. Bidirectional actuator behavior replicates in 6/7 candidate heads, and the same low-rank structure appears independently in \idefics{}. Under iterative control, the four dimensions lose no reachability relative to the full 128-dimensional head (91 versus 92 of 112) (\S\ref{sec:res_mncsa}, \S\ref{sec:res_mea}).
  \item \textbf{Gap 2, local is not finite: a local control law and an iterative target controller.} First-order predictions are near-exact for small edits and drift at finite scale. \mea{} re-linearizes the target margin after every step and reaches 81.25\% of oracle targets, more than five times the 15.18\% of a fixed direction with identical step norms and far above an activation-aligned control. The effect appears in behavior: given the ground-truth target, the model's own greedy answer is correct on 80/112 examples with the \mea{} edit and on 0/112 without it. The pattern replicates in \idefics{} on 500 newly generated images: \mea{} reaches 53.2\% of oracle targets versus 2.8\% for a fixed direction, and raises correct generated answers from 124 to 218 of 470 (97:3, exact $p=2.6\times10^{-25}$) (\S\ref{sec:res_mea}).
  \item \textbf{Gap 3, reachable is not useful: target specification decides utility.} With a validated actuator, the target decides deployment utility. A reachable but wrong target rule loses $18.54\pp$, whereas a ground-truth-free runner-up target with a learned gate improves the reference \ugr{} policy by $+4.29\pp$ on a prospectively specified Stage-2 population (exact McNemar $p=4.6\times10^{-7}$) while editing 0.8\% of inputs. \mea{} realizes 97\% of assigned targets and makes the model itself generate the selected answer (54/63 versus 7/63), establishing that the answer is causally realizable through the validated actuator rather than imposed on the output. The target signal is already inside the model: the gate isolates zero-count overcounting, a form of object-existence hallucination, and a threshold on $p(0\mid x)$ reads it directly from the model's numeral distribution (\S\ref{sec:deployable_mea}).
\end{itemize}

\section{Related Work}
\label{sec:related}

\paragraph{Counting in VLMs.}
Counting has long exposed the limits of generic attention in VQA \citep{zhang2018counting,trott2018counting,acharya2019tallyqa}, and large multimodal models remain unreliable counters beyond the subitizing range and in cluttered scenes \citep{testolin2025enumeration,tamarapalli2025countqa,qharabagh2026lvlmcount,rong2026unicbench,deng2026holocount}. Mechanistically, \citet{che2026counting} trace counting to functionally distinct attention heads and regularize them during fine-tuning; \citet{hasani2026countscope} follow numerical information through language and vision--language models with causal mediation and activation patching; \citet{ledinhanh2026counting} find that count evidence is strongest at the modality projection and under-used in later language layers; and \citet{elshangiti2026count} show that the correct count is often decodable but misaligned with the verbalized answer, and correct errors through probe-direction steering and re-prompting. Attention-based interventions reweight count-relevant visual tokens at inference time \citep{sengupta2025count,alzahrani2026countercount}. \method{} keeps the VLM frozen and acts on a causally validated actuator rather than on attention maps, probe directions, or a dedicated counter \citep{qian2025t2icount}.

\paragraph{Mechanistic intervention and activation steering.}
Causal mediation analysis and causal tracing made intervention the standard of evidence for mechanistic claims \citep{pearl2009causality,vig2020causal,meng2022rome}, and recent work localizes multimodal components that causally influence grounding, hallucination, and task behavior \citep{golovanevsky2025notice,wang2025vseam,rudman2026prompt,zhang2026vib,wang2026hones,rahmanzadehgervi2025tab}. Activation engineering steers hidden states at inference time along contrastive directions \citep{turner2023acteng,rimsky2024caa,zou2023repe} or along probe directions at selected attention heads \citep{li2023iti}, with studies of intervention magnitude \citep{stoehr2024scaling}, conditional deployment \citep{lee2025cast}, and VLM-specific steering \citep{sivakumar2025steervlm,yin2026dynamic,yang2026crvlm}. Supervised alternatives learn a low-rank subspace in which to intervene \citep{geiger2024das,wu2024reft}. Our subspace is read off the cardinality gradient inside an actuator that has already passed bidirectional intervention tests, and we compare it with contrastive and principal-component directions at equal displacement (\S\ref{sec:res_mncsa}). Subspace interpretations require held-out intervention \citep{makelov2024subspace}, and representation steering often trails prompting in end-task utility \citep{wu2025axbench}; both observations motivate separating mechanistic reachability from deployment utility. \mea{}'s per-step update is the minimum-norm linearized boundary step of DeepFool \citep{moosavi2016deepfool}, used here as an internal target controller inside a causally validated subspace rather than as an input-space perturbation. GrAInS \citep{nguyen2025grains} constructs fixed steering directions from gradient attribution, token ablation, and representation differences; two task-adapted implementations serve as our external baselines (Appendix~\ref{app:external_controls}). \method{} tests each boundary between validated actuator, sensitivity geometry, finite target execution, target choice, and gating explicitly (Appendix~\ref{app:closest_work}).

\section{AROMA}
\label{sec:method}

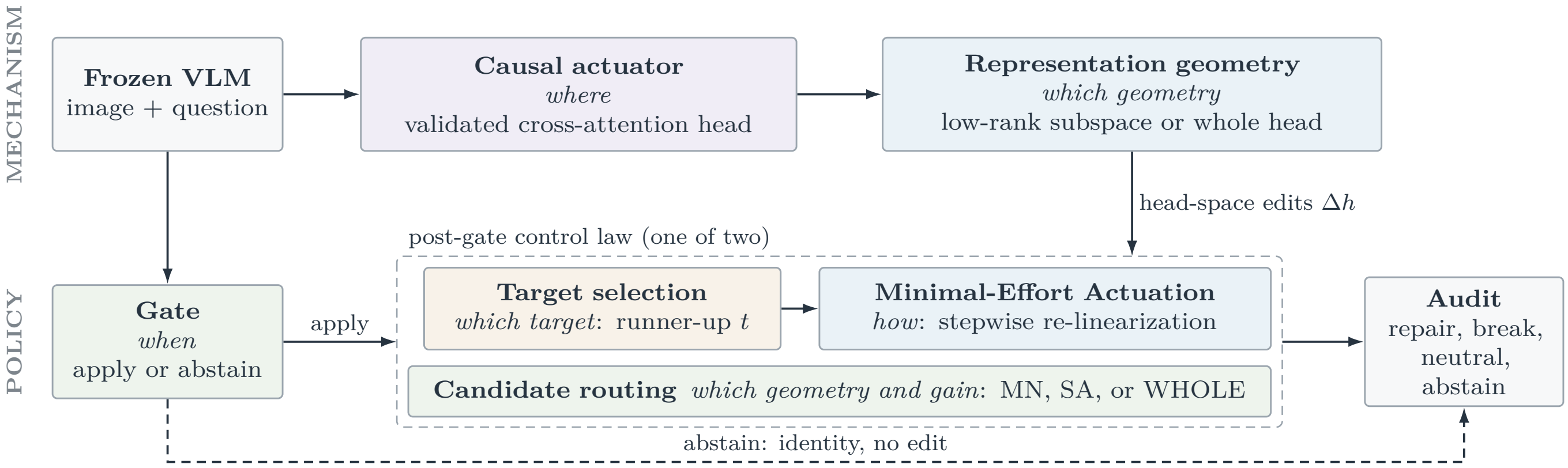
\begin{figure*}[!t]
\centering
\pgfmathsetlengthmacro{\aromaU}{\textwidth/100}%
\begin{tikzpicture}[
  x=\aromaU,y=\aromaFigY,
  box/.style={draw=aromaRule,rounded corners=1.5pt,line width=.55pt,
    align=center,inner sep=4pt,minimum height=\aromaBoxH,
    font=\rmfamily\fontsize{8}{9.5}\selectfont,text=aromaInk},
  lane/.style={box,minimum height=\aromaLaneH},
  slim/.style={box,minimum height=\aromaSlimH},
  arrow/.style={-{Latex[length=1.8mm,width=1.2mm]},draw=aromaInk,line width=.7pt},
  lab/.style={font=\rmfamily\fontsize{7.5}{9}\selectfont,text=aromaInk,inner sep=2pt},
  band/.style={font=\rmfamily\bfseries\fontsize{7}{8.5}\selectfont,text=aromaInk!60,
    rotate=90,anchor=center,inner sep=0pt}
]
\node[band] at (.8,0) {MECHANISM};
\node[box,fill=aromaPlain,text width=13*\aromaU] (vlm) at (10.5,0)
  {\textbf{Frozen VLM}\\image + question};
\node[box,fill=aromaMechanism,text width=26*\aromaU] (act) at (36.5,0)
  {\textbf{Causal actuator}\\\emph{where}\\validated cross-attention head};
\node[box,fill=aromaGeometry,text width=30*\aromaU] (geo) at (71.5,0)
  {\textbf{Representation geometry}\\\emph{which geometry}\\low-rank subspace or whole head};
\draw[arrow] (vlm.east)--(act.west);
\draw[arrow] (act.east)--(geo.west);
\node[band] at (.8,-2.84) {POLICY};
\node[box,fill=aromaPolicy,text width=13*\aromaU] (gate) at (10.5,-2.84)
  {\textbf{Gate}\\\emph{when}\\apply or abstain};
\draw[draw=aromaRule,densely dashed,rounded corners=2pt,line width=.5pt] (25,-1.86) rectangle (81,-3.82);
\node[lane,fill=aromaDecision,text width=21*\aromaU] (tgt) at (38,-2.46)
  {\textbf{Target selection}\\\emph{which target}: runner-up $t$};
\node[lane,fill=aromaGeometry,text width=27*\aromaU] (mea) at (66,-2.46)
  {\textbf{Minimal-Effort Actuation}\\\emph{how}: stepwise re-linearization};
\node[slim,fill=aromaPolicy,text width=53*\aromaU] (rout) at (53,-3.41)
  {\textbf{Candidate routing}\enspace\emph{which geometry and gain}: MN, SA, or WHOLE};
\node[box,fill=aromaPlain,text width=11*\aromaU] (out) at (92.5,-2.84)
  {\textbf{Audit}\\repair, break,\\neutral, abstain};
\draw[arrow] (tgt.east)--(mea.west);
\draw[arrow] (vlm.south)--(gate.north);
\draw[arrow] (gate.east)--node[lab,above]{apply}(25,-2.84);
\draw[arrow] (81,-2.84)--(out.west);
\draw[arrow] (geo.south)--node[lab,right]{head-space edits $\Delta h$}(71.5,-1.86);
\draw[arrow,densely dashed] (gate.south)--(10.5,-4.22)--node[lab,above,pos=.5]{abstain: identity, no edit}(92.5,-4.22)--(out.south);
\node[lab,anchor=south west] at (25.3,-1.86) {post-gate control law (one of two)};
\end{tikzpicture}
\caption{\textbf{The AROMA control stack.} \emph{Mechanism} (top): a frozen VLM exposes a causally validated cross-attention head; its Top-4 sensitivity subspace supports magnitude-normalized MN/SA edits, while WHOLE is an alternative full-head intervention geometry on the same actuator. \emph{Policy} (bottom): a ground-truth-free gate decides \emph{when} to act; on gated examples, either a target rule selects \emph{which target} to pursue and \mea{} determines \emph{how} to reach it, or a candidate router selects geometry and gain without an explicit target.}
\label{fig:aroma_stack}
\end{figure*}

Figure~\ref{fig:aroma_stack} shows the full control stack. A frozen VLM exposes a causally validated actuator (\S\ref{sec:formulation}); its sensitivity geometry defines low-rank edit directions (\S\ref{sec:mncsa}); a local response law and Minimal-Effort Actuation reach explicit targets (\S\ref{sec:mea_method}); and a gate, a target rule, and a router decide when, toward what, and with which geometry to act (\S\ref{sec:routing}).

\begin{figure*}[!t]
\centering
\pgfmathsetlengthmacro{\csaHalf}{0.5*\textwidth}\pgfmathsetlengthmacro{\csaWidth}{0.5*\textwidth-\aromaPairPad}
\pgfmathsetlengthmacro{\csaRight}{\csaHalf+1.15cm}
\begin{tikzpicture}
\pgfplotsset{csaaxis/.style={scale only axis,width=\csaWidth,height=\aromaHa,anchor=north west,
axis x line*=bottom,axis y line*=left,axis line style={draw=aromaInk!60,line width=.45pt},
tick style={draw=aromaInk!60,line width=.4pt},tick align=outside,major tick length=2pt,
tick label style={font=\rmfamily\fontsize{7.5}{9}\selectfont,text=aromaInk},
label style={font=\rmfamily\fontsize{8}{9.5}\selectfont,text=aromaInk},ymajorgrids=true,
grid style={draw=aromaGrid,line width=.35pt},scaled ticks=false,clip=true}}
\tikzset{csatitle/.style={anchor=south west,align=left,text width=\dimexpr.5\textwidth-10pt\relax,inner sep=0pt,text=aromaInk,font=\rmfamily\fontsize{9}{11}\selectfont\bfseries}}

\node[csatitle] at (0,0) {(a) Sensitivity concentration by rank};
\begin{axis}[csaaxis,at={(1.0cm,-.90cm)},xmin=.7,xmax=8.3,ymin=0,ymax=1.04,
xlabel={Top-$r$ sensitivity directions},ylabel={Explained sensitivity},
xtick={1,2,3,4,5,6,7,8},xticklabels={1,2,4,8,16,32,64,128},
ytick={0,.2,.4,.6,.8,1.0},legend style={font=\rmfamily\fontsize{7.5}{9}\selectfont,draw=none,fill=none,at={(.5,1.02)},anchor=south,legend columns=2,/tikz/every even column/.append style={column sep=5pt}}]
\addplot[color=aromaBlue,line width=1pt,mark=*] coordinates {(1,.76216173)(2,.84171753)(3,.89822936)(4,.94462435)(5,.97719458)(6,.99346874)(7,.99923134)(8,1.0)};
\addplot[color=aromaOrange,line width=1pt,mark=square*] coordinates {(1,.69528345)(2,.79779084)(3,.86777530)(4,.92797816)(5,.97030351)(6,.99150319)(7,.99900109)(8,1.0)};
\legend{Total second moment,Centered residual}
\draw[densely dashed,aromaGrayMid,line width=.55pt] (axis cs:.7,.219176)--(axis cs:8.3,.219176);
\ifaclstyle
\node[font=\rmfamily\fontsize{7.5}{9}\selectfont,text=aromaInk!75,anchor=north east] at (axis cs:8.25,.195) {rank-4 null 99th pct. $\approx0.219$};
\else
\node[font=\rmfamily\fontsize{7.5}{9}\selectfont,text=aromaInk!75,anchor=south] at (axis cs:4.2,.245) {rank-4 null 99th pct. $\approx0.219$};
\fi
\ifaclstyle
\node[font=\rmfamily\fontsize{7.5}{9}\selectfont\bfseries,text=aromaBlue,anchor=south west] at (axis cs:2.3,.53) {Top-4: $89.82\%$};
\node[font=\rmfamily\fontsize{7.5}{9}\selectfont\bfseries,text=aromaOrange,anchor=north west] at (axis cs:2.3,.49) {Centered: $86.78\%$};
\else
\node[font=\rmfamily\fontsize{7.5}{9}\selectfont\bfseries,text=aromaBlue,anchor=south west] at (axis cs:1.0,.58) {Top-4: $89.82\%$};
\node[font=\rmfamily\fontsize{7.5}{9}\selectfont\bfseries,text=aromaOrange,anchor=north west] at (axis cs:1.0,.52) {Centered: $86.78\%$};
\fi
\end{axis}

\node[csatitle] at (\csaHalf,0) {(b) Activation-norm share of learned Top-4};
\begin{axis}[csaaxis,at={(\csaRight,-.90cm)},xmin=-.2,xmax=6.2,ymin=0,ymax=.56,
xlabel={Quantile (\%) of $\rho=\|P_Uh\|/\|h\|$},ylabel={$\rho$},
xtick={0,1,2,3,4,5,6},xticklabels={0,10,25,50,75,90,100},
ytick={0,.1,.2,.3,.4,.5}]
\addplot[color=aromaGreen,line width=1pt,mark=*] coordinates {(0,.054286)(1,.232631)(2,.280526)(3,.308572)(4,.339287)(5,.379001)(6,.510192)};
\draw[densely dashed,aromaGrayMid,line width=.55pt] (axis cs:-.2,.308572)--(axis cs:6.2,.308572);
\node[font=\rmfamily\fontsize{7.5}{9}\selectfont\bfseries,text=aromaGreen!70!black,anchor=south east] at (axis cs:5.9,.08) {median $\rho=0.309$};
\node[font=\rmfamily\fontsize{7.5}{9}\selectfont,text=aromaInk,anchor=north west,align=left] at (axis cs:.15,.52) {Median energy share\\$\rho^2=9.52\%$};
\end{axis}
\end{tikzpicture}
\caption{\textbf{Sensitivity is not activation energy.}
\textbf{(a)} The top four dimensions explain 89.82\% of the uncentered cardinality-sensitivity second moment and 86.78\% after centering; both are far above the rank-matched null distribution.
\textbf{(b)} The same Top-4 subspace contains little of the pooled activation norm: median $\rho=0.309$, corresponding to only 9.52\% median projected activation energy. This mismatch predicts why same-$\alpha$ low-rank scaling under-actuates and motivates magnitude normalization.}
\label{fig:csa_geometry}
\end{figure*}
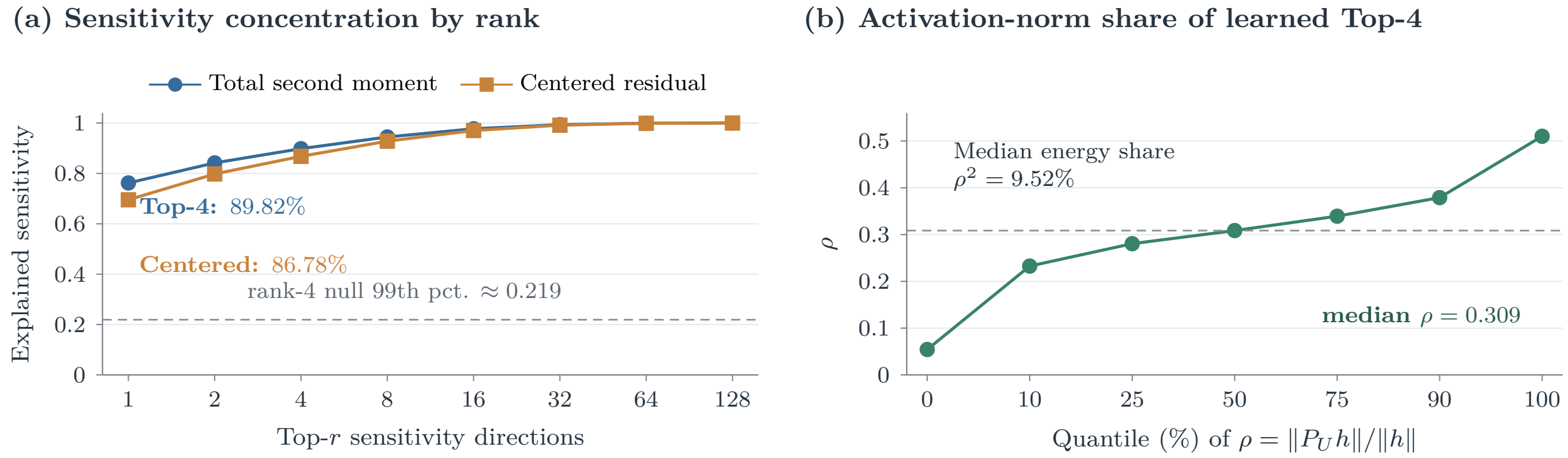

\subsection{Problem Formulation}
\label{sec:formulation}

\paragraph{Control objective.}
For an image--question pair $x$ with integer answer $y$, let $c_i(a)=\ind[\hat y_a(x_i)=y_i]$ denote correctness under action $a$, with $a=1$ the identity. Realized repair utility is $u_i(a)=c_i(a)-c_i(1)\in\{-1,0,+1\}$, so a controller trades repairs against breaks rather than maximizing intervention magnitude.

\paragraph{Actuators.}
Let $H=[H^{(1)};\ldots;H^{(m)}]$ be the concatenated cross-attention head state immediately before its output projection, and reserve \emph{baseline} for the unmodified model. A localized head can be scaled as a whole, $H^{(j)}_{\alpha}=\alpha H^{(j)}$, where $\alpha=1$ is identity, $0<\alpha<1$ attenuates, and $\alpha>1$ amplifies. We call this \textbf{WHOLE} or \emph{full-head actuation}; it is one intervention geometry on the same actuator and serves throughout as a strong comparator to low-rank geometries. A candidate actuator is characterized by the normalized expected numeral $\mu_\alpha(x)$ over numeral support $\mathcal V_N$ and qualifies through bidirectional finite-step behavior under intervention, beyond attention or gradient saliency.

\subsection{Sensitivity Geometry and Magnitude-Normalized Actuation}
\label{sec:mncsa}

For the selected head, let $H(x)\in\R^{T\times d}$ be the token-level head output and $h(x)=T^{-1}\sum_t H_t(x)$ its pooled coordinate. Predictions are the argmax of $p(n\mid x)$, a softmax over the logits of the 16 numerals $0,\ldots,15$ at the final prompt position. Because every edit adds one shared offset to all token states, we define local cardinality sensitivity as the gradient with respect to that offset,
\begin{equation}
g(x)=\nabla_{\delta}\,\mu\big(H(x)+\mathbf{1}_T\delta^{\top}\big)\Big|_{\delta=0}=\textstyle\sum_{t=1}^{T}\nabla_{H_t}\mu(x),
\label{eq:shared_offset_grad}
\end{equation}
where $\mu$ is the normalized expected numeral over the procedural support $\{1,\ldots,10\}$; the same shared-offset convention is used for every margin gradient below. Its second moment is $S=\frac1N\sum_{i=1}^N g_i g_i^\top$, with $U_r$ the top-$r$ eigenspace and $P_U=U_rU_r^\top$ its projector. $S$ proposes \emph{candidate directions} inside an actuator already validated by intervention, and causal status is established by intervention. To quantify how much of the actuator state the subspace can act on, we measure the projected activation-energy share $\rho(x)^2$, where $\rho(x)=\|P_Uh(x)\|_2/\|h(x)\|_2$.

A naive projected analogue of whole-head scaling is $\Delta h_{\rm naive}=(\alpha-1)P_Uh$. When $U_r$ captures high sensitivity but little activation norm (Fig.~\ref{fig:csa_geometry}), this edit is much smaller than whole-head scaling. With the activation-aligned direction $d_{\mathrm{MN}}=P_Uh/\|P_Uh\|_2$, \method{} therefore uses magnitude-normalized causal subspace actuation (\mncsa{}),
\begin{equation}
\boxed{\Delta h_{\rm MN}=(\alpha-1)\|h\|_2\, d_{\mathrm{MN}}},
\label{eq:mncsa}
\end{equation}
so $\|\Delta h_{\rm MN}\|_2=|\alpha-1|\|h\|_2$, matching the pooled displacement magnitude of whole-head scaling. Because $d_{\mathrm{MN}}$ is collinear with $P_Uh$, Eq.~\eqref{eq:mncsa} equals the naive edit rescaled per example by $1/\rho(x)$: \mncsa{} is a closed-form equal-norm calibration that fixes the low-rank displacement to the one already validated for whole-head scaling, so it requires no gain search, and the mismatch $\rho$ predicts how far naive projection falls short. We apply the same $\Delta h$ to every token state, preserving token residual differences. A gain $\beta$ rescales the edit, $\Delta h_{\beta}=\beta\,\Delta h_{\rm MN}$: $\beta=1$ is the equal-norm test, and $\beta>1$ is a stronger edit reserved for calibrated comparisons.

\subsection{Local Response Law and Minimal-Effort Actuation}
\label{sec:mea_method}

\paragraph{Local response.}
For a small edit $\Delta h$ in the Top-4 subspace, the response obeys the first-order law $\mu(h+\Delta h)-\mu(h)\approx g^\top\Delta h$ with $g=\nabla_h\mu(h)$. Among equal-norm directions in the subspace, the sensitivity-aligned direction $d_{\mathrm{SA}}=P_Ug/\|P_Ug\|_2$ is locally optimal for increasing $\mu$, whereas $d_{\mathrm{MN}}$ is activation-aligned. Their cosine $\eta=d_{\mathrm{SA}}^\top d_{\mathrm{MN}}=\cos(P_Ug,P_Uh)$ is the relative local efficiency of \mncsa{}; it diagnoses local geometry, and finite-gain correctness is measured directly.

\paragraph{Target-specific margin control.}
For an explicit target numeral $t$, let $c$ be the current highest-scoring non-target numeral, $m_{c\rightarrow t}(h)=\ell_c(h)-\ell_t(h)$ the logit margin, and $q=P_U\nabla_h m_{c\rightarrow t}$ its projected gradient. The minimum-norm first-order step that reaches the local decision boundary is $-(m_{c\rightarrow t}/\|q\|_2^2)\,q$, the linearized boundary step of DeepFool \citep{moosavi2016deepfool}. Because the local law loses calibration at finite scale (\S\ref{sec:res_mea}), \mea{} applies this step inside a trust region,
\begin{equation}
\Delta h_{\rm MEA}
=-\min\!\left(\frac{m_{c\rightarrow t}}{\|q\|_2},\tau\right)
\frac{q}{\|q\|_2},
\label{eq:mea_step}
\end{equation}
with trust radius $\tau=0.10B$ and per-example reference budget $B=|\alpha_{\rm prop}-1|\|h\|_2$ (with $\|h\|_2$ taken at the current state), where $\alpha_{\rm prop}$ is the Stage-1 proposal of \S\ref{sec:routing}. $B$ is the displacement scale that Stage~1 has already proposed for this example, so the trust region keeps each step a small fraction of an intervention size whose effects are audited elsewhere. The model is re-evaluated after every step; $c$, the margin, and $q$ are recomputed for at most 20 iterations. \mea{} takes each step inside the validated four-dimensional subspace, under a trust region, with the competitor re-selected and the margin re-linearized after every step.

\paragraph{Oracle and deployable targets.}
Oracle \mea{} sets $t$ to the ground-truth numeral and measures reachable actuator capacity. Deployable \mea{} never observes ground truth: on gated examples (\S\ref{sec:routing}) it pursues the runner-up target
\begin{equation}
 t_{\rm run}(x)=\arg\max_{n\in\{0,\ldots,15\},\,n\neq b(x)}p(n\mid x),
\label{eq:runnerup_target}
\end{equation}
where $b(x)$ is the Stage-1 baseline prediction and ties choose the smaller numeral. The rule is deliberately simple, so that \emph{which target} can be studied separately from \emph{how to reach} it.

\subsection{Deployment Policy: Gating and Routing}
\label{sec:routing}

\paragraph{Stage 1: Proposal Controller.}
A \textbf{Proposal Controller} maps 39 ground-truth-free features (numeral probabilities, uncertainty statistics, and geometric summaries) to a coarse cardinality action $\alpha_{\rm prop}\in\{0,1,1.5,2,4\}$. Proposals with $\alpha_{\rm prop}\leq1$ are executed directly; upward proposals ($\alpha_{\rm prop}>1$) are refined by Stage~2.

\paragraph{Stage 2: candidate-level utility routing.}
Stage~2 selects, per example, among abstention (IDENTITY) and candidate edits $\Delta h_{m,\beta}=\beta(\alpha_{\rm prop}-1)\|h\|_2\,d_m$. The \textbf{Candidate Utility Router} operates within the low-rank family $m\in\{\mathrm{MN},\mathrm{SA}\}$ (15 decisions per example); the \textbf{Unified Geometry Router} (\ugr{}) additionally routes the intervention family itself, $m\in\{\mathrm{MN},\mathrm{SA},\mathrm{WHOLE}\}$ (43 candidates per example, with WHOLE applied at the matched scale). Each router scores every candidate $c$ with a candidate-conditioned L2-regularized logistic model $Q_\theta(x,c)=\Pr[\hat y_c(x)=y\mid\Phi(x,c)]$ over ground-truth-free features and deploys $c^*(x)=\arg\max_{c}Q_\theta(x,c)$, breaking ties toward IDENTITY and smaller gain. For natural images, \ugr{} is factorized into a gate that decides whether to keep the identity prediction or apply an edit and a pairwise ranker that selects among non-identity candidates when the gate fires. Ground truth is used only to fit and evaluate routers; gain grids are given in Appendix~\ref{app:routing}.

\paragraph{Two deployment regimes.}
\method{} supports two post-gate control laws. \textbf{Candidate routing} constructs no explicit target: it decides \emph{when} to intervene and selects a bounded geometry/gain action. \textbf{Targeted \mea{}} keeps the same gate, uses Eq.~\eqref{eq:runnerup_target} for \emph{which target}, and uses iterative margin control for \emph{how to reach}. The hierarchical \ugr{} policy, which we call the \emph{reference} \ugr{}, is therefore the natural reference for targeted \mea{}: both act on the identical gated set, and only the post-gate control law changes. An upward proposal $\alpha_{\rm prop}>1$ triggers Stage~2 and sets a reference budget; the final prediction may move in either direction.

\section{Experimental Setup}
\label{sec:setup}

\paragraph{Models and scoring.}
The primary pipeline uses frozen \modelname{} \citep{meta2024llama32} with a deterministic numeral proxy over 0--15. Cross-model experiments use frozen \idefics{}, a Flamingo-style model with gated cross-attention blocks \citep{alayrac2022flamingo,huggingface2023idefics}, scored by teacher-forced sequence log-likelihood over complete numeral strings. We compare directions and effects within each model.

\paragraph{Data.}
Procedural counting uses \proc{} cohorts with zero seed and pixel-hash overlap: v1 ($N=300$) for mechanism discovery, v2 ($N=1000$) for controller development, v3 ($N=2000$) for confirmation and CSA development, and v4 ($N=2000$), generated only after \mncsa{} was fixed. Routers are developed and confirmed on separately generated cohorts. Natural evaluation uses TallyQA \citep{acharya2019tallyqa,goyal2017vqa,krishna2017visualgenome}: disjoint samples for zero-shot transfer and Stage-1 recalibration, a held-out 8,000-example cohort for natural \ugr{}, and a second held-out, pixel-deduplicated cohort of $N=7{,}992$, constructed only after runner-up targeting and \mea{} were fully specified and sharing no question, image ID, or pixels with earlier natural data. HoloCount ($N=2480$) is an external benchmark and GQA \citep{hudson2019gqa} an off-target stress test.

\paragraph{Evaluation protocol and evidence levels.}
We call a result \emph{prospective} when its method, comparator, endpoint, and statistical test were fixed before the evaluation cohort was run; the local-response audit and oracle \mea{} validate the mechanism and controller. Paired correctness is tested with two-sided exact McNemar tests, with paired bootstrap intervals where these were part of the planned analysis; directional steering uses exact binomial tests with Holm correction. Geometry baselines, free-generation checks, and output-space controls reuse existing cohorts; their directions and gains are fit or selected on development data only, each test configuration is fixed before evaluation and run once, and all are marked post hoc (Appendix~\ref{app:additional}). Oracle \mea{} runs in float32 and natural-image \mea{} in BF16. Stage-2 primary endpoints are conditional by design, since the compared policies act on the prespecified upward-proposal branch; we additionally report an identity-fallback reconstruction over the full natural cohort, computed from the same per-example predictions. The main text reports percentages and percentage-point (pp) differences to at most two decimals and $p$-values to two significant figures; appendix tables report values at full precision. Appendix~\ref{app:timeline} gives the full timeline and statistics, and Table~\ref{tab:headline} lists every headline result with its evidence level.

\section{Results}
\label{sec:results}

\subsection{Gap 1: Sensitivity Is Not Energy}
\label{sec:res_mncsa}

\paragraph{Actuator discovery.}
Causal screening over 256 multimodal cross-attention heads yields seven candidates, and held-out intervention evidence selects \headname{} as the primary actuator. On $N=300$, $\alpha=0.5$ lowers $\mu$ on 91.33\% of examples, $\alpha=1.5$ raises it on 90.67\%, and $\mu_{0.5}<\mu_1<\mu_{1.5}$ holds on 88.67\%.

\paragraph{The sensitivity--energy mismatch.}
For $r=4$, the learned subspace explains 89.82\% of the uncentered sensitivity second moment (86.78\% after centering), above all 10,000 rank-matched isotropic null replicates. Across 653 non-\noop{} v3 examples, its median projected activation-energy share is only $\rho^2=9.52\%$ (Fig.~\ref{fig:csa_geometry}). The mismatch predicts the behavior of same-$\alpha$ projected scaling quantitatively: naive projection moves the state by only $\rho\,|\alpha-1|\,\|h\|_2$, a threefold displacement deficit at the median. On the development intervention subset, naive projection reaches 33.23\%, whereas \mncsa{} reaches 52.53\% ($+19.30\pp$, McNemar $p=2.6\times10^{-28}$) and matches whole-head actuation (51.00\%; $p=0.24$). The recovery is specific to the sensitivity subspace: under identical rank and displacement, Bottom-4 reaches 26.80\% and 20 random rank-4 subspaces drawn in advance average 27.18\% (maximum 30.78\%).

\paragraph{Held-out confirmation.}
On v4, generated only after \mncsa{} was fixed, \mncsa{} improves accuracy from 50.25\% to 58.25\% ($+8.00\pp$, 95\% CI $[+6.7,+9.3]\pp$, McNemar $p=2.7\times10^{-36}$) and exceeds same-policy whole-head actuation (57.25\%) by $+1.00\pp$ (95\% CI $[+0.3,+1.7]\pp$, $p=0.0078$), with every visual condition improving over baseline (Table~\ref{tab:v4_condition}). A four-dimensional subspace of the 128-dimensional head thus delivers the finite control of the full head once its displacement is magnitude-matched.

\paragraph{Comparison with other geometries.}
We compare it against two strong directions that require no gradients, both fit only on the discovery cohort v1 and applied at the same site with identical per-example displacement: a mean-difference count direction (high-count minus low-count pooled activations, as in contrastive activation addition; \citealp{rimsky2024caa}) and the top-four activation principal components. On v4 under the Stage-1 policy, \mncsa{} (58.25\%) exceeds the mean-difference direction (55.45\%) by $+2.80\pp$ (95\% CI $[+1.95,+3.70]$; 67:11, $p=6.1\times10^{-11}$) and activation PCA-4 (52.75\%) by $+5.50\pp$ ($[+4.40,+6.65]$; 125:15, $p=8.9\times10^{-23}$), with 160 net repairs versus 104 and 50. With development-tuned gains, all geometries, including naive projection, reach within $2.5\pp$ of one another (Appendix~\ref{app:add_geometry}): gain acts as a separate calibration variable, and the sensitivity subspace reaches its accuracy at the smallest displacement and with no gain search.

\paragraph{Replication across heads and models.}
\label{sec:res_replication}
Fixing the complete seven-head candidate pool with no head-specific tuning, six heads, including five of six non-primary candidates, satisfy the bidirectional criterion on a fresh $N=300$ cohort after Holm correction across 14 directional tests (Appendix~\ref{app:multi_actuator}). Repeating the search in \idefics{}, rather than transplanting Llama coordinates, identifies head B2H28, whose raw Top-4 subspace explains 91.36\% of sensitivity trace while carrying only 4.14\% mean activation energy; with a development-selected gain ($r=4$, $\beta=2$), low-rank actuation recovers 81.09\%/81.49\% of same-run whole-head mean effect on a held-out cohort (Appendix~\ref{app:idefics}). Low-rank causal geometry replicates across architectures, and finite gain is calibrated per model.

\subsection{Gap 2: Local Is Not Finite}
\label{sec:res_mea}

\begin{figure*}[!t]
\centering
\begin{tikzpicture}
\pgfplotsset{meax/.style={scale only axis,width=\aromaTripleW,height=\aromaHb,axis x line*=bottom,axis y line*=left,
  tick align=outside,axis line style={draw=aromaInk!60,line width=.45pt},tick style={draw=aromaInk!60},
  tick label style={font=\rmfamily\fontsize{7.5}{9}\selectfont},label style={font=\rmfamily\fontsize{7.5}{9}\selectfont},
  title style={font=\rmfamily\fontsize{8.5}{10.5}\selectfont\bfseries,align=center,text width=.29\textwidth,yshift=8pt},ymajorgrids=true,grid style={draw=aromaRule!30}}}
\begin{axis}[meax,at={(1.0cm,0)},anchor=north west,title={(a) Local first-order calibration},xmin=.035,xmax=.215,ymin=.90,ymax=1.005,
  xlabel={$\beta$},ylabel={Pearson $r$},xtick={.05,.10,.20},xticklabels={0.05,0.10,0.20},ytick={.90,.94,.98,1.00}]
\addplot[color=aromaBlue,mark=*,mark size=2pt,line width=1pt] coordinates {(.05,.9936)(.10,.9769)(.20,.9371)};
\end{axis}
\begin{axis}[meax,at={(.32\textwidth+1.0cm,0)},anchor=north west,title={(b) Oracle MEA reachability},xmin=0,xmax=21,ymin=0,ymax=90,
  xlabel={Max iterations},ylabel={Target success (\%)},xtick={1,5,10,15,20},ytick={0,20,40,60,80}]
\addplot[color=aromaGreen,mark=*,mark size=2pt,line width=1pt] coordinates {(1,5.36)(2,10.71)(3,12.50)(5,20.54)(10,55.36)(15,75.89)(20,81.25)};
\end{axis}
\begin{axis}[meax,at={(.64\textwidth+1.0cm,0)},anchor=north west,title={(c) Matched-budget controls},ybar,bar shift=0pt,bar width=15pt,enlarge x limits=.22,
  symbolic x coords={MEA,Frozen,Activation},xtick={MEA,Frozen,Activation},xticklabels={MEA,{Fixed\\direction},{Activation\\aligned}},xticklabel style={align=center},
  ymin=0,ymax=90,ylabel={Success (\%)},ytick={0,20,40,60,80},nodes near coords,every node near coord/.append style={font=\rmfamily\fontsize{7.5}{9}\selectfont},legend image code/.code={\path[#1] (0pt,-2pt) rectangle (8pt,3pt);}]
\addplot[draw=none,fill=aromaGreen] coordinates {(MEA,81.25)};
\addplot[draw=none,fill=aromaBlue] coordinates {(Frozen,15.18)};
\addplot[draw=none,fill=aromaGray] coordinates {(Activation,0.89)};
\end{axis}
\end{tikzpicture}
\caption{\textbf{Local projected geometry is quantitatively valid at small scale, but global target control requires iterative re-linearization.} (a) Pearson agreement between first-order predicted and realized numeral movement is near-perfect at $\beta=.05$ and declines as intervention scale grows. (b) Oracle \mea{} accumulates reachability as it recomputes the current target--competitor margin and projected gradient. (c) Matched-budget controls under the same realized per-step norms: iterative \mea{}, a fixed initial target-gradient direction, and activation-aligned control.}
\label{fig:mea_local_global}
\end{figure*}
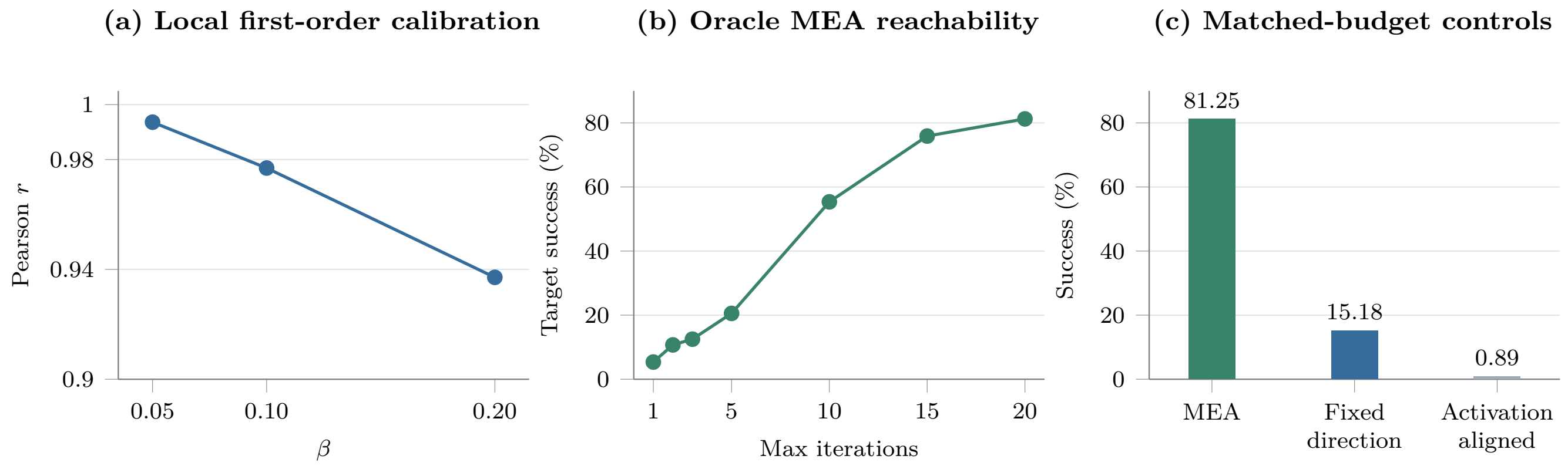

\paragraph{Local response law.}
On the $N=160$ continuous-response cohort, the median cosine between the activation- and sensitivity-aligned directions is 0.79, so \mncsa{} retains 79\% of the locally optimal efficiency. At $\beta=0.05$, observed response versus first-order prediction has Pearson $r=0.9936$, Spearman $\rho=0.9990$, and 100\% sign agreement (Fig.~\ref{fig:mea_local_global}a); correlation remains high at $\beta=0.10$ ($r=0.9769$) and $0.20$ ($r=0.9371$) while magnitude error grows. The projected gradient is a valid \emph{local} control object: what changes with scale is finite-step calibration, while the derivative stays accurate.

\paragraph{Oracle reachability.}
On $N=112$ baseline-wrong examples, oracle \mea{} reaches the ground-truth target on $91/112=81.25\%$ within 20 trust-region steps, accumulating from 5.36\% after one step to 55.36\% after ten (Fig.~\ref{fig:mea_local_global}b). Given exactly the realized \mea{} step norms, a fixed initial target-gradient direction reaches 15.18\% (exact paired $p=1.1\times10^{-22}$), and an iterative activation-aligned control 0.89\% ($p=1.6\times10^{-27}$; Fig.~\ref{fig:mea_local_global}c). Path length and norm cannot explain the gain; what matters is recomputing the target-specific projected margin, including the active competitor, at every step. Reached edits stay compact: median net displacement is $0.99B$, and 88\% of oracle trajectories end within $4B$, the largest gain already audited in the routing candidate grid (Appendix~\ref{app:add_cost}).

\paragraph{Free-form generation.}
All accuracies above use the closed-set numeral proxy. To test whether \mea{} changes behavior beyond a score, we re-run the same protocol, which reproduces every original final prediction exactly (175/175), and let the edited model answer by greedy free-form generation. With the edit applied, generated answers are correct on 80/112 oracle examples versus 0/112 without it (80:0 paired discordance, exact $p=1.7\times10^{-24}$; 79.5\% agreement with the proxy argmax, with most disagreements one numeral short because \mea{} stops at the decision boundary). Applying the edit only at prompt positions or also at every generated position yields the same answer on 174/175 examples, so the edited prompt state alone carries the change. Iterative target control moves the numeral the model actually says.

\paragraph{Whole-head ablation.}
Running \mea{} on the full 128-dimensional head instead of the Top-4 subspace, with every other setting identical, leaves oracle reachability unchanged (92 versus 91 of 112; 10:9 discordance, exact $p=1.0$), and generated answers are correct on 88 versus 80 of 112 ($p=0.13$; Appendix~\ref{app:wholehead_mea}). The sensitivity subspace carries essentially all of the head's iterative controllability.

\paragraph{Replication in \idefics{}.}
We repeat oracle \mea{} in \idefics{} at head B2H28 with its own Top-4 basis, select only the trust-region scale on development data, and evaluate once on 500 newly generated images with zero seed and pixel overlap with every earlier cohort (Appendix~\ref{app:idefics_mea}). On the 470 baseline-wrong examples, \mea{} reaches the ground-truth target on 250 (53.2\%), against 2.8\% for a fixed direction and 1.5\% for an activation-aligned direction with identical step norms (238:1 and 243:0 discordance; exact $p=5.4\times10^{-70}$ and $1.4\times10^{-73}$). The edits again change behavior: correct greedy answers rise from 124 to 218 of 470 (97:3; $p=2.6\times10^{-25}$), with no parse failures. In this head the first-order law remains near-exact over the tested edit sizes ($r\geq0.998$), yet a fixed direction still fails, so the advantage of re-linearization does not rest on strong curvature: recomputing the target-specific margin and its active competitor at every step is what a fixed direction lacks.

\subsection{Gap 3: Reachable Is Not Useful}
\label{sec:deployable_mea}

Gaps~1 and~2 establish that the actuator can be moved and steered to a chosen numeral. This section asks what then limits utility. Once execution is reliable, deployment accuracy is set by \emph{when} to act and \emph{which} target to pursue, and the controls below show that the deciding signal is already present in the model's own numeral distribution. Because \mea{} realizes almost every assigned target, the gate and the target rule become the operative decisions, and \mea{} establishes that the selected answer is causally realizable through the validated actuator and generated by the model itself.

\paragraph{Gating and routing.}
\label{sec:res_routing}
The Stage-1 Proposal Controller improves held-out v3 from 49.95\% to 58.05\% ($+8.10\pp$; 173 repairs against 11 breaks). Within Stage~2, MN, SA, and WHOLE reach nearly tied peak accuracies (61.4\%, 61.0\%, 61.6\%) at different gains, so the edit is chosen per example. On a held-out upward-proposal population ($N=294$), the Candidate Utility Router improves a global MN comparator ($\beta=2$) from 60.54\% to 66.33\% ($+5.78\pp$, McNemar $p=1.5\times10^{-5}$) with 41\% fewer breaks (16 vs.\ 27). \ugr{} then routes the intervention family itself: on its held-out final Stage-2 population ($N=633$), it reaches 73.14\% versus 70.93\% for the stronger single-geometry router selected in development ($+2.21\pp$, McNemar $p=0.038$; 27:13 paired discordance) while using every intervention family (Appendix~\ref{app:routing}). On natural images, the Top-4 geometry transfers to TallyQA without refitting ($+1.40\pp$ over baseline for \mncsa{}), while the Stage-1 utility map is domain-specific: the dominant error type reverses from undercounts to overcounts, the procedurally trained map changes accuracy by $-2.98\pp$, and recalibrating only this map on $K=500$ natural examples raises accuracy on a disjoint 4,000-example set from 64.9\% to 67.1\% (85 more correct answers; $p=9.2\times10^{-11}$). Fixed and evaluated on the prespecified Stage-2 population of a new 8,000-example cohort ($N=631$), hierarchical \ugr{} reaches 35.34\% versus 33.44\% for hierarchical Whole-only routing and 31.85\% for identity ($+1.90\pp$, McNemar $p=0.0042$; 14:2 paired discordance), intervening on 52/631 examples with 26 repairs against 4 breaks (Appendix~\ref{app:natural}). This hierarchical policy is the reference \ugr{} for targeted deployment.

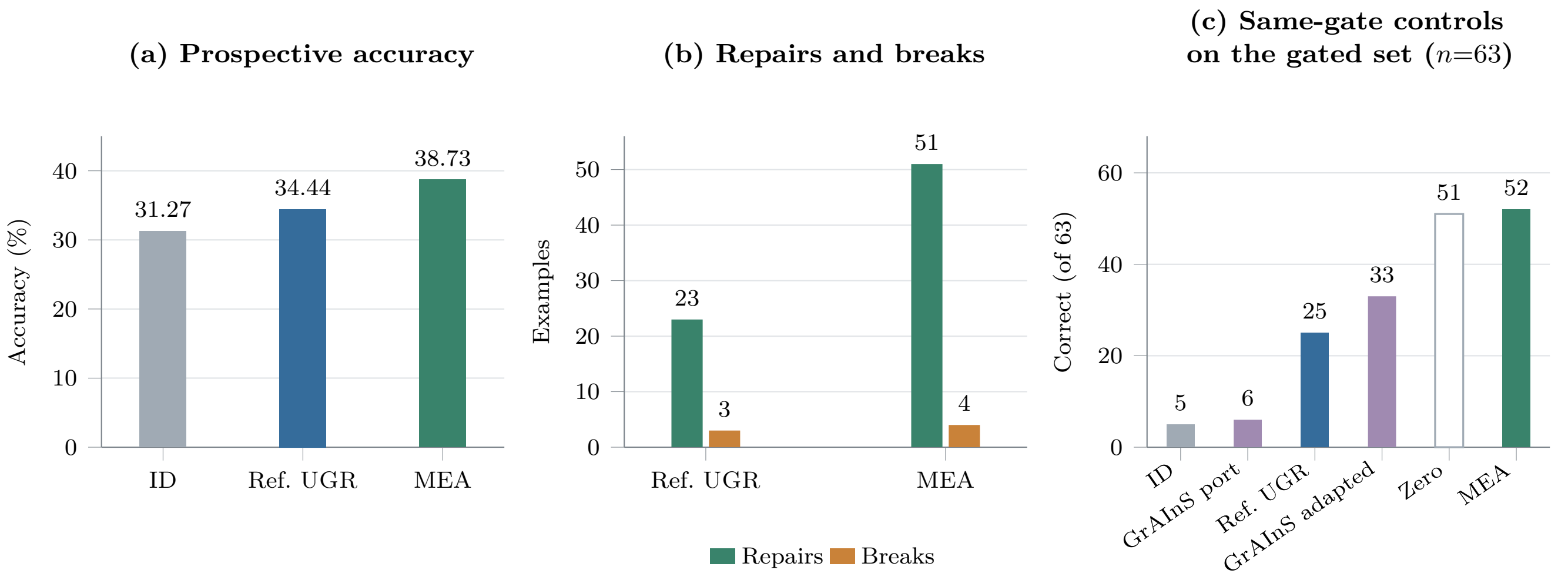
\begin{figure*}[!t]
\centering

\begin{tikzpicture}
\begin{groupplot}[group style={group size=3 by 1,horizontal sep=\aromaGroupSep},
 scale only axis,width=.25\textwidth,height=\aromaHc,
 axis x line*=bottom,axis y line*=left,tick align=outside,legend image code/.code={\path[#1] (0pt,-2pt) rectangle (8pt,3pt);}]
\nextgroupplot[
  title={(a) Prospective accuracy},
  title style={font=\rmfamily\fontsize{8.5}{10.5}\selectfont\bfseries,align=center,text width=.275\textwidth,yshift=12pt},
  ybar,bar shift=0pt,bar width=15pt,ymin=0,ymax=45,
  ytick={0,10,20,30,40},ylabel={Accuracy (\%)},
  symbolic x coords={ID,Old,MEA},xtick={ID,Old,MEA},enlarge x limits=0.22,
  xticklabels={ID,{Ref.\ \ugr{}},MEA},
  tick label style={font=\rmfamily\fontsize{7.5}{9}\selectfont},
  label style={font=\rmfamily\fontsize{7.5}{9}\selectfont},
  axis line style={draw=aromaInk!60},tick style={draw=aromaInk!60},
  ymajorgrids=true,grid style={draw=aromaRule!30},
  nodes near coords,
  every node near coord/.append style={font=\rmfamily\fontsize{7.5}{9}\selectfont\bfseries,anchor=south,yshift=1pt}]
\addplot[draw=none,fill=aromaGray] coordinates {(ID,31.270)};
\addplot[draw=none,fill=aromaBlue] coordinates {(Old,34.444)};
\addplot[draw=none,fill=aromaGreen] coordinates {(MEA,38.730)};
\nextgroupplot[
  title={(b) Repairs and breaks},
  title style={font=\rmfamily\fontsize{8.5}{10.5}\selectfont\bfseries,align=center,text width=.275\textwidth,yshift=12pt},
  ybar,bar width=10pt,ymin=0,ymax=56,
  ytick={0,10,20,30,40,50},ylabel={Examples},
  symbolic x coords={Old,MEA},xtick=data,enlarge x limits=0.34,
  xticklabels={{Ref.\ \ugr{}},MEA},
  tick label style={font=\rmfamily\fontsize{7.5}{9}\selectfont},
  label style={font=\rmfamily\fontsize{7.5}{9}\selectfont},
  axis line style={draw=aromaInk!60},tick style={draw=aromaInk!60},
  ymajorgrids=true,grid style={draw=aromaRule!30},
  legend image code/.code={\path[#1] (0pt,-2pt) rectangle (8pt,3pt);},
  legend style={draw=none,fill=none,font=\rmfamily\fontsize{7.5}{9}\selectfont,at={(.5,-.28)},anchor=north,legend columns=2},
  nodes near coords,
  every node near coord/.append style={font=\rmfamily\fontsize{7.5}{9}\selectfont\bfseries,anchor=south,yshift=1pt}]
\addplot[draw=none,fill=aromaGreen] coordinates {(Old,23)(MEA,51)};
\addplot[draw=none,fill=aromaOrange] coordinates {(Old,3)(MEA,4)};
\legend{Repairs,Breaks}
\nextgroupplot[
  title={(c) Same-gate controls\\on the gated set ($n{=}63$)},
  title style={font=\rmfamily\fontsize{8.5}{10.5}\selectfont\bfseries,align=center,text width=.275\textwidth,yshift=12pt},
  ybar,bar shift=0pt,bar width=9pt,ymin=0,ymax=68,
  ytick={0,20,40,60},ylabel={Correct (of 63)},
  symbolic x coords={ID,Port,Old,Gra,Zero,MEA},xtick={ID,Port,Old,Gra,Zero,MEA},enlarge x limits=0.10,
  xticklabels={ID,{GrAInS port},{Ref.\ \ugr{}},{GrAInS adapted},Zero,MEA},
  xticklabel style={font=\rmfamily\fontsize{7}{8.5}\selectfont,rotate=35,anchor=east},
  tick label style={font=\rmfamily\fontsize{7.5}{9}\selectfont},
  label style={font=\rmfamily\fontsize{7.5}{9}\selectfont},
  axis line style={draw=aromaInk!60},tick style={draw=aromaInk!60},
  ymajorgrids=true,grid style={draw=aromaRule!30},clip=false,
  nodes near coords,
  every node near coord/.append style={font=\rmfamily\fontsize{7.5}{9}\selectfont\bfseries,anchor=south,yshift=1pt}]
\addplot[draw=none,fill=aromaGray] coordinates {(ID,5)};
\addplot[draw=none,fill=aromaPurple!75] coordinates {(Port,6)(Gra,33)};
\addplot[draw=none,fill=aromaBlue] coordinates {(Old,25)};
\addplot[draw=aromaGray,line width=.6pt,fill=white] coordinates {(Zero,51)};
\addplot[draw=none,fill=aromaGreen] coordinates {(MEA,52)};
\end{groupplot}
\end{tikzpicture}
\par\vspace{3pt}
\caption{\textbf{Target quality converts controllability into prospectively confirmed deployment utility.} (a) Accuracy on the prespecified $N=630$ Stage-2 population ($+4.29\pp$ over the reference \ugr{}; 29:2 paired discordance; exact McNemar $p=4.6\times10^{-7}$). (b) Repairs and breaks relative to identity. (c) Correct predictions on the 63 gated cases under the same gate; the GrAInS minimal port, the counting-adapted GrAInS-inspired comparator, and the constant-zero rule are post hoc controls.}
\label{fig:deployable_mea}
\end{figure*}

\paragraph{An upward-only target rule.}
The first deployable target rule, $t=\arg\max_{n>b}p(n)$, restricts candidates to numerals larger than the baseline prediction. Across 623 targetable examples, \mea{} reaches 417 targets (66.93\%), and forced deployment changes accuracy by $-18.54\pp$ (17 repairs against 134 breaks); under the gate, breaks fall to four and one repair remains, because only one of the 48 gated targets equals ground truth. Reachability therefore amplifies whatever target rule it is given, which makes target construction an explicit design variable. Natural development data identify the runner-up rule of Eq.~\eqref{eq:runnerup_target}: on the 30 development gated decisions, the runner-up target matches ground truth on 24 (80.0\%), and 28 of the 30 require a downward correction. The rule is then fixed; the development record and a retrospective bridge evaluation are given in Appendix~\ref{app:target}.

\paragraph{Prospective confirmation on new natural data.}
With the target rule, gate, reference-\ugr{} actions, and \mea{} protocol all fixed, we evaluate the new cohort of $N=7{,}992$ questions. Stage~1 emits $N=630$ upward proposals, and the gate fires on 63 of them. On this Stage-2 population, identity is correct on 197/630 (31.27\%), the reference \ugr{} on 217/630 (34.44\%), and runner-up + \mea{} on 244/630 (38.73\%; Fig.~\ref{fig:deployable_mea}a). The primary paired improvement over the reference \ugr{} is $+4.29\pp$, with 29 \mea{}-only versus 2 reference-only correct examples (exact two-sided McNemar $p=4.6\times10^{-7}$). Relative to identity, the reference \ugr{} produces 23 repairs and 3 breaks, and targeted \mea{} produces 51 repairs and 4 breaks: the improvement comes from additional repairs at an essentially unchanged break count.

\paragraph{Execution on the gated set.}
The runner-up target equals ground truth on 51/63 (80.95\%) gated examples, and \mea{} reaches 61/63 (96.83\%) of assigned targets, including all 51 correct ones, in a median of 9 iterations against the 20-step horizon. The realized targets carry over to free-form generation: with the edit applied, the model's own greedy answer is correct on 54/63 gated examples versus 7/63 unedited (51:4 paired discordance, exact $p=2.1\times10^{-11}$; 95.2\% agreement with the proxy). The remaining error is therefore set by which target is selected. Execution costs $20.4\times$ a single forward pass at the median 9 steps ($42.9\times$ at 20), paid on 0.79\% of inputs.

\paragraph{Full-cohort utility.}
\label{sec:res_controls}
The Stage-2 $N=630$ is exactly the upward-proposal subset of the $N=7{,}992$ cohort: all 630 Stage-1 and Stage-2 identity predictions agree. Keeping identity predictions outside the 63 gated cases yields 5,254/7,992 correct (65.74\%) for runner-up + \mea{}, versus 5,207 (65.15\%) at identity and 5,227 (65.40\%) for the reference \ugr{}: 47 net correct predictions over identity ($+0.59\pp$; exact paired $p=2.0\times10^{-11}$) with 0.79\% of inputs edited. The conditional gain carries over to the whole natural cohort.

\paragraph{Locating the gain.}
Controls on the same gated set locate the deployment gain (Fig.~\ref{fig:deployable_mea}c; Appendix~\ref{app:target}). Emitting the selected runner-up numeral on the gated set gives 243/630, and a gate-conditional zero rule 51/63, against 244/630 and 52/63 for \mea{} (exact paired $p=1$): under closed-set scoring, the gate and the target rule carry the gain, as the saturation result predicts. \mea{} contributes what emission cannot: it realizes the target through the validated actuator, so the model itself generates the corrected answer (54/63 versus 7/63), and it is correct on 5/12 non-zero cases, against 4/12 for emission and none for the constant rule. Steering families differ sharply under the same gate: a counting-adapted GrAInS-inspired comparator is correct on 33/63 and an image-preserving GrAInS minimal port on 6/63, against 52/63 for \mea{} (244/630 versus 225/630 for the adapted comparator, $+3.02\pp$, exact paired $p=3.1\times10^{-4}$). Steering all 630 examples, the adapted comparator reaches 247/630 ($p=0.81$ versus \mea{}), so selective \mea{} matches a global steering policy while editing one tenth of the examples.

\paragraph{Zero-count overcounting.}
Of the 63 gated examples, 51 have true count zero: the learned gate reaches 81\% precision for zero-count questions, against base rates of 48.7\% in the Stage-2 population and 15.8\% in the full cohort. Zero-count overcounting is object-existence hallucination \citep{li2023pope}: the model asserts objects that are absent. Selectivity is essential: answering zero everywhere scores 307/630 on the zero-enriched Stage-2 population and 1,264/7,992 (15.8\%) over the cohort, and a development-calibrated confidence threshold at a matched firing rate yields 198/630 (Appendix~\ref{app:add_output}). The signal the gate isolates is present in the model's output: a development-calibrated threshold fixed before evaluation on $p(0\mid x)$ matches the learned gate on the Stage-2 branch (251 versus 244; $p=0.28$) and, applied to all inputs, reaches 5,317/7,992 ($+1.38\pp$ over identity), or 5,326/7,992 ($+1.49\pp$) when it emits zero (Appendix~\ref{app:add_gate}). Reading the zero-count signal from $p(0\mid x)$ thus extends deployment beyond the Stage-2 branch, which holds 177 of the cohort's 638 zero-count overcounts. The contribution of iterative control itself is isolated by the local-response, oracle-target, matched-budget, and free-generation experiments (\S\ref{sec:res_mea}; Appendix~\ref{app:mea}).

\section{Discussion}
\label{sec:discussion}

\paragraph{Three gaps and their remedies.}
The experiments replace ``find a causal component and steer it'' with a sharper picture in which localization, actuation, reachability, and utility are distinct properties, each established by its own held-out experiment. Sensitivity concentrates in a subspace that carries little activation energy, and equal-norm calibration makes that subspace the most displacement-efficient control geometry tested. Projected gradients predict infinitesimal movement exactly, and iterative re-linearization extends this local law to high target reachability that also changes generated answers. Reachability amplifies the target rule it is given, and a fixed runner-up rule with a gate that isolates zero-count overcounting yields prospectively confirmed utility from the same actuator. \method{} reports these as three endpoints, mechanistic reachability, policy-level deployment utility, and population-level gain, because each answers a different question about control.

\paragraph{Generalization.}
Each component generalizes at its own level. Actuator behavior replicates across Llama heads, the low-rank hierarchy reappears independently in IDEFICS with finite gain calibrated per model, and \mea{}'s advantage over a fixed direction replicates there on newly generated data. Llama's geometry transfers to natural images without refitting, and the routing policy transfers after recalibrating its utility map on 500 examples. On HoloCount, the transferred policy leaves accuracy unchanged while the action family retains $+12.26\pp$ of oracle headroom (Appendix~\ref{app:natural}), a concrete target for richer readouts. Off-target behavior is preserved: across matched GQA conditions, WHOLE changes 1.44\% of baseline-correct answers and MN 1.19\% (Appendix~\ref{app:offtarget}). These results are interpretable component by component because discovery, development, and confirmation used disjoint data (Appendix~\ref{app:discussion}).

\paragraph{Target specification.}
On the active branch, assigned-target reachability is 96.83\% and target accuracy 80.95\%, so every improvement in target construction translates into accuracy. The ground-truth-in-top-2 bound on the gated set is 248/630, and gates that read $p(0\mid x)$ beyond the Stage-2 branch already reach zero-count overcounts that the learned gate does not see. Combining such gates with \mea{}, which realizes any selected target in generation, is a direct route to larger population-level gains with the same actuator.

\section{Conclusion}
\label{sec:conclusion}

Localizing a behavior is the beginning of control. We presented \method{}, a mechanism-to-control framework for frozen VLMs that identifies three gaps between a causally validated component and a useful controller and closes each with a remedy confirmed on held-out data. A sensitivity--energy mismatch predicts how far naive low-rank edits fall short, and equal-norm calibration delivers whole-head control from four of 128 dimensions with no gain search. Iterative re-linearization extends a locally exact response law to high target reachability and changes the answers the model generates, in two architectures. A learned gate with a ground-truth-free runner-up target turns this capability into a prospectively confirmed gain on new natural images while editing fewer than 1\% of inputs, and shows that object-existence hallucination can be read directly from the model's numeral distribution. Interpretability becomes operational when causal relevance, intervention geometry, finite calibration, target specification, and deployment policy are each validated as a stage of control.

\section*{Limitations}
\method{} is developed on visual counting, whose discrete outcome space makes every edit auditable as a repair, a break, or no change. The deployment pipeline runs on Llama \headname{}; actuator behavior replicates across six Llama heads, and low-rank geometry and oracle \mea{} reachability replicate in IDEFICS, whose gated cross-attention blocks provide an independent architecture. In IDEFICS, the closed-set scoring proxy and free-form answers often disagree (22\% agreement on development data), so reachability and generated-answer correctness are reported as separate readouts, and the trust-region scale was selected at the edge of the development grid. The deployment gain inherits the scope of the Stage-2 branch, which contains 177 of the cohort's 638 zero-count overcounts, and oracle \mea{} runs in float32 whereas natural-image \mea{} runs in BF16. Three directions follow from these results. First, extending the actuator search to VLMs that fuse image and text through shared self-attention tests the three gaps beyond cross-attention fusion. Second, target construction is the highest-leverage component: the runner-up rule reaches 80.95\% target accuracy on the active branch, the ground-truth-in-top-2 bound is 248/630, and gates that read $p(0\mid x)$ on all inputs can address the 461 zero-count overcounts outside the Stage-2 branch. Third, \mea{}'s state-dependent re-linearization applies to any discrete target space, and extending it to self-attention architectures, numerical precisions, and non-counting behaviors is a natural next step.

\section*{Ethical Considerations}
Inference-time intervention can repair specific behaviors without updating model weights, and the same capability can introduce regressions when deployed outside validated conditions. Our results therefore support reporting repairs and breaks, abstention behavior, intervention magnitude, off-target break burden, finite-step diagnostics, transfer results, and confirmation after calibration alongside accuracy gains. The GQA stress test provides a prospective collateral-behavior check outside the counting task. Target-directed control is also dual-use: under the upward-only rule, \mea{} reached 66.93\% of assigned targets even when they were wrong, and edited states change generated answers, so a party with weight access could steer an automated counting system toward chosen outputs. Monitoring activation displacement at validated actuators is a natural detection signal for such interventions.

\FloatBarrier
\ifaclstyle
\small
\else
\nobalance
\small
\setlength{\bibsep}{1.5pt plus .3pt minus .3pt}
\afterpage{\balance}
\fi

\clearpage
\nobalance
\appendix
\ifaclstyle\normalsize\else\onecolumn\fontsize{10}{12}\selectfont\fi

\Needspace{5\baselineskip}
\section{Appendix Roadmap, Evidence Levels, and Timeline}
\label{app:timeline}
This appendix gives the details behind the main text, grouped by the three gaps. Appendix~\ref{app:timeline} states the evidence levels, the experimental timeline, and the analyses added after confirmation. \emph{Gap~1:} Appendix~\ref{app:mechanism} records mechanism discovery, the sensitivity geometry, specificity controls, the held-out v4 confirmation, and equal-displacement geometry baselines in Llama; Appendices~\ref{app:multi_actuator} and~\ref{app:idefics} give the multi-actuator and IDEFICS replications. \emph{Gap~2:} Appendix~\ref{app:mea} covers the local response law, oracle \mea{}, matched-budget controls, free-form generation under \mea{} edits, and \mea{} displacement and compute cost. \emph{Gap~3:} Appendix~\ref{app:routing} covers routing over direction, gain, and geometry; Appendix~\ref{app:natural} natural-image transfer, hierarchical natural \ugr{}, and HoloCount; and Appendix~\ref{app:target} the complete targeted-deployment chain, full-population accounting, external steering and output-space controls, and gate analyses. Appendix~\ref{app:offtarget} reports the off-target stress test and finite-step diagnostics, Appendix~\ref{app:discussion} an extended discussion, the scope of claims, and positioning against related work, and Appendix~\ref{app:reproducibility} reproducibility details.

\ifaclstyle\begin{table*}[!tp]\else\begin{table}[!t]\fi
\centering
\fontsize{8.5}{10.5}\selectfont
\caption{\textbf{Headline evidence across the AROMA control stack.} The table separates mechanism/controller validation from held-out prospective confirmation. The final targeted-deployment row uses a new natural cohort evaluated only after target, gate, reference, and MEA execution semantics are fixed.}
\label{tab:headline}
\setlength{\tabcolsep}{3.5pt}
\begin{tabularx}{\linewidth}{@{}P{3.05cm}r P{4.75cm}Y P{2.55cm}@{}}
\toprule
\bfseries Experiment & \bfseries $N$ & \bfseries Primary result & \bfseries Interpretation & \bfseries Evidence level \\
\midrule
Llama directional characterization & 300 & 91.33\% down; 90.67\% up; 88.67\% strict & Primary bidirectional causal actuator & Held-out characterization \\
Llama Top-4 geometry & 300 / 653 & 89.82\% sensitivity ($N=300$); 9.52\% median projected energy ($N=653$ audit) & Sensitivity--energy mismatch & Mechanism validation \\
Llama held-out v4 & 2000 & 50.25\% base; 58.25\% MN-CSA; $+8.00\pp$ & Same-policy low-rank utility & Prospective held-out \\
Equal-displacement geometry baselines & 2000 & MN-CSA 58.25\% vs mean-difference 55.45\% ($+2.80\pp$) and PCA-4 52.75\% ($+5.50\pp$) & Sensitivity subspace is the most displacement-efficient geometry & Post hoc (protocol fixed) \\
Continuous local response & 160 & at $\beta=.05$: Pearson $r=.9936$, Spearman $=.9990$, sign 100\% & First-order projected control is locally accurate & Mechanism validation \\
Oracle MEA reachability & 112 & 81.25\% target success; fixed direction 15.18\%; activation-aligned 0.89\% & Re-linearization converts local geometry into global reachability & Controller validation \\
MEA free-form generation & 112 / 63 & Edited greedy answers correct 80/112 vs 0/112 (oracle); 54/63 vs 7/63 (natural, gated) & Edits change generated answers, not only proxy scores & Post hoc (protocol fixed) \\
IDEFICS oracle MEA & 470 & 53.2\% target success; fixed direction 2.8\%; generated answers correct 218 vs 124 & Re-linearization replicates in a second architecture & Held-out (new cohort, evaluated once) \\
Whole-head MEA (Llama) & 112 & 92/112 whole head vs 91/112 Top-4 ($p=1.0$) & Four dimensions carry iterative controllability & Ablation (protocol fixed) \\
Utility Router held-out final & 294 & 60.54\% global MN $\rightarrow$ 66.33\%; $+5.782\pp$ & Low-rank candidate routing & Prospective held-out \\
Procedural UGR held-out final & 633 & 70.93\% restricted $\rightarrow$ 73.14\%; $+2.212\pp$ & Geometry routing & Prospective held-out \\
Natural hierarchical UGR final & 631 & 33.439\% Whole $\rightarrow$ 35.341\% Unified; $+1.902\pp$ & Natural geometry routing & Prospective held-out \\
Prospective runner-up + MEA & 630 & 34.444\% reference \ugr{} $\rightarrow$ 38.730\% ($+4.286\pp$; 29:2; $p=4.63\times10^{-7}$) & GT-free target + fixed gate converts reachability into utility & Prospective held-out \\
Full-cohort identity fallback & 7992 & 5207/7992 identity $\rightarrow$ 5254/7992 gated MEA ($+0.588\pp$) & Audited whole-population selective utility, derived from fixed outputs & Post hoc reconstruction \\
Gate selectivity & 7992 & Gate precision for zero-count 81\% vs 15.8\% base; global emit-zero 1264/7992 & Gate isolates zero-count overcounting & Post hoc (stored outputs) \\
External GrAInS-inspired comparison & 630 & Same-gate 225/630 vs MEA 244/630; global GrAInS-inspired 247/630 & Policy allocation and deployment scope are distinct & Post hoc supplemental \\
Llama multi-actuator robustness & 300 & 6/7 heads replicate bidirectionally; 5/6 non-primary & Within-model actuator robustness & Prospective robustness \\
IDEFICS actuator / geometry & 300 & 82.00\% down; 82.67\% up; calibrated rank-4 retains 81.09\%/81.49\% effect & Independent cross-model hierarchy & Held-out replication \\
GQA off-target stress test & 1000 & WHOLE 1.443\% vs MN 1.186\% break burden; MN lower in 6/7 budgets & Low collateral burden for both geometries & Prospective stress test \\
\bottomrule
\end{tabularx}
\ifaclstyle\end{table*}\else\end{table}\fi

\subsection{Experimental Timeline and Statistics}
\label{app:timeline_details}
\paragraph{Data timeline.}
For Llama, \proc{} v1 ($N=300$) supports mechanism discovery, v2 ($N=1000$) original router development, v3 ($N=2000$) fixed whole-head confirmation and later CSA development, and v4 ($N=2000$) is generated only after \mncsa{} is closed; v4 has zero seed/pixel-hash overlap with v1--v3. Natural evaluation uses two disjoint 4,000-example TallyQA samples for transfer and Stage-1 recalibration, a later held-out 8,000-example TallyQA cohort for natural hierarchical UGR, and HoloCount ($N=2480$) as an external boundary. The 8,000-example Natural UGR cohort is balanced 4,000 simple/4,000 complex and has zero question, image-ID, and pixel overlap with the earlier natural data. IDEFICS uses mutually disjoint evaluation cohorts: held-out actuator confirmation ($N=300$), geometry estimation ($N=300$), equal-norm \mncsa{} ($N=300$), screen ($N=100$), dev validation ($N=200$), and final held-out confirmation ($N=300$); a further $N=500$ reserve remains unused.

\paragraph{Unified-geometry routing timeline.}
UGR development uses a new 2,000-example procedural cohort generated under the same renderer/distribution rules but with zero overlap against the earlier procedural cohorts.
The Stage-1 Proposal Controller invokes Stage~2 on 654 examples.
Grouped OOF development yields 68.81\% for both restricted routers; the whole-head restricted router is selected prospectively because it has fewer breaks (34 versus 37).
The unified router reaches 71.87\% OOF.
Only after the complete UGR protocol, feature map, candidate grid, trained routers, comparator, and statistics are fixed do we generate a separate held-out 2,000-example final cohort.
Its realized Stage-2 population contains 633 examples ($\alpha_{\rm prop}>1$), and no router retraining or candidate-family modification is permitted after final outcomes are observed.

\paragraph{Natural hierarchical UGR timeline.}
Natural Stage-2 development uses the $N=292$ upward proposals produced by the natural Proposal Controller from the earlier TallyQA development population. A flat unified candidate classifier underperforms Whole-only routing (35.959\% versus 38.014\%), and pairwise ranking alone remains slightly below Whole (36.301\% versus 36.986\%). These are development results. The selected hierarchical design separates IDENTITY-vs-intervene gating from non-identity pairwise ranking and reaches 35.616\% versus 34.247\% for the matched hierarchical Whole comparator ($+1.370\pp$), with 16 versus 12 repairs and zero breaks for both. The complete hierarchical policies, 47-D feature construction, 43-candidate action space, gate threshold, and model hyperparameters are then fixed before a new 8,000-example TallyQA final cohort is evaluated. Stage~1 produces 631 upward proposals; all 43 candidates are exhaustively executed for each, yielding 27,133 candidate outcomes. No final outcome is used to refit either hierarchical router.

\paragraph{Prospective targeted-MEA timeline.}
The runner-up target rule is proposed only after diagnosing the failure of the upward-only rule and is first supported on natural development OOF. Its initial comparison on the earlier $N=631$ population is reported as retrospective because that population had already been inspected. The rule is then fixed without modification before a new prospective natural cohort is processed. Candidate selection and pixel deduplication are outcome-independent and completed before model inference; the final cohort contains 7,992 examples (3,996 simple/3,996 complex), zero historical question/image-ID/pixel overlap, and zero within-cohort pixel duplicates. Stage~1 yields $N=630$ upward Stage-2 examples. The hierarchical gate, re-executed and checked against its stored outputs, produces 63 prospective gated decisions. The 630 gate decisions, runner-up targets, and reference-\ugr{} method/gain actions are all fixed before the 63 reference actions and 63 MEA trajectories are executed. No 43-candidate prospective sweep, router refit, target retuning, or development update is performed.

\paragraph{Statistics and evidence levels.}
We report exact McNemar tests for paired correctness, exact directional binomial tests for sign consistency, and paired bootstrap intervals for mean shifts, retention ratios, routing accuracy differences, and off-target break burden. In the multi-actuator study, attenuation and amplification are tested separately against 0.5 directional consistency and Holm correction is applied across the 14 head-by-direction tests; a head is counted as bidirectionally replicated only when both corrected directional tests pass. The low-rank Utility Router final primary test is two-sided exact McNemar on the $N=294$ upward population, with a 20,000-repetition paired bootstrap interval. The procedural UGR final primary test compares fixed unified versus prospectively selected whole-head restricted routing on the realized held-out $N=633$ Stage-2 population using two-sided exact McNemar and a 20,000-repetition paired bootstrap interval. The natural hierarchical UGR primary test analogously compares the hierarchical Unified and Whole policies on the prespecified held-out $N=631$ upward Stage-2 population using two-sided exact McNemar and a paired bootstrap interval. The prospective targeted-MEA primary test compares the reference \ugr{} against runner-up + MEA on the identical fixed $N=630$ Stage-2 population using two-sided exact McNemar; the observed discordance is 29 MEA-only versus 2 reference-only. The GQA stress test reports the paired mean difference in break burden across seven budgets with a paired bootstrap interval; later robustness includes seed-stability checks and a 1,000,000-draw paired sign-flip randomization test.

\subsection{Analyses Added After Confirmation}
\label{app:additional}
Five analyses were carried out after the prospective results were final: equal-displacement geometry baselines (Appendix~\ref{app:add_geometry}), free-form generation under \mea{} edits (Appendix~\ref{app:add_freegen}), output-space and gate controls (Appendix~\ref{app:add_output}), gate recall and non-learned gates (Appendix~\ref{app:add_gate}), and \mea{} displacement and compute cost (Appendix~\ref{app:add_cost}). Two further analyses replicate oracle \mea{} in \idefics{} on a newly generated cohort (Appendix~\ref{app:idefics_mea}) and ablate the Top-4 restriction of \mea{} in Llama (Appendix~\ref{app:wholehead_mea}); each has its own protocol, fixed before its test evaluation. The first five analyses reuse fixed cohorts and protocols; directions, subspaces, and gains are fit or selected on development data only, and each test configuration is fixed before evaluation and evaluated once. Before any new comparison, the original protocol is re-executed: on the v3 development intervention subset, baseline (175/653), \mncsa{} (343/653), and naive projection (217/653) match the original per-example outputs exactly, as do baseline (163/604) and \mncsa{} (323/604) on the edited v4 rows; re-executed \mea{} reproduces all 175 final predictions. The geometry, generation, and cost analyses use the same GPU class as the v4 confirmation, and the five analyses together use 11.2 GPU-hours; the \idefics{} replication and the ablation use a further 7.9. Existing result files were checksummed beforehand and verified unchanged afterwards.

\FloatBarrier
\Needspace{6\baselineskip}
\section{Mechanism Discovery and Low-Rank Geometry in Llama}
\label{app:mechanism}

\subsection{Actuator Discovery and the Stage-1 Action Space}
The final method focuses on \headname{}, but the discovery process provides evidence for adaptive rather than fixed mechanistic control.
Exhaustive screening across 256 cross-attention heads produced seven candidates: L33H1, L3H4, L18H13, L8H30, L3H11, L13H11, and L33H21.
Held-out candidate-vs-random validation gave mean interaction $-0.2859$ with bootstrap 95\% CI $[-0.4533,-0.1089]$ under the experiment's sign convention.
A per-sample best-head oracle over 27 held-out pairs gave mean correct-answer log-probability improvement $+0.0932$ (95\% CI $[+0.0613,+0.1269]$) on baseline-wrong examples versus $+0.0126$ on baseline-correct examples, for specificity $+0.0806$ (95\% CI $[+0.0457,+0.1163]$).
Every candidate head wins on at least one sample.
A multi-head equalization attempt was not robust because nonlinear interactions could cancel, motivating one stable actuator plus per-example routing.
In the 300-sample characterization of \headname{} (Fig.~\ref{fig:steering}), attenuation lowers $\mu$ on $274/300$ examples and amplification raises it on $272/300$.

\ifaclstyle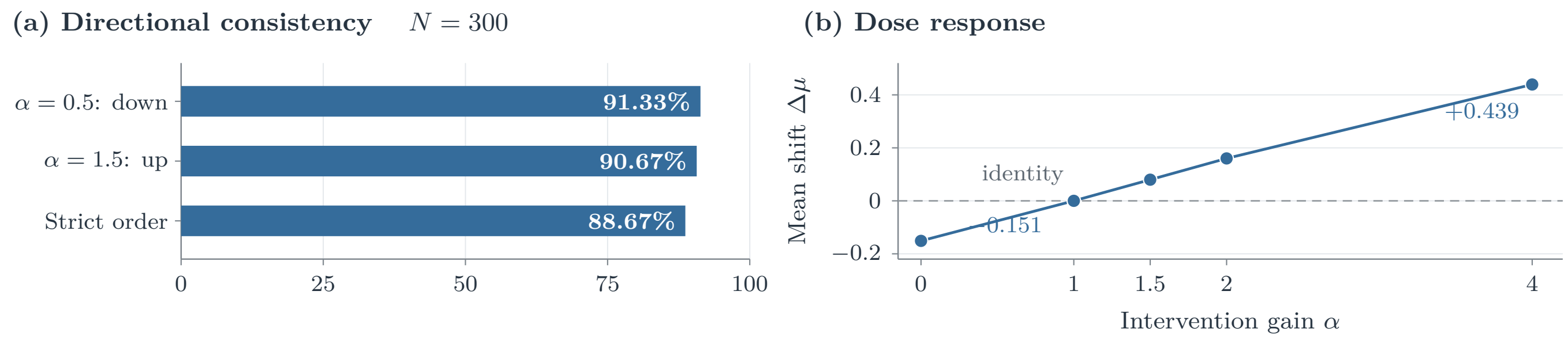
\begin{figure*}[!tp]\else\begin{figure}[!htbp]\fi
\centering

\pgfmathsetlengthmacro{\steerPlotWidth}{.5*\linewidth-2.55cm}

\begin{tikzpicture}

\pgfplotsset{
    steer axis/.style={
        scale only axis,
        width=\steerPlotWidth,
        axis background/.style={fill=white},
        axis x line*=bottom,
        axis y line*=left,
        axis line style={
            draw=aromaInk!60,
            line width=0.45pt
        },
        tick align=outside,
        tick style={
            draw=aromaInk!60,
            line width=0.4pt
        },
        major tick length=2pt,
        tick label style={
            font=\rmfamily\fontsize{7.5}{9}\selectfont,
            text=aromaInk
        },
        label style={
            font=\rmfamily\fontsize{8}{9.5}\selectfont,
            text=aromaInk
        },
        grid style={
            draw=aromaGrid,
            line width=0.35pt
        },
        scaled ticks=false,
        clip=true
    }
}

\node[
    anchor=west,
    inner sep=0pt,
    text=aromaInk,
    font=\rmfamily\fontsize{8.5}{10.5}\selectfont
] at (0,0) {
    \textbf{(a) Directional consistency}
    \quad $N=300$
};

\begin{axis}[
    steer axis,
    at={(1.95cm,-0.45cm)},
    anchor=north west,
    height=2.25cm,
    xbar,
    xmin=0,
    xmax=100,
    symbolic y coords={
        Strict order,
        $\alpha=1.5$: up,
        $\alpha=0.5$: down
    },
    ytick=data,
    yticklabel style={
        font=\rmfamily\fontsize{7.5}{9}\selectfont,
        text=aromaInk,
        inner sep=2pt
    },
    xtick={0,25,50,75,100},
    xticklabels={0,25,50,75,100},
    xmajorgrids=true,
    ymajorgrids=false,
    enlarge y limits=0.32,
    bar width=10pt
]

\addplot[
    draw=none,
    fill=aromaBlue,
    nodes near coords,
    point meta=explicit symbolic,
    every node near coord/.append style={
        font=\rmfamily\fontsize{7.5}{9}\selectfont\bfseries,
        text=white,
        anchor=east,
        xshift=-3pt,
        inner sep=0pt
    }
]
coordinates {
    (91.33,{$\alpha=0.5$: down}) [91.33\%]
    (90.67,{$\alpha=1.5$: up})   [90.67\%]
    (88.67,{Strict order})      [88.67\%]
};

\end{axis}

\node[
    anchor=west,
    inner sep=0pt,
    text=aromaInk,
    font=\rmfamily\fontsize{8.5}{10.5}\selectfont
] at (.5\linewidth,0) {
    \textbf{(b) Dose response}
};

\begin{axis}[
    steer axis,
    at={(.5\linewidth+1.10cm,-.45cm)},width=\dimexpr.5\linewidth-1.45cm\relax,
    anchor=north west,
    height=2.25cm,
    xmin=-0.15,
    xmax=4.20,
    ymin=-0.22,
    ymax=0.52,
    xlabel={Intervention gain $\alpha$},
    ylabel={Mean shift $\Delta\mu$},
    xtick={0,1,1.5,2,4},
    xticklabels={0,1,1.5,2,4},
    ytick={-0.2,0,0.2,0.4},
    yticklabels={$-0.2$,$0$,$0.2$,$0.4$},
    xmajorgrids=false,
    ymajorgrids=true
]

\addplot[
    draw=aromaInk!50,
    densely dashed,
    line width=0.5pt,
    forget plot
]
coordinates {
    (-0.15,0)
    (4.20,0)
};

\addplot[
    color=aromaBlue,
    line width=0.9pt,
    mark=*,
    mark size=2.2pt,
    mark options={
        fill=aromaBlue,
        draw=white,
        line width=0.4pt
    }
]
coordinates {
    (0,   -0.151121)
    (1,    0.000000)
    (1.5,  0.079846)
    (2,    0.160119)
    (4,    0.439136)
};

\node[
    font=\rmfamily\fontsize{7.5}{9}\selectfont,
    text=aromaInk!75,
    anchor=south east,
    xshift=-3pt,
    yshift=5pt,
    inner sep=0pt
] at (axis cs:1,0) {identity};

\node[
    font=\rmfamily\fontsize{7.5}{9}\selectfont\bfseries,
    text=aromaBlue,
    anchor=north,
    xshift=0pt,
    yshift=0pt,
    inner sep=0pt
] at (axis cs:.55,-0.06) {$-0.151$};

\node[
    font=\rmfamily\fontsize{7.5}{9}\selectfont\bfseries,
    text=aromaBlue,
    anchor=north east,
    xshift=-4pt,
    yshift=-6pt,
    inner sep=0pt
] at (axis cs:4,0.439136) {$+0.439$};

\end{axis}

\end{tikzpicture}

\caption{\textbf{Bidirectional cardinality steering.}
\textbf{(a)} In the 300-sample characterization, $\alpha=0.5$ shifts the normalized expected numeral downward on 91.33\% of samples, and $\alpha=1.5$ shifts it upward on 90.67\%. Strict ordering $\mu_{0.5}<\mu_1<\mu_{1.5}$ holds on 88.67\%; downward and upward response magnitudes are strongly coupled (Spearman $\rho=0.9514$, $p=1.59\times10^{-154}$).
\textbf{(b)} On \proc{} v2, gains $\alpha\in\{0,1,1.5,2,4\}$ produce monotonic mean expected-numeral shifts, from $-0.151$ at $\alpha=0$ to $+0.439$ at $\alpha=4$; $\alpha=1$ exactly reproduces baseline. These results support a bidirectional, dose-responsive cardinality actuator.}
\label{fig:steering}
\ifaclstyle\end{figure*}\else\end{figure}\fi

\ifaclstyle\begin{table*}[!tp]\else\begin{table}[H]\fi
\centering\fontsize{8.5}{10.5}\selectfont
\caption{Full v2 gain sweep; repairs are among the 497 baseline-wrong examples.}
\label{tab:fullactions}
\begin{tabular}{@{}lrrrrrrrrrr@{}}
\toprule
\bfseries $\alpha$ & \bfseries 0 & \bfseries .25 & \bfseries .5 & \bfseries .75 & \bfseries 1 & \bfseries 1.25 & \bfseries 1.5 & \bfseries 2 & \bfseries 3 & \bfseries 4 \\
\midrule
Repairs & 20 & 16 & 13 & 8 & 0 & 18 & 35 & 60 & 80 & 89 \\
\bottomrule
\end{tabular}
\ifaclstyle\end{table*}\else\end{table}\fi
Under grouped out-of-fold evaluation on v2 ($N=1000$), the Proposal Controller raises accuracy from 50.30\% to 58.60\% ($+8.30\pp$). The full-action oracle repairs 139/497 errors and reaches 64.2\% ($+13.9\pp$).
The compressed set $\{0,1,1.5,2,4\}$ repairs 138/497 errors and reaches 64.1\% ($+13.8\pp$), retaining essentially all oracle headroom with a smaller deployable action space.

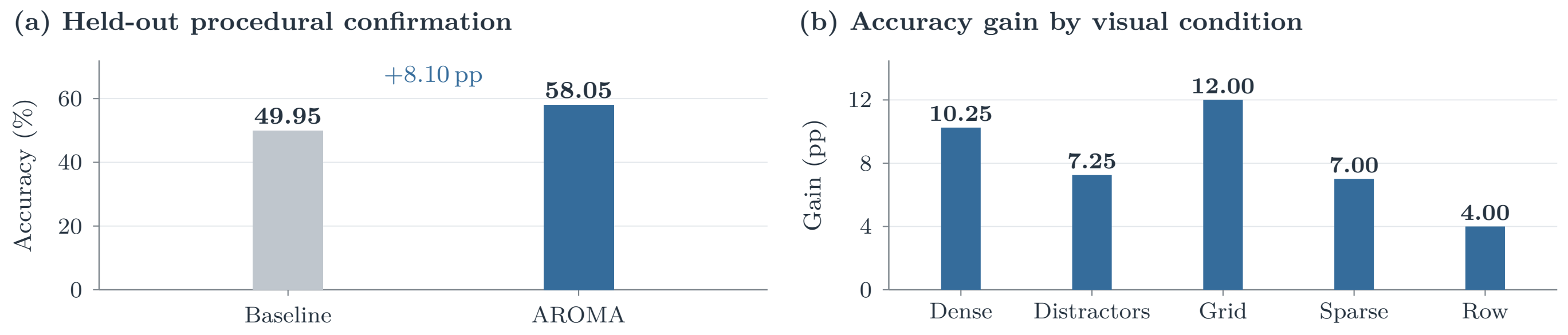
\ifaclstyle\begin{figure*}[!tp]\else\begin{figure}[!htbp]\fi
\centering

\pgfmathsetlengthmacro{\procPlotWidth}{.5*\linewidth-1.35cm}

\begin{tikzpicture}

\pgfplotsset{
    proc axis/.style={
        scale only axis,
        width=\procPlotWidth,
        axis background/.style={fill=white},
        axis x line*=bottom,
        axis y line*=left,
        axis line style={
            draw=aromaInk!60,
            line width=0.45pt
        },
        tick align=outside,
        major tick length=2pt,
        tick style={
            draw=aromaInk!60,
            line width=0.4pt
        },
        tick label style={
            font=\rmfamily\fontsize{7.5}{9}\selectfont,
            text=aromaInk
        },
        label style={
            font=\rmfamily\fontsize{8}{9.5}\selectfont,
            text=aromaInk
        },
        ymajorgrids=true,
        xmajorgrids=false,
        grid style={
            draw=aromaGrid,
            line width=0.35pt
        },
        scaled ticks=false,
        clip=true
    }
}

\node[
    anchor=west,
    inner sep=0pt,
    text=aromaInk,
    font=\rmfamily\fontsize{8.5}{10.5}\selectfont
] at (0,0) {
    \textbf{(a) Held-out procedural confirmation}
};

\begin{axis}[
    proc axis,
    at={(1.00cm,-0.43cm)},
    anchor=north west,
    height=2.65cm,
    ybar,
    bar width=23pt,
    bar shift=0pt,
    xmin=-0.65,
    xmax=1.65,
    ymin=0,
    ymax=72,
    ylabel={Accuracy (\%)},
    xtick={0,1},
    xticklabels={Baseline,AROMA},
    ytick={0,20,40,60},
    nodes near coords,
    point meta=explicit symbolic,
    every node near coord/.append style={
        font=\rmfamily\fontsize{8}{9.5}\selectfont\bfseries,
        text=aromaInk,
        anchor=south,
        yshift=2pt,
        inner sep=0pt
    }
,legend image code/.code={\path[#1] (0pt,-2pt) rectangle (8pt,3pt);}]

\addplot[
    fill=aromaGrayLight,
    draw=none
]
coordinates {
    (0,49.95) [49.95]
};

\addplot[
    fill=aromaBlue,
    draw=none
]
coordinates {
    (1,58.05) [58.05]
};

\node[
    font=\rmfamily\fontsize{8}{9.5}\selectfont\bfseries,
    text=aromaBlue,
    anchor=center,
    inner sep=0pt
] at (axis cs:0.5,67) {$+8.10\,\mathrm{pp}$};

\end{axis}

\node[
    anchor=west,
    inner sep=0pt,
    text=aromaInk,
    font=\rmfamily\fontsize{8.5}{10.5}\selectfont
] at (.5\linewidth,0) {
    \textbf{(b) Accuracy gain by visual condition}
};

\begin{axis}[
    proc axis,
    at={(.5\linewidth+1.05cm,-.43cm)},
    anchor=north west,
    height=2.65cm,
    ybar,
    bar width=13pt,
    bar shift=0pt,
    xmin=-0.55,
    xmax=4.55,
    ymin=0,
    ymax=14.5,
    ylabel={Gain (pp)},
    xtick={0,1,2,3,4},
    xticklabels={Dense,Distractors,Grid,Sparse,Row},
    xticklabel style={
        font=\rmfamily\fontsize{7.5}{9}\selectfont,
        text=aromaInk,
        inner sep=2pt
    },
    ytick={0,4,8,12},
    nodes near coords,
    point meta=explicit symbolic,
    every node near coord/.append style={
        font=\rmfamily\fontsize{7.5}{9}\selectfont\bfseries,
        text=aromaInk,
        anchor=south,
        yshift=2pt,
        inner sep=0pt
    }
,legend image code/.code={\path[#1] (0pt,-2pt) rectangle (8pt,3pt);}]

\addplot[
    fill=aromaBlue,
    draw=none
]
coordinates {
    (0,10.25) [10.25]
    (1, 7.25) [7.25]
    (2,12.00) [12.00]
    (3, 7.00) [7.00]
    (4, 4.00) [4.00]
};

\end{axis}
\end{tikzpicture}

\par\vspace{4pt}

{%
\rmfamily\fontsize{8}{9.5}\selectfont
\setlength{\tabcolsep}{0pt}
\renewcommand{\arraystretch}{1.18}
\begin{tabular*}{\linewidth}{
    @{\extracolsep{\fill}}cccc@{}
}
\toprule
\textbf{Repairs} &
\textbf{Breaks} &
\textbf{Net repairs} &
\textbf{Intervention rate} \\
\midrule
173 & 11 & \textbf{+162} & 32.65\% \\
\bottomrule
\end{tabular*}
\par
}

\caption{\textbf{Procedural confirmation.}
The synthetic router is fixed before evaluation on the held-out 2,000-example \proc{} v3 set.
\textbf{(a)} Accuracy increases from $999/2000=49.95\%$ to $1161/2000=58.05\%$ ($+8.10\pp$; paired 95\% CI $[+6.85,+9.40]\pp$). There are 173 repairs, 11 breaks, and $+162$ net repairs, with interventions on 32.65\% of examples.
\textbf{(b)} All prespecified visual conditions improve: \texttt{dense}, $+10.25\pp$; \texttt{distractors}, $+7.25\pp$; \texttt{grid}, $+12.00\pp$; \texttt{random\_sparse}, $+7.00\pp$; and \texttt{row}, $+4.00\pp$. These gains support generalization beyond v2 within the procedural benchmark family.}
\label{fig:supp_procedural}
\ifaclstyle\end{figure*}\else\end{figure}\fi

\subsection{Sensitivity Spectrum and Null Statistics}
The rank-4 sensitivity experiment uses $N=300$, head dimension 128, rank $r=4$, and 10,000 norm-matched isotropic random-direction null replicates.
Both the uncentered and the centered Top-4 shares exceed the empirical 99th percentile and every null replicate.
\ifaclstyle\begin{table*}[!tp]\else\begin{table}[H]\fi
\centering\fontsize{8.5}{10.5}\selectfont
\caption{Explained cardinality-sensitivity energy by rank.}
\label{tab:csa_spectrum}
\setlength{\tabcolsep}{3.5pt}
\fontsize{8.5}{10.5}\selectfont
\begin{tabular*}{\linewidth}{@{\extracolsep{\fill}}lrrrrrrrr@{}}
\toprule
\bfseries $r$ & \bfseries 1 & \bfseries 2 & \bfseries 4 & \bfseries 8 & \bfseries 16 & \bfseries 32 & \bfseries 64 & \bfseries 128 \\
\midrule
\bfseries Total second moment & 0.762 & 0.842 & 0.898 & 0.945 & 0.977 & 0.993 & 0.999 & 1.00 \\
\bfseries Centered residual & 0.695 & 0.798 & 0.868 & 0.928 & 0.970 & 0.992 & 0.999 & 1.00 \\
\bottomrule
\end{tabular*}
\ifaclstyle\end{table*}\else\end{table}\fi
Figure~\ref{fig:csa_geometry} contrasts sensitivity concentration with projected activation norm.
The mean-gradient direction accounts for 0.233 of total trace.
For $r=4$, the total null median/q95/q99 are 0.211/0.217/0.219, yielding 4.249$\times$ enrichment and empirical $p=9.999\times10^{-5}$; centered null median/q95/q99 are 0.211/0.217/0.219, yielding 4.103$\times$ enrichment and the same empirical $p$.
Both enrichments are reported against a prespecified auxiliary $5\times$ enrichment threshold, and CSA was accordingly carried through development and confirmed on the independent v4 cohort (Appendix~\ref{app:v4}).

\subsection{Actuation-Strength Audit and Specificity Controls}
For the 653 fixed non-\noop{} v3 examples, $\rho=\|P_Uh\|/\|h\|$ has quantiles 0/10/25/50/75/90/100\% of 0.0543, 0.2326, 0.2805, 0.3086, 0.3393, 0.3790, and 0.5102.
Mean $\rho$ is 0.3082 and median projected energy share $\rho^2$ is 0.0952.
By selected action, mean $\rho$ is 0.339 for $\alpha=0$, 0.442 for $\alpha=1.5$, 0.357 for $\alpha=2$, and 0.293 for $\alpha=4$.
This audit is label-free and diagnoses an intervention-strength mismatch rather than selecting examples by outcome.

\ifaclstyle\begin{table*}[!tp]\else\begin{table}[H]\fi
\centering
\begin{minipage}[t]{0.48\linewidth}\vspace{0pt}
\captionsetup{justification=raggedright}
\centering\fontsize{8.5}{10.5}\selectfont
\caption{Magnitude-normalization recovery on the 653 non-\noop{} v3 intervention examples.}
\label{tab:csa_recovery}
\setlength{\tabcolsep}{3pt}\begin{tabularx}{\linewidth}{@{}Yrrrr@{}}
\toprule
\bfseries Method & \bfseries Accuracy & \bfseries Repairs & \bfseries Breaks & \bfseries Net \\
\midrule
Baseline & 26.799\% & -- & -- & -- \\
Whole-head & 50.995\% & 169 & 11 & 158 \\
Same-$\alpha$ CSA & 33.231\% & 49 & 7 & 42 \\
MN-CSA Top-4 & 52.527\% & 185 & 17 & 168 \\
\bottomrule
\end{tabularx}
\end{minipage}\hfill
\begin{minipage}[t]{0.49\linewidth}\vspace{0pt}
\captionsetup{justification=raggedright}
\centering\fontsize{8.5}{10.5}\selectfont
\caption{Rank-matched specificity controls on the same intervention subset.}
\label{tab:csa_specificity}
\setlength{\tabcolsep}{3pt}\begin{tabularx}{\linewidth}{@{}Yrrr@{}}
\toprule
\bfseries Comparison & \bfseries Accuracy & \shortstack[r]{\bfseries Difference\\\bfseries vs Top-4} & \bfseries 95\% CI \\
\midrule
Top-4 MN-CSA & 52.527\% & -- & -- \\
Bottom-4 MN-CSA & 26.799\% & $-25.727\pp$ & $[-29.556,-22.052]$ \\
Random-4 mean (20) & 27.175\% & $-25.352\pp$ & $[-28.936,-21.792]$ \\
Best random-4 & 30.781\% & $-21.746\pp$ & -- \\
\bottomrule
\end{tabularx}
\end{minipage}
\ifaclstyle\end{table*}\else\end{table}\fi
MN-CSA minus whole-head is $+1.531\pp$ with paired bootstrap 95\% CI $[-0.766,+3.828]\pp$ and McNemar $p=0.237$; MN-CSA minus same-$\alpha$ CSA is $+19.296\pp$ with 95\% CI $[+16.08,+22.665]\pp$ and McNemar $p=2.64\times10^{-28}$.

No one of the 20 fixed random rank-4 controls reaches Top-4; the empirical rank probability is $1/21=0.0476$, the minimum attainable with 20 controls.

\ifaclstyle\begin{figure*}[!tp]\else\begin{figure}[!htbp]\fi
\centering
\pgfmathsetlengthmacro{\cbHalf}{0.5*\textwidth}
\pgfmathsetlengthmacro{\cbWidth}{0.5*\textwidth-1.55cm}
\pgfmathsetlengthmacro{\cbRight}{\cbHalf+1.15cm}
\begin{tikzpicture}
\pgfplotsset{cbaxis/.style={scale only axis,width=\cbWidth,height=2.72cm,anchor=north west,
axis x line*=bottom,axis y line*=left,axis line style={draw=aromaInk!60,line width=.45pt},
tick style={draw=aromaInk!60,line width=.4pt},tick align=outside,major tick length=2pt,
tick label style={font=\rmfamily\fontsize{7.5}{9}\selectfont,text=aromaInk},label style={font=\rmfamily\fontsize{8}{9.5}\selectfont,text=aromaInk},
ymajorgrids=true,grid style={draw=aromaGrid,line width=.35pt},scaled ticks=false,clip=true}}
\tikzset{cbtitle/.style={anchor=west,inner sep=0pt,text=aromaInk,font=\rmfamily\fontsize{8.5}{10.5}\selectfont\bfseries}}

\node[cbtitle] at (0,0) {(a) Calibration recovers low-rank control};
\begin{axis}[cbaxis,at={(1.0cm,-.7cm)},ybar,bar width=17pt,xmin=-.6,xmax=4.6,ymin=0,ymax=60,
ylabel={Accuracy (\%)},bar shift=0pt,xtick={0,1,2,3,4},xticklabels={Baseline,Whole,Naive CSA,MN-CSA,Bottom-4},
xticklabel style={font=\rmfamily\fontsize{7.5}{9}\selectfont,rotate=18,anchor=north east},ytick={0,20,40,60},legend image code/.code={\path[#1] (0pt,-2pt) rectangle (8pt,3pt);}]
\addplot[draw=none,fill=aromaGray] coordinates {(0,26.799)};
\addplot[draw=none,fill=aromaBlue] coordinates {(1,50.995)};
\addplot[draw=none,fill=aromaOrange] coordinates {(2,33.231)};
\addplot[draw=none,fill=aromaGreen] coordinates {(3,52.527)};
\addplot[draw=none,fill=aromaPurple] coordinates {(4,26.799)};
\node[font=\rmfamily\fontsize{7.5}{9}\selectfont\bfseries,text=aromaGreen!70!black,anchor=south] at (axis cs:3,53.0) {52.53};
\end{axis}

\node[cbtitle] at (\cbHalf,0) {(b) Random rank-4 controls ($n=20$)};
\begin{axis}[cbaxis,at={(\cbRight,-.7cm)},xmin=-1,xmax=20,ymin=20,ymax=56,
xlabel={Random subspace index},ylabel={Accuracy (\%)},xtick={0,4,8,12,16,19},ytick={20,30,40,50}]
\addplot[only marks,mark=*,mark size=2.1pt,color=aromaGray] coordinates {
(0,26.340)(1,25.421)(2,30.781)(3,23.890)(4,28.943)(5,26.799)(6,29.556)(7,26.493)(8,26.034)(9,27.871)
(10,26.187)(11,26.953)(12,27.871)(13,26.493)(14,26.953)(15,27.565)(16,30.475)(17,24.962)(18,29.097)(19,24.809)};
\draw[aromaGreen,line width=1pt] (axis cs:-1,52.527)--(axis cs:20,52.527);
\draw[densely dashed,aromaPurple,line width=.7pt] (axis cs:-1,27.175)--(axis cs:20,27.175);
\node[font=\rmfamily\fontsize{7.5}{9}\selectfont\bfseries,text=aromaGreen!70!black,anchor=south east] at (axis cs:19.2,46.5) {Top-4 52.53\%};
\node[font=\rmfamily\fontsize{7.5}{9}\selectfont,text=aromaPurple,anchor=south east] at (axis cs:19.7,34.5) {random mean 27.18\%};
\end{axis}

\node[cbtitle] at (0,-4.65cm) {(c) Held-out v4 confirmation};
\begin{axis}[cbaxis,at={(1.0cm,-5.55cm)},ybar,bar width=28pt,xmin=-.6,xmax=2.6,ymin=0,ymax=68,
ylabel={Accuracy (\%)},bar shift=0pt,xtick={0,1,2},xticklabels={Baseline,Whole-head,MN-CSA},ytick={0,20,40,60},
nodes near coords,every node near coord/.append style={font=\rmfamily\fontsize{7.5}{9}\selectfont\bfseries,text=aromaInk,anchor=south,yshift=2pt},legend image code/.code={\path[#1] (0pt,-2pt) rectangle (8pt,3pt);}]
\addplot[draw=none,fill=aromaGray] coordinates {(0,50.25)};
\addplot[draw=none,fill=aromaBlue] coordinates {(1,57.25)};
\addplot[draw=none,fill=aromaGreen] coordinates {(2,58.25)};
\end{axis}

\node[cbtitle] at (\cbHalf,-4.65cm) {(d) v4 repair / break decomposition};
\begin{axis}[cbaxis,at={(\cbRight,-5.55cm)},ybar,bar width=17pt,xmin=-.6,xmax=1.6,ymin=0,ymax=195,
ylabel={Examples},xtick={0,1},xticklabels={Whole-head,MN-CSA},ytick={0,50,100,150,190},
legend style={font=\rmfamily\fontsize{7.5}{9}\selectfont,draw=none,fill=none,at={(.5,1.02)},anchor=south,legend columns=2,/tikz/every even column/.append style={column sep=5pt}},legend image code/.code={\path[#1] (0pt,-2pt) rectangle (8pt,3pt);}]
\addplot[draw=none,fill=aromaGreen,bar shift=-10pt,forget plot,nodes near coords,every node near coord/.append style={font=\rmfamily\fontsize{7.5}{9}\selectfont,text=aromaInk,yshift=1pt}] coordinates {(0,150)(1,174)};
\addplot[draw=none,fill=aromaOrange,bar shift=10pt,forget plot,nodes near coords,every node near coord/.append style={font=\rmfamily\fontsize{7.5}{9}\selectfont,text=aromaInk,yshift=1pt}] coordinates {(0,10)(1,14)};
\addlegendimage{area legend,fill=aromaGreen,draw=none}\addlegendentry{Repairs}
\addlegendimage{area legend,fill=aromaOrange,draw=none}\addlegendentry{Breaks}
\end{axis}
\end{tikzpicture}
\caption{\textbf{From low-rank sensitivity to independently confirmed control.}
\textbf{(a)} Same-$\alpha$ rank-4 CSA severely under-actuates; magnitude normalization restores whole-head accuracy on the v3 development intervention subset (52.53\% vs 50.995\%; $p=0.24$).
\textbf{(b)} The learned Top-4 subspace exceeds every one of 20 fixed random rank-4 controls; random mean is 27.18\% and random maximum is 30.78\%, versus 52.53\% for Top-4.
\textbf{(c)} After the CSA method is closed, a newly generated v4 confirmation set yields 50.25\% baseline, 57.25\% whole-head, and 58.25\% MN-CSA.
\textbf{(d)} MN-CSA obtains 174 repairs and 14 breaks versus 150 repairs and 10 breaks for whole-head, producing 20 additional net repairs.}
\label{fig:csa_behavior}
\ifaclstyle\end{figure*}
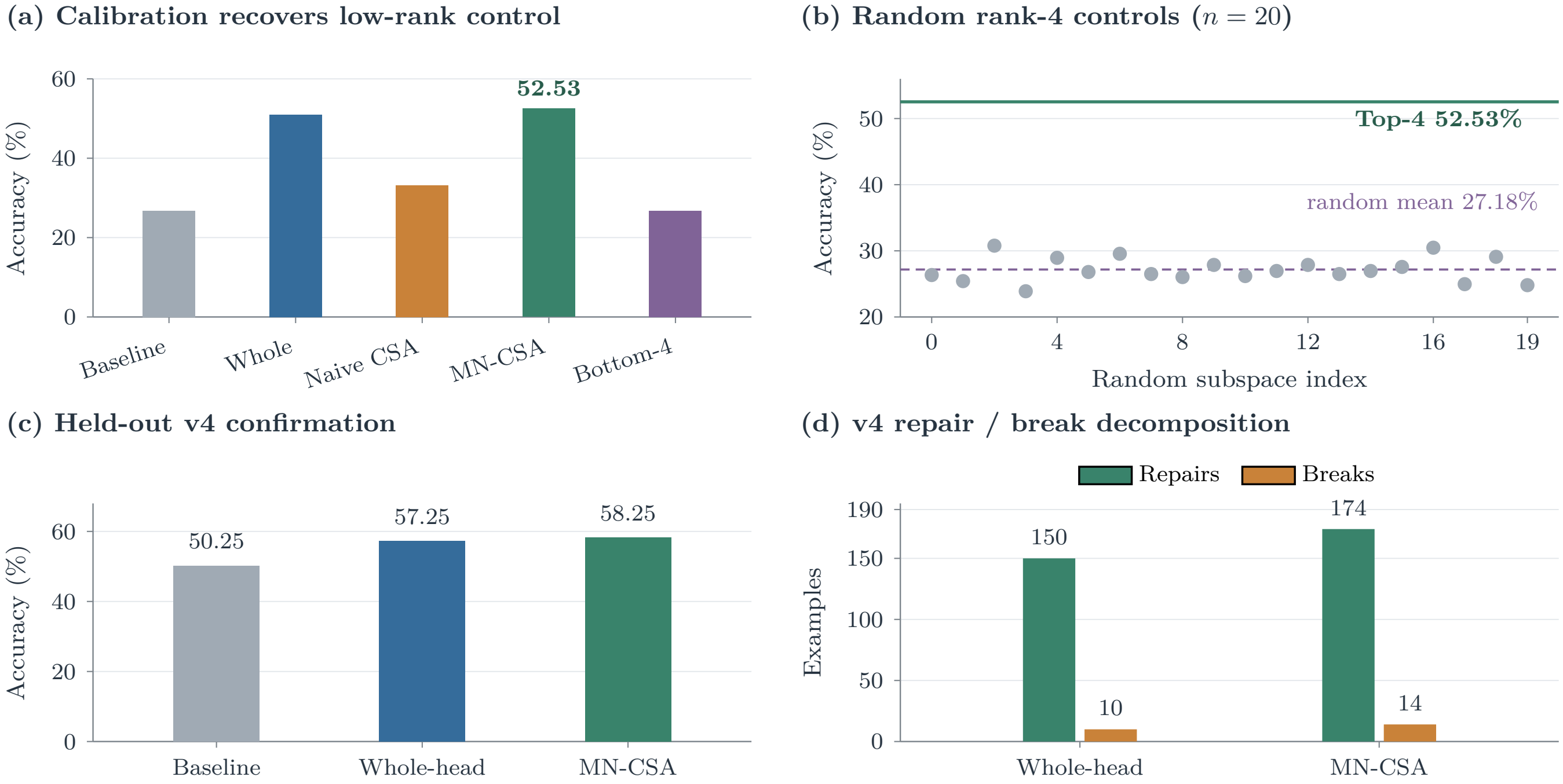\else\end{figure}\fi

\subsection{Held-out v4 Confirmation}
\label{app:v4}
\ifaclstyle\begin{table*}[!tp]\else\begin{table}[H]\fi
\centering
\begin{minipage}[t]{0.72\textwidth}
\vspace{0pt}
\captionsetup{justification=raggedright}
\setlength{\tabcolsep}{3pt}
\centering\fontsize{8.5}{10.5}\selectfont
\caption{v4 result by visual condition.}
\label{tab:v4_condition}
\vspace{2pt}
\begin{tabular*}{\linewidth}{@{\extracolsep{\fill}}lrrrrrr@{}}
\toprule
\bfseries Condition & \bfseries $N$ & \bfseries Base & \bfseries Whole & \bfseries MN & \bfseries MN--Base & \bfseries MN--Whole \\
\midrule
Dense & 400 & 45.50 & 54.00 & 55.25 & +9.75 & +1.25 \\
Distractors & 400 & 60.00 & 65.75 & 65.25 & +5.25 & -0.50 \\
Grid & 400 & 48.25 & 55.75 & 59.75 & +11.50 & +4.00 \\
Random sparse & 400 & 55.50 & 64.75 & 63.25 & +7.75 & -1.50 \\
Row & 400 & 42.00 & 46.00 & 47.75 & +5.75 & +1.75 \\
\bottomrule
\end{tabular*}
\end{minipage}\hfill
\begin{minipage}[t]{0.25\textwidth}
\vspace{0pt}
\captionsetup{justification=raggedright}
\setlength{\tabcolsep}{3pt}
\centering\fontsize{8.5}{10.5}\selectfont
\caption{v4 action distribution of the synthetic router.}
\label{tab:v4_actions}
\begin{tabular*}{\linewidth}{@{\extracolsep{\fill}}rrr@{}}
\toprule
\bfseries $\alpha$ & \bfseries Count & \bfseries Fraction \\
\midrule
0 & 27 & 1.35\% \\
1 & 1396 & 69.80\% \\
1.5 & 32 & 1.60\% \\
2 & 47 & 2.35\% \\
4 & 498 & 24.90\% \\
\bottomrule
\end{tabular*}
\end{minipage}
\ifaclstyle\end{table*}\else\end{table}\fi

\ifaclstyle\begin{table*}[!tp]\else\begin{table}[H]\fi
\centering\fontsize{8.5}{10.5}\selectfont
\caption{v4 result by ground-truth count. Accuracies are fractions; differences are percentage points.}
\label{tab:v4_count}
\begin{tabular}{@{}rrrrrrr@{}}
\toprule
\bfseries Count & \bfseries Base & \bfseries Whole & \bfseries MN & \bfseries Whole--Base & \bfseries MN--Base & \bfseries MN--Whole \\
\midrule
1 & 1.000 & 1.000 & 1.000 & 0.0 & 0.0 & 0.0 \\
2 & .995 & 1.000 & 1.000 & +0.5 & +0.5 & 0.0 \\
3 & .880 & .960 & .940 & +8.0 & +6.0 & -2.0 \\
4 & .750 & .860 & .850 & +11.0 & +10.0 & -1.0 \\
5 & .515 & .690 & .730 & +17.5 & +21.5 & +4.0 \\
6 & .425 & .545 & .565 & +12.0 & +14.0 & +2.0 \\
7 & .135 & .300 & .360 & +16.5 & +22.5 & +6.0 \\
8 & .145 & .155 & .155 & +1.0 & +1.0 & 0.0 \\
9 & .125 & .120 & .135 & -0.5 & +1.0 & +1.5 \\
10 & .055 & .095 & .090 & +4.0 & +3.5 & -0.5 \\
\bottomrule
\end{tabular}
\ifaclstyle\end{table*}\else\end{table}\fi
The primary fixed comparison is MN-CSA versus baseline: $+8.00\pp$, 95\% CI $[+6.7,+9.3]\pp$, exact McNemar $p=2.67\times10^{-36}$.
MN-CSA versus whole-head is a prespecified secondary comparison: $+1.00\pp$, 95\% CI $[+0.3,+1.7]\pp$, exact McNemar $p=0.0078$.
Both comparisons are statistically supported.

\subsection{Geometry Baselines at Equal and Tuned Displacement}
\label{app:add_geometry}
All geometries act on the pooled L18H13 coordinate and add the same $\Delta h$ to every token, under the Stage-1 policy of the v4 confirmation. G1 is naive projection with a global gain $\kappa$, $\Delta h=\kappa(\alpha-1)P_Uh$. G2 is \mncsa{} (Eq.~\ref{eq:mncsa}) with gain $\beta$. G3 is a mean-difference count direction, the difference between mean pooled activations of high-count ($\ge6$) and low-count ($\le3$) v1 examples, applied as $\beta(\alpha-1)\|h\|_2\,d$. G4 is the top-four centered principal-component subspace of pooled v1 activations, with exactly the same projection and per-example norm matching as \mncsa{}. All directions and subspaces are fit on v1 ($N=300$), the cohort on which $U_4$ was estimated; v3 is used only for gain selection and v4 only for the single committed evaluation. The mean-difference direction lies almost entirely inside the PCA-4 span ($\|P_Vd\|_2=0.997$) and only partly inside $U_4$ ($\|P_Ud\|_2=0.60$).

\ifaclstyle\begin{table*}[!tp]\else\begin{table}[H]\fi
\centering\fontsize{8.5}{10.5}\selectfont
\caption{Geometry baselines on the v3 development subset and held-out v4 ($N=2000$; 604 edited rows, \noop{} rows keep the baseline prediction). $\Delta$ is relative to \mncsa{} at $\beta=1$ over all 2,000 rows, with paired bootstrap 95\% CIs; discordance is configuration-only : \mncsa{}-only. Starred gains are selected on development.}
\label{tab:add_geometry}
\setlength{\tabcolsep}{3pt}
\begin{tabular}{@{}lrrrrrrr@{}}
\toprule
\bfseries Configuration & \bfseries Dev /653 & \bfseries v4 /2000 & \bfseries Repairs & \bfseries Breaks & \bfseries $\Delta$ (pp) [95\% CI] & \bfseries Disc. & \bfseries $p$ \\
\midrule
Baseline & 175 & 1,005 & -- & -- & $-8.00$ & 14:174 & $2.7\times10^{-36}$ \\
\midrule
\multicolumn{8}{@{}l}{\itshape Equal displacement} \\
\mncsa{} (G2), $\beta=1$ & 343 & \textbf{1,165} & 174 & 14 & -- & -- & -- \\
Mean-difference (G3), $\beta=1$ & 282 & 1,109 & 111 & 7 & $-2.80$ $[-3.70,-1.95]$ & 11:67 & $6.1\times10^{-11}$ \\
Activation PCA-4 (G4), $\beta=1$ & 226 & 1,055 & 53 & 3 & $-5.50$ $[-6.65,-4.40]$ & 15:125 & $8.9\times10^{-23}$ \\
Naive projection (G1), $\kappa=1$ & 217 & 1,047 & 44 & 2 & $-5.90$ $[-7.10,-4.75]$ & 16:134 & $2.2\times10^{-24}$ \\
\midrule
\multicolumn{8}{@{}l}{\itshape Development-tuned gain} \\
\mncsa{} (G2), $\beta^*=1.5$ & 393 & 1,189 & 223 & 39 & $+1.20$ $[+0.25,+2.15]$ & 57:33 & 0.015 \\
Naive projection (G1), $\kappa^*=5$ & 395 & 1,201 & 225 & 29 & $+1.80$ $[+0.90,+2.70]$ & 61:25 & $1.3\times10^{-4}$ \\
Mean-difference (G3), $\beta^*=2.5$ & 397 & 1,193 & 219 & 31 & $+1.40$ $[+0.55,+2.25]$ & 53:25 & 0.002 \\
Activation PCA-4 (G4), $\beta^*=3$ & 350 & 1,152 & 171 & 24 & $-0.65$ $[-1.35,+0.05]$ & 17:30 & 0.079 \\
\bottomrule
\end{tabular}
\ifaclstyle\end{table*}\else\end{table}\fi

At equal displacement, the sensitivity subspace is the most effective geometry tested, ahead of the mean-difference direction, activation PCA-4, and the uncalibrated naive edit; plotted against realized displacement norm, the naive and \mncsa{} dose--response curves coincide, as expected from their collinearity. With development-tuned gains the geometries converge (tuned \mncsa{} versus tuned mean-difference: $-0.20\pp$, $[-0.65,+0.25]$, $p=0.52$), and larger gains buy repairs at the price of roughly twice as many breaks (24--39 versus 14). This matches the conclusion of Appendix~\ref{app:geometry_trajectory_ablation} that gain is a separate calibration variable: the geometries differ in the displacement they need, and \mncsa{} reaches its accuracy at the smallest displacement and without any gain search. The selected $\kappa^*$ and $\beta^*_{\rm G4}$ lie on the edge of the prespecified grids, which were not extended.

\FloatBarrier
\Needspace{6\baselineskip}
\section{Multi-Actuator Replication in Llama}
\label{app:multi_actuator}
The original geometry/routing pipeline deliberately concentrates on \headname{}, but this leaves open whether its finite controllability is an idiosyncratic property of the selected head.
The multi-actuator experiment revisits the complete historical Llama candidate pool after the primary L18H13 mechanism/control pipeline has already been established. The seven coordinates are L33H1, L3H4, L18H13, L8H30, L3H11, L13H11, and L33H21. All seven are evaluated on the same fresh $N=300$ cohort with $\alpha\in\{0.5,1,1.5\}$ and no head-specific tuning. Baseline outputs are identical across runs, and all raw outcomes (900 per head) are fixed before the primary endpoint is computed.

Six of seven candidates satisfy the prespecified bidirectional replication criterion after Holm correction across 14 directional tests: L33H1, L3H4, L18H13, L3H11, L13H11, and L33H21. This includes five of the six non-primary candidates. L8H30 does not replicate: attenuation consistency is 50.67\% and amplification consistency is 47.00\%, both near chance and in the wrong mean-shift regime. In contrast, the replicated heads span a wide effect-strength range, from L3H11 (64.0\% attenuation; 57.67\% amplification) to L18H13 (91.67\%; 92.67\%). The actuator phenomenon thus generalizes beyond L18H13, and L8H30 shows that bidirectional finite control is verified by intervention rather than inferred from candidacy.

\ifaclstyle\begin{table*}[!tp]\else\begin{table}[H]\fi
\centering\fontsize{8.5}{10.5}\selectfont
\caption{Fresh-cohort multi-actuator endpoint. ``Att.'' and ``Amp.'' are directional consistencies for $\alpha=0.5$ and $1.5$; reported $p$-values are Holm-adjusted across all 14 head-by-direction tests. Replication requires both adjusted directional tests to pass.}
\label{tab:multi_actuator}
\setlength{\tabcolsep}{3.0pt}
\begin{tabular}{@{}lrrrrrrc@{}}
\toprule
\bfseries Head & \bfseries Att. & \bfseries $p_{\rm Holm}$ & \bfseries Amp. & \bfseries $p_{\rm Holm}$ & \bfseries Strict & \bfseries Mean shift (D/U) & \bfseries Rep. \\
\midrule
L33H1 & 85.33\% & $8.74\times10^{-37}$ & 86.00\% & $2.69\times10^{-38}$ & 76.67\% & $-0.04161/+0.04012$ & yes \\
L3H4 & 79.67\% & $2.34\times10^{-25}$ & 69.67\% & $2.70\times10^{-11}$ & 56.67\% & $-0.01034/+0.00565$ & yes \\
L18H13 & 91.67\% & $1.37\times10^{-53}$ & 92.67\% & $9.46\times10^{-57}$ & 89.00\% & $-0.07725/+0.07312$ & yes \\
L8H30 & 50.67\% & 0.8625 & 47.00\% & 0.8637 & 24.67\% & $+0.00054/-0.00267$ & no \\
L3H11 & 64.00\% & $3.55\times10^{-6}$ & 57.67\% & 0.0139 & 40.00\% & $-0.00822/+0.00270$ & yes \\
L13H11 & 71.33\% & $3.81\times10^{-13}$ & 72.67\% & $9.54\times10^{-15}$ & 49.67\% & $-0.00947/+0.00819$ & yes \\
L33H21 & 61.00\% & $3.31\times10^{-4}$ & 67.33\% & $5.70\times10^{-9}$ & 40.00\% & $-0.00436/+0.00328$ & yes \\
\bottomrule
\end{tabular}
\ifaclstyle\end{table*}\else\end{table}\fi

The endpoint yields 6/7 replicated heads and 5/6 replicated non-primary heads. Because the seven-head pool was historically screened, the endpoint is a within-model robustness check: the primary L18H13 phenomenon survives a prospectively fixed test across the complete historical candidate set, and L8H30 supplies a negative control that separates candidacy from finite controllability.

\ifaclstyle\begin{figure*}[!tp]\else\begin{figure}[!htbp]\fi
\centering
\pgfmathsetlengthmacro{\maHalf}{0.5*\textwidth}
\pgfmathsetlengthmacro{\maWidth}{0.5*\textwidth-1.70cm}
\pgfmathsetlengthmacro{\maRight}{\maHalf+1.20cm}
\begin{tikzpicture}
\pgfplotsset{maaxis/.style={scale only axis,width=\maWidth,height=3.15cm,anchor=north west,
axis x line*=bottom,axis y line*=left,axis line style={draw=aromaInk!60,line width=.45pt},
tick style={draw=aromaInk!60,line width=.4pt},tick align=outside,major tick length=2pt,
tick label style={font=\rmfamily\fontsize{7.5}{9}\selectfont,text=aromaInk},label style={font=\rmfamily\fontsize{8}{9.5}\selectfont,text=aromaInk},
ymajorgrids=true,grid style={draw=aromaGrid,line width=.35pt},scaled ticks=false,clip=false}}
\tikzset{matitle/.style={anchor=south west,align=left,text width=\dimexpr.5\textwidth-10pt\relax,inner sep=0pt,text=aromaInk,font=\rmfamily\fontsize{8.5}{10.5}\selectfont\bfseries}}

\node[matitle] at (0,0) {(a) Directional consistency on a fresh $N=300$ cohort};
\begin{axis}[maaxis,at={(1.05cm,-.95cm)},ybar,bar width=7pt,ymin=.4,ymax=1.0,
ylabel={Consistency},symbolic x coords={L33H1,L3H4,L18H13,L8H30,L3H11,L13H11,L33H21},xtick=data,
xticklabel style={rotate=35,anchor=east},ytick={.4,.5,.6,.7,.8,.9,1.0},
legend style={font=\rmfamily\fontsize{7.5}{9}\selectfont,draw=none,fill=none,at={(.5,1.03)},anchor=south,legend columns=2},legend image code/.code={\path[#1] (0pt,-2pt) rectangle (8pt,3pt);}]
\addplot[draw=none,fill=aromaBlue,bar shift=-4pt] coordinates {(L33H1,.853333)(L3H4,.796667)(L18H13,.916667)(L8H30,.506667)(L3H11,.64)(L13H11,.713333)(L33H21,.61)};
\addplot[draw=none,fill=aromaOrange,bar shift=4pt] coordinates {(L33H1,.86)(L3H4,.696667)(L18H13,.926667)(L8H30,.47)(L3H11,.576667)(L13H11,.726667)(L33H21,.673333)};
\legend{Attenuation,Amplification}
\draw[densely dashed,aromaGray,line width=.6pt] (axis cs:L33H1,.5) -- (axis cs:L33H21,.5);
\end{axis}

\node[matitle] at (\maHalf,0) {(b) Strict bidirectional ordering and replication status};
\begin{axis}[maaxis,at={(\maRight,-.95cm)},ybar,bar width=17pt,ymin=0,ymax=1.0,
ylabel={Strict-order fraction},symbolic x coords={L33H1,L3H4,L18H13,L8H30,L3H11,L13H11,L33H21},xtick=data,
xticklabel style={rotate=35,anchor=east},ytick={0,.2,.4,.6,.8,1.0},legend image code/.code={\path[#1] (0pt,-2pt) rectangle (8pt,3pt);}]
\addplot[draw=none,fill=aromaGreen] coordinates {(L33H1,.766667)(L3H4,.566667)(L18H13,.89)(L8H30,.246667)(L3H11,.4)(L13H11,.496667)(L33H21,.4)};
\draw[densely dashed,aromaGray,line width=.6pt] (axis cs:L33H1,.5) -- (axis cs:L33H21,.5);
\node[font=\rmfamily\fontsize{7.5}{9}\selectfont\bfseries,text=aromaInk,anchor=north east,align=right] at (axis cs:L33H21,.96) {6/7 replicated\\5/6 non-primary};
\node[font=\rmfamily\fontsize{7.5}{9}\selectfont,text=aromaInk!80,anchor=south] at (axis cs:L8H30,.29) {not replicated};
\end{axis}
\end{tikzpicture}
\caption{\textbf{Bidirectional actuator behavior generalizes across the historical candidate pool.}
\textbf{(a)} Attenuation ($\alpha=0.5$) and amplification ($\alpha=1.5$) directional consistency for all seven candidates on the same fresh $N=300$ cohort; the dashed line marks 0.5. Six heads pass both one-sided exact directional tests after Holm correction across the 14 head-by-direction tests. \textbf{(b)} The stricter per-example ordering $\mu_{0.5}<\mu_1<\mu_{1.5}$ varies substantially across heads and is not itself the replication criterion. L8H30 provides a useful negative case: historical causal candidacy does not automatically imply stable finite controllability.}
\label{fig:multi_actuator}
\ifaclstyle\end{figure*}
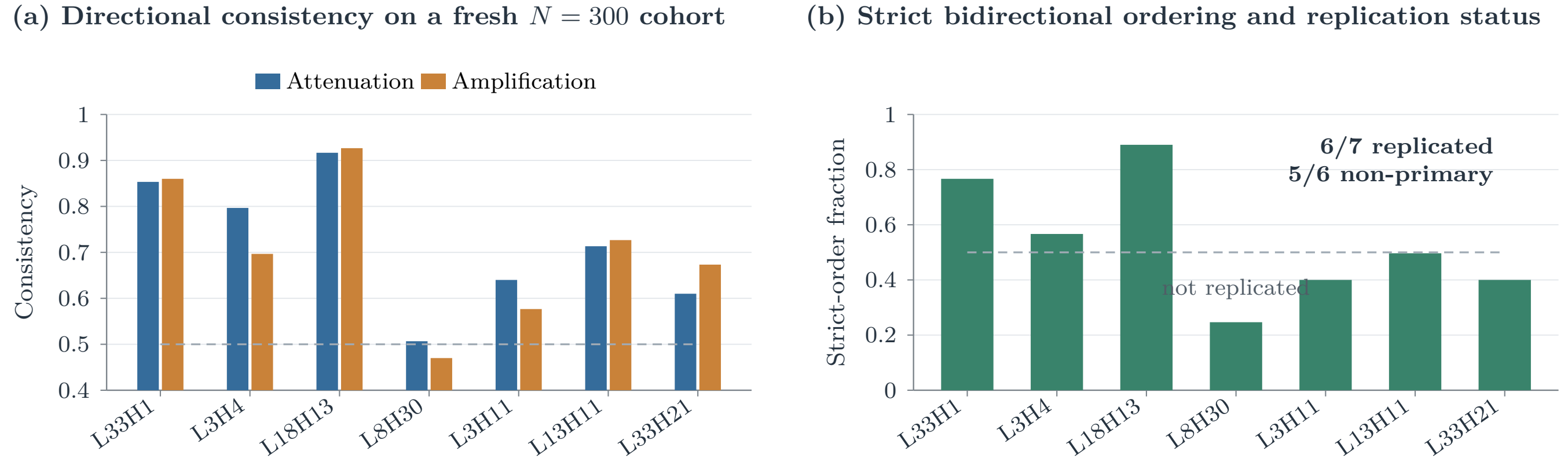\else\end{figure}\fi
\FloatBarrier
\Needspace{6\baselineskip}
\section{IDEFICS Cross-Model Replication}
\label{app:idefics}
\subsection{Actuator and Scoring}
Cross-model experiments use \path{HuggingFaceM4/idefics-9b-instruct} at revision \path{7065afb8a5bb03015700b315f6751dc3dd4e16eb}.
The model exposes eight gated cross-attention blocks with 32 heads per block and 128-dimensional head slices, for 256 screened cross-attention heads.
The intervention is applied to the B2H28 slice at the input to the block-2 cross-attention output projection.
Because decimal numeral strings are multi-token under the IDEFICS tokenizer, the expected numeral is computed from complete teacher-forced sequence log-likelihoods over strings 0--15, including the end-of-utterance terminator; token-length normalization is not used.

The held-out actuator confirmation contains $N=300$ examples, balanced with 30 examples for each count 1--10.
For B2H28, attenuation gives 246/300 downward shifts (82.00\%), amplification gives 248/300 upward shifts (82.67\%), and strict ordering holds on 203/300 (67.67\%).
The corresponding one-sided exact binomial $p$-values are $9.81\times10^{-31}$ and $4.53\times10^{-32}$, both Holm-significant.
The protocol's bidirectional replication criterion is satisfied; strict ordering (67.67\%) is reported alongside the prespecified auxiliary 70\% strong-ordering threshold.

\Needspace{5\baselineskip}
\subsection{IDEFICS Sensitivity Geometry and Robustness}
On a disjoint geometry cohort ($N=300$), the raw uncentered Top-4 explains 0.914 of sensitivity trace ($r_{80}=2$, $r_{90}=4$), and the centered Top-4 explains 0.840 ($r_{80}=4$, $r_{90}=7$).
The raw mean-gradient direction accounts for 0.560 of trace.
Against 10,000 random rank-4 subspaces, no null replicate reaches the raw or centered observed concentration (empirical $p=1/10001$ for each).
The mean projected activation-energy share of the raw Top-4 is only 0.0414, with mean $\rho\approx0.2006$, again separating sensitivity concentration from activation energy.

\ifaclstyle\begin{table*}[!tp]\else\begin{table}[H]\fi
\centering
\fontsize{8.5}{10.5}\selectfont
\caption{Raw and unit-gradient sensitivity robustness in the two model-specific actuator coordinates. Unit normalization changes the estimand by giving every example equal gradient norm; it is therefore reported as robustness, not as a replacement for the raw analysis.}
\label{tab:gradient_norm_robustness}
\begin{tabular}{@{}lrrrr@{}}
\toprule
\bfseries Model / estimand & \bfseries Raw Top-4 & \bfseries Raw centered & \bfseries Unit Top-4 & \bfseries Unit centered \\
\midrule
Llama \headname{} & 0.898 & 0.868 & 0.835 & 0.650 \\
IDEFICS B2H28 & 0.914 & 0.840 & 0.899 & 0.785 \\
\bottomrule
\end{tabular}
\ifaclstyle\end{table*}\else\end{table}\fi
For Llama, the raw mean-gradient trace share is 0.233 and the mean unit-gradient share is 0.564; for IDEFICS the corresponding values are 0.560 and 0.615.
Thus the large raw cross-model difference in shared-direction concentration is substantially reduced after equalizing per-example gradient magnitude.
The normalized analysis shows that the cross-model contrast is estimand-sensitive.
The leading subspaces remain strongly low-dimensional under both estimands, with basis membership in the lower modes varying with normalization.

A separate activation-scale audit finds B2H28 median pooled norm 0.457, ranking 214/256 globally by median norm but 18/32 within cross-attention block B2; its median norm is about 0.89 times the per-image 256-head median.
The head is therefore globally low-scale but locally typical within its block rather than near-dead.
Because projected energy is normalized within the head, the 4.14\% Top-4 activation-energy share is independent of this head-level scale.

\Needspace{5\baselineskip}
\subsection{IDEFICS Equal-Norm MN-CSA Confirmation}
The primary cross-model finite-control test uses the raw uncentered Top-4 and the same magnitude-matched operator as Eq.~\eqref{eq:mncsa} on a fresh $N=300$ cohort.
Table~\ref{tab:idefics_equalnorm} summarizes the result.

\ifaclstyle\begin{table*}[!tp]\else\begin{table}[H]\fi
\centering
\fontsize{8.5}{10.5}\selectfont
\caption{IDEFICS equal-norm rank-4 MN-CSA on a fresh $N=300$ cohort. ``Retention'' is the ratio of mean MN-CSA directional shift to the same-run whole-head directional shift.}
\label{tab:idefics_equalnorm}
\begin{tabular}{@{}lrrr@{}}
\toprule
\bfseries Metric & \bfseries Down & \bfseries Up & \bfseries Strict \\
\midrule
MN-CSA consistency & 65.33\% & 66.00\% & 37.67\% \\
Whole-head consistency & 85.33\% & 88.00\% & 74.00\% \\
Mean MN-CSA shift & 0.01336 & 0.01269 & -- \\
Mean whole-head shift & 0.03300 & 0.03702 & -- \\
Effect retention & 40.47\% & 34.28\% & -- \\
One-sided exact $p$ & $5.98\times10^{-8}$ & $1.59\times10^{-8}$ & -- \\
\bottomrule
\end{tabular}
\ifaclstyle\end{table*}\else\end{table}\fi
Both directional tests remain significant after Holm correction.
This stage establishes causal transfer of the low-rank geometry at matched displacement magnitude, before any gain calibration.

\Needspace{5\baselineskip}
\subsection{IDEFICS Rank and Gain Development Record}
The rank/gain development sequence is disjoint from the equal-norm confirmation and from the final confirmation.
A screen cohort uses replicates 18--19 ($N=100$); a development-validation cohort uses replicates 20--23 ($N=200$); the held-out final cohort uses replicates 24--29 ($N=300$); replicates 30--39 ($N=500$) remain unused reserve data.
All four splits have zero overlap with prior IDEFICS evaluation cohorts.

\ifaclstyle\begin{table*}[!tp]\else\begin{table}[H]\fi
\centering
\fontsize{8.5}{10.5}\selectfont
\caption{IDEFICS rank screen at $\beta=1$ on the development-only screening set. Sensitivity share rises monotonically, whereas balanced finite causal shift peaks at rank 4.}
\label{tab:idefics_rank_screen}
\begin{tabular}{@{}rrrrrr@{}}
\toprule
\bfseries $r$ & \bfseries Sens. share & \bfseries Down cons. & \bfseries Up cons. & \bfseries Balanced shift & \bfseries Down/Up retention \\
\midrule
2  & 85.32\% & 62\% & 65\% & 0.01289 & 40.69\% / 40.50\% \\
4  & 91.36\% & 63\% & 68\% & 0.01417 & 44.71\% / 43.37\% \\
8  & 96.12\% & 63\% & 63\% & 0.00898 & 28.35\% / 39.73\% \\
16 & 98.36\% & 61\% & 61\% & 0.00845 & 26.66\% / 25.79\% \\
\bottomrule
\end{tabular}
\ifaclstyle\end{table*}\else\end{table}\fi

The top two ranks ($r=4,2$) then enter a bounded gain screen with $\beta\in\{0.75,1,1.25,1.5,2\}$.
The predeclared primary ranking metric is balanced mean shift, with minimum directional consistency and strict ordering as secondary criteria.
The top two configurations are $(r=2,\beta=2)$ and $(r=4,\beta=2)$.
On the independent development-validation set, the ordering reverses: $(r=4,\beta=2)$ has balanced shift 0.02344 versus 0.02275 for rank 2 and is selected for final confirmation.
Its development-validation directional consistency is 73.0\%/74.0\%, with 73.69\%/73.62\% effect retention.
No further parameter search is performed after this selection.

\Needspace{5\baselineskip}
\subsection{IDEFICS Held-out Gain-Calibrated Final Confirmation}
The final cohort contains $N=300$ examples from replicates 24--29 and is first evaluated only after $r=4,\beta=2$ is selected on the development-validation set.
The raw Top-4 basis explains 91.36\% of raw sensitivity trace.

\ifaclstyle\begin{table*}[!tp]\else\begin{table}[H]\fi
\centering
\fontsize{8.5}{10.5}\selectfont
\caption{IDEFICS held-out final confirmation for the fixed gain-calibrated rank-4 intervention ($r=4,\beta=2$).}
\label{tab:idefics_final}
\begin{tabular}{@{}lrr@{}}
\toprule
\bfseries Metric & \bfseries Down & \bfseries Up \\
\midrule
MN directional consistency & 79.67\% & 79.67\% \\
Whole-head directional consistency & 84.00\% & 84.33\% \\
MN mean shift & 0.02555 & 0.02880 \\
Whole-head mean shift & 0.03151 & 0.03534 \\
Effect retention & 81.09\% & 81.49\% \\
Retention 95\% CI & [71.21, 92.15]\% & [71.90, 92.15]\% \\
Directional exact $p$ & $2.34\times10^{-26}$ & $2.34\times10^{-26}$ \\
MN--whole mean difference & $-0.00596$ & $-0.00654$ \\
MN--whole 95\% CI & $[-0.00969,-0.00230]$ & $[-0.01063,-0.00257]$ \\
\bottomrule
\end{tabular}
\ifaclstyle\end{table*}\else\end{table}\fi
Strict ordering holds on 187/300 = 62.33\%, versus 211/300 = 70.33\% for whole-head intervention.
Both directional exact tests survive Holm correction.
The gain-calibrated edit therefore recovers 81\% of the whole-head mean effect.
Because $\beta=2$, these retention ratios quantify recovered whole-head behavioral effect under a calibrated low-rank displacement.
The fixed low-rank controller is directionally consistent on 239/300 examples in each direction, with paired mean MN-minus-whole differences reported in Table~\ref{tab:idefics_final}. The rank screen also rules out a simple ``capture more spectral mass'' explanation and motivates treating finite control as a separate behavioral property (Fig.~\ref{fig:idefics_crossmodel}). The cross-model conclusion is structural: low-rank causal geometry replicates, and finite gain is calibrated per model.

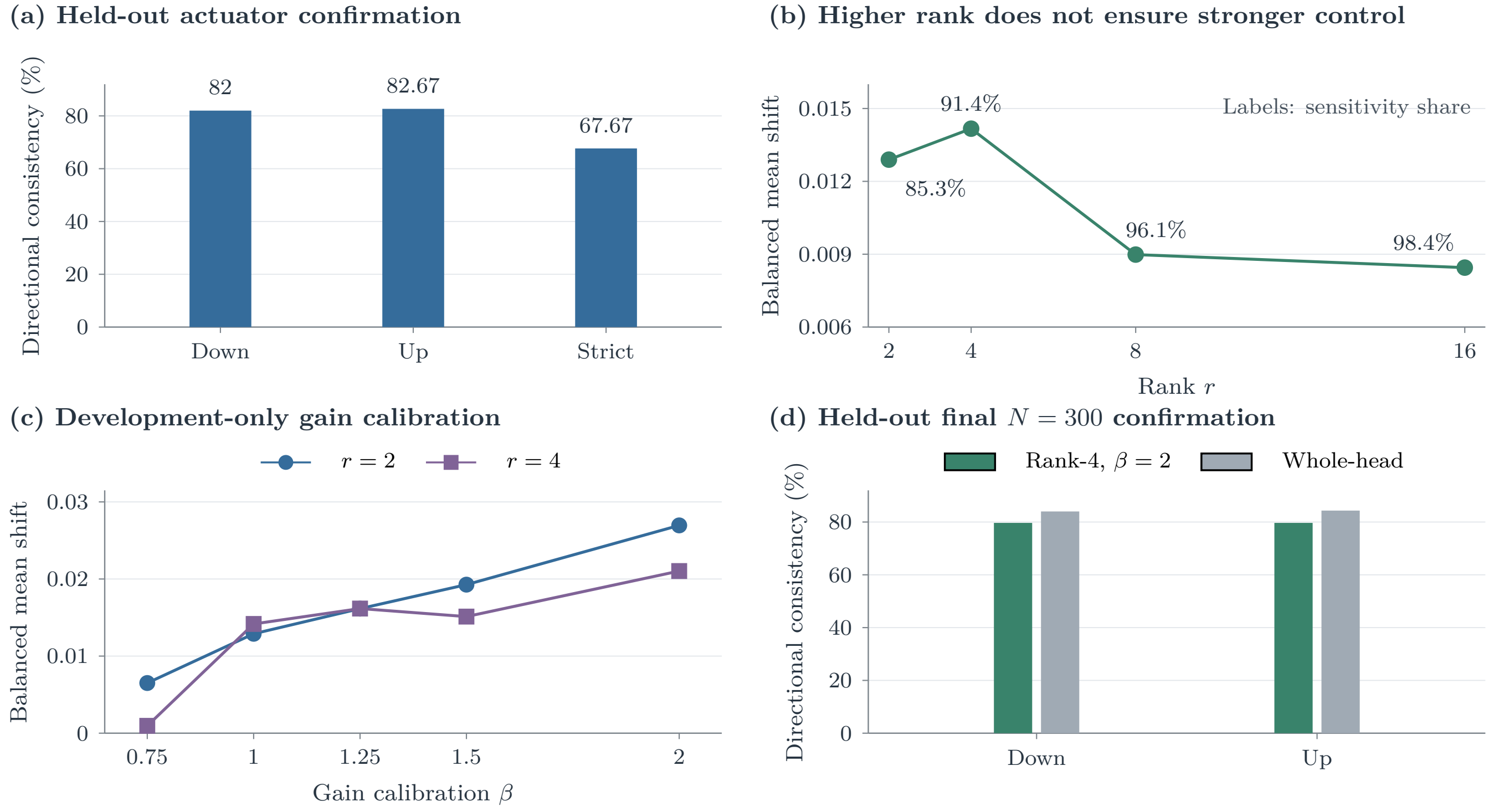
\ifaclstyle\begin{figure*}[!tp]\else\begin{figure}[!htbp]\fi
\centering

\pgfmathsetlengthmacro{\xmHalf}{0.5*\textwidth}
\pgfmathsetlengthmacro{\xmWidth}{0.5*\textwidth-1.70cm}
\pgfmathsetlengthmacro{\xmRight}{\xmHalf+1.20cm}

\begin{tikzpicture}

\pgfplotsset{
  xmaxis/.style={
    scale only axis,
    width=\xmWidth,
    height=2.90cm,
    anchor=north west,
    axis x line*=bottom,
    axis y line*=left,
    axis line style={
      draw=aromaInk!60,
      line width=.45pt
    },
    tick style={
      draw=aromaInk!60,
      line width=.4pt
    },
    tick align=outside,
    major tick length=2pt,
    tick label style={
      font=\rmfamily\fontsize{7.5}{9}\selectfont,
      text=aromaInk
    },
    label style={
      font=\rmfamily\fontsize{8}{9.5}\selectfont,
      text=aromaInk
    },
    ymajorgrids=true,
    grid style={
      draw=aromaGrid,
      line width=.35pt
    },
    scaled ticks=false,
    clip=true
  },
  xmlegend/.style={
    legend style={
      font=\rmfamily\fontsize{7.5}{9}\selectfont,
      draw=none,
      fill=none,
      inner sep=0pt,
      at={(.5,1.05)},
      anchor=south,
      legend columns=2,
      column sep=8pt
    }
  }
}

\tikzset{
  xmtitle/.style={
    anchor=west,
    inner sep=0pt,
    text=aromaInk,
    font=\rmfamily\fontsize{8.5}{10.5}\selectfont\bfseries
  },
  xmannotation/.style={
    font=\rmfamily\fontsize{7.5}{9}\selectfont,
    text=aromaInk
  }
}

\node[xmtitle] at (0,0)
  {(a) Held-out actuator confirmation};

\begin{axis}[
  xmaxis,
  at={(1.15cm,-.80cm)},
  ybar,
  bar width=21pt,
  xmin=-.6,
  xmax=2.6,
  ymin=0,
  ymax=92,
  ylabel={Directional consistency (\%)},
  xtick={0,1,2},
  xticklabels={Down,Up,Strict},
  ytick={0,20,40,60,80},
  nodes near coords,
  every node near coord/.append style={
    font=\rmfamily\fontsize{7.5}{9}\selectfont\bfseries,
    text=aromaInk,
    anchor=south,
    yshift=2pt
  }
,legend image code/.code={\path[#1] (0pt,-2pt) rectangle (8pt,3pt);}]
\addplot[draw=none,fill=aromaBlue]
  coordinates {
    (0,82.00)
    (1,82.67)
    (2,67.67)
  };
\end{axis}

\node[xmtitle] at (\xmHalf,0)
  {(b) Higher rank does not ensure stronger control};

\begin{axis}[
  xmaxis,
  at={(\xmRight,-.80cm)},
  xmin=1.5,
  xmax=16.5,
  ymin=.006,
  ymax=.016,
  xlabel={Rank $r$},
  ylabel={Balanced mean shift},
  xtick={2,4,8,16},
  ytick={.006,.009,.012,.015},
  yticklabels={0.006,0.009,0.012,0.015}
]
\addplot[
  color=aromaGreen,
  line width=1pt,
  mark=*,
  mark size=2.3pt
]
coordinates {
  (2,.0128943)
  (4,.0141695)
  (8,.00898424)
  (16,.00844754)
};

\node[
  xmannotation,
  anchor=north west,
  xshift=2pt,
  yshift=-3pt
] at (axis cs:2,.0128943) {85.3\%};

\node[xmannotation,anchor=south]
  at (axis cs:4,.01443) {91.4\%};

\node[xmannotation,anchor=south]
  at (axis cs:8.5,.00923) {96.1\%};

\node[xmannotation,anchor=south east]
  at (axis cs:16,.00870) {98.4\%};

\node[
  xmannotation,
  text=aromaInk!85,
  anchor=north east
] at (axis cs:16.4,.01585)
  {Labels: sensitivity share};
\end{axis}

\node[xmtitle] at (0,-4.80cm)
  {(c) Development-only gain calibration};

\begin{axis}[
  xmaxis,
  xmlegend,
  at={(1.15cm,-5.65cm)},
  xmin=.65,
  xmax=2.10,
  ymin=0,
  ymax=.0315,
  xlabel={Gain calibration $\beta$},
  ylabel={Balanced mean shift},
  xtick={.75,1,1.25,1.5,2},
  ytick={0,.01,.02,.03},
  yticklabels={0,0.01,0.02,0.03}
]
\addplot[
  color=aromaBlue,
  line width=1pt,
  mark=*,
  mark size=2.2pt
]
coordinates {
  (.75,.0065070)
  (1,.0128943)
  (1.25,.0161549)
  (1.5,.0192676)
  (2,.0269541)
};

\addplot[
  color=aromaPurple,
  line width=1pt,
  mark=square*,
  mark size=2.2pt
]
coordinates {
  (.75,.0009541)
  (1,.0141695)
  (1.25,.0161662)
  (1.5,.0151236)
  (2,.0210314)
};

\legend{$r=2$,$r=4$}
\end{axis}

\node[xmtitle] at (\xmHalf,-4.80cm)
  {(d) Held-out final $N=300$ confirmation};

\begin{axis}[
  xmaxis,
  xmlegend,
  at={(\xmRight,-5.65cm)},
  ybar,
  bar width=13pt,
  xmin=-.6,
  xmax=1.6,
  ymin=0,
  ymax=92,
  ylabel={Directional consistency (\%)},
  xtick={0,1},
  xticklabels={Down,Up},
  ytick={0,20,40,60,80}
,legend image code/.code={\path[#1] (0pt,-2pt) rectangle (8pt,3pt);}]
\addplot[
  draw=none,
  fill=aromaGreen,
  bar shift=-8pt,
  forget plot
]
coordinates {
  (0,79.67)
  (1,79.67)
};

\addplot[
  draw=none,
  fill=aromaGray,
  bar shift=8pt,
  forget plot
]
coordinates {
  (0,84.00)
  (1,84.33)
};

\addlegendimage{area legend,fill=aromaGreen,draw=none}
\addlegendentry{Rank-4, $\beta=2$}

\addlegendimage{area legend,fill=aromaGray,draw=none}
\addlegendentry{Whole-head}
\end{axis}

\end{tikzpicture}

\par\vspace{4pt}
{\footnotesize\color{aromaInk}
Final mean-effect retention:
\textbf{81.1\% downward / 81.5\% upward}
\quad ($r=4$, $\beta=2$; gain-calibrated).\par}

\caption{
\textbf{Cross-model replication and held-out gain calibration
in IDEFICS-9B-Instruct.}
\textbf{(a)} Independently screened B2H28 shows held-out
bidirectional actuator behavior.
\textbf{(b)} Sensitivity trace rises monotonically with rank
(point labels), but finite causal steering peaks at rank 4
and deteriorates at ranks 8 and 16.
\textbf{(c)} On the development-only screen set, larger gain
substantially increases balanced low-rank steering; only the
top two configurations are carried forward.
\textbf{(d)} Independent dev validation selects
$r=4,\beta=2$ before the final cohort is touched.
On the held-out final set, the fixed low-rank intervention
reaches 79.67\% directional consistency in both directions
and recovers about 81\% of same-run whole-head mean shift.
Because $\beta=2$, this is a gain-calibrated, not equal-norm,
comparison.
}
\label{fig:idefics_crossmodel}
\ifaclstyle\end{figure*}\else\end{figure}\fi
\FloatBarrier
\Needspace{6\baselineskip}
\section{Local Response Law and Minimal-Effort Actuation}
\label{app:mea}

This section validates the local response law and the \mea{} controller (Gap~2). All experiments use the same fixed Llama model revision, causal actuator \headname{}, and Top-4 basis, and comprise local mechanism validation, oracle control with matched-budget controls, free-form generation under the realized edits, and displacement and compute accounting.

\subsection{Continuous Local-Response Validation}
For each example, let $P_U=U_4U_4^\top$, $d_{\mathrm{MN}}=P_Uh/\|P_Uh\|$, and $d_{\mathrm{SA}}=P_Ug/\|P_Ug\|$. The $N=160$ run evaluates continuous numeral movement under matched-norm activation- and gradient-aligned directions. The median $\cos(P_Ug,P_Uh)$ is 0.788, with 25th/75th percentiles 0.734/0.841; only 1/160 examples has non-positive cosine. At $\beta=.05$, the regression of realized on first-order predicted movement has slope 1.0573, intercept $-1.29\times10^{-4}$, Pearson $r=.9936$, Spearman $\rho=.9990$, MAE .001313, RMSE .003178, and 100\% sign agreement. At $\beta=.10$ and $.20$, Pearson correlation remains .9769 and .9371, while magnitude error increases. A four-example finite-difference check using relative steps .01 and .02 yields 100\% sign agreement and sub-percent median relative error, independently validating the implemented derivative.

\ifaclstyle\begin{table*}[!tp]\else\begin{table}[H]\fi
\centering\fontsize{8.5}{10.5}\selectfont
\caption{Continuous local-response validation. The local law remains directionally reliable as scale grows, but magnitude calibration degrades.}
\label{tab:local_response}
\begin{tabular}{@{}rrrrrr@{}}
\toprule
\bfseries $\beta$ & \bfseries Slope & \bfseries Pearson & \bfseries Spearman & \bfseries MAE & \bfseries Sign agreement \\
\midrule
0.05 & 1.0573 & 0.9936 & 0.9990 & 0.00131 & 100\% \\
0.10 & 1.0994 & 0.9769 & 0.9964 & 0.00523 & 100\% \\
0.20 & 1.1000 & 0.9371 & 0.9873 & 0.0194 & 100\% \\
\bottomrule
\end{tabular}
\ifaclstyle\end{table*}\else\end{table}\fi

\subsection{Oracle MEA Protocol and Horizon}
Oracle \mea{} sets the target to ground truth only for the purpose of measuring controllability. At each step it identifies the current highest non-target numeral, computes the target--competitor logit margin, projects the margin gradient into the Top-4 subspace, takes the minimum predicted crossing step clipped to $\tau=.10|\alpha_{\rm prop}-1|\|h\|$, and re-linearizes. The primary 20-step run contains 112 baseline-wrong examples and no projected-gradient abstentions.

\ifaclstyle\begin{table*}[!tp]\else\begin{table}[H]\fi
\centering\fontsize{8.5}{10.5}\selectfont
\caption{Cumulative oracle MEA target reachability.}
\label{tab:mea_horizon}
\begin{tabular}{@{}rrrrrrrr@{}}
\toprule
\bfseries Max steps & \bfseries 1 & \bfseries 2 & \bfseries 3 & \bfseries 5 & \bfseries 10 & \bfseries 15 & \bfseries 20 \\
\midrule
Success (\%) & 5.36 & 10.71 & 12.50 & 20.54 & 55.36 & 75.89 & 81.25 \\
\bottomrule
\end{tabular}
\ifaclstyle\end{table*}\else\end{table}\fi

The strict primary success count is 91/112. Of the 21 unresolved examples, nine finish within margin $\le10^{-3}$ of the target boundary, but they are counted as failures. Unresolved trajectories are associated with larger initial effort, sustained clipping, and more active-competitor changes.
In BF16, an exact zero margin can still leave deterministic argmax at the competitor; in that boundary case \mea{} takes one legal trust-radius step in the same projected gradient direction rather than introducing a new clearance threshold.

\subsection{Matched-Budget Controls}
The fixed-direction control computes the initial target-specific projected margin direction once and then receives the exact realized \mea{} step norms. It reaches 17/112=15.18\% versus 91/112=81.25\% for \mea{}; the paired table is 17 both, 74 MEA-only, 0 fixed-only, and 21 neither (exact $p=1.06\times10^{-22}$). An iterative activation-aligned control using the same realized budget reaches 1/112=0.89\%; the paired table is 1 both, 90 MEA-only, 0 activation-only, and 21 neither (exact $p=1.62\times10^{-27}$). These controls establish target-specific, state-dependent re-linearization as the operative ingredient under the tested budget. The fixed-direction control holds both the projected target-gradient direction and the active competitor fixed, so it isolates state-dependent re-linearization as a whole.

\subsection{Free-Form Generation under MEA Edits}
\label{app:add_freegen}
Because the cumulative \mea{} edit is not stored, the protocol is re-executed for each cohort with its original settings (oracle: float32 and numerals 1--10; natural gated set: BF16 and numerals 0--15). The re-runs reproduce every original final prediction (112/112 and 63/63) with zero per-step norm deviation. The edited model then answers by greedy decoding (at most 16 new tokens), with the edit applied either at prompt positions only or also at every generated position; identity generation is run on the same examples. The first integer in the output is parsed, including English number words up to fifteen and ``no''/``none'' as zero; no output failed to parse.

\ifaclstyle\begin{table*}[!tp]\else\begin{table}[H]\fi
\centering\fontsize{8.5}{10.5}\selectfont
\caption{Greedy free-form answers with and without the \mea{} edit. Discordance is edited-only : identity-only. Agreement is between the parsed generated answer and the closed-set proxy argmax under the edit.}
\label{tab:add_freegen}
\begin{tabular}{@{}lrrrrrr@{}}
\toprule
\bfseries Cohort & \bfseries $N$ & \bfseries Identity correct & \bfseries Edited correct & \bfseries Disc. & \bfseries Exact paired $p$ & \bfseries Agreement with proxy \\
\midrule
Oracle \mea{} (procedural) & 112 & 0 & 80 & 80:0 & $1.7\times10^{-24}$ & 79.5\% \\
Deployed \mea{} (natural, gated) & 63 & 7 & 54 & 51:4 & $2.1\times10^{-11}$ & 95.2\% \\
\bottomrule
\end{tabular}
\ifaclstyle\end{table*}\else\end{table}\fi

The two application modes give the same answer on 174/175 examples. Oracle disagreements with the proxy are mostly one numeral below the target, because \mea{} stops as soon as the target becomes the argmax and leaves the edited state at the decision boundary.

\subsection{MEA Displacement and Compute Cost}
\label{app:add_cost}
Because the trust radius scales with the current-state norm, the net displacement of a trajectory is not bounded by $2B$; we therefore measure it directly. Net displacement and path length are nearly equal, so trajectories move consistently rather than oscillating.

\ifaclstyle\begin{table*}[!tp]\else\begin{table}[H]\fi
\centering\fontsize{8.5}{10.5}\selectfont
\caption{Realized \mea{} net displacement relative to the per-example reference budget $B=|\alpha_{\rm prop}-1|\|h_0\|_2$ (unedited state). $4B$ is the largest displacement already audited in the routing candidate grid.}
\label{tab:add_cost}
\begin{tabular}{@{}lrrrr@{}}
\toprule
\bfseries Cohort & \bfseries $N$ & \bfseries Median net$/B$ & \bfseries Net $\le B$ & \bfseries Net $\le 4B$ \\
\midrule
Oracle \mea{} (procedural) & 112 & 0.99 & 50\% & 88\% \\
Prospective deployed \mea{} (natural) & 63 & 1.58 & 32\% & 97\% \\
Retrospective deployed \mea{} (natural) & 52 & 1.00 & 50\% & 98\% \\
\bottomrule
\end{tabular}
\ifaclstyle\end{table*}\else\end{table}\fi

On the oracle cohort, required displacement grows with the distance between target and baseline prediction (median net$/B$ of 0.75, 3.75, and 5.79 for distances 1, 2, and 3), and unresolved trajectories exhaust the 20-step horizon. On the evaluation GPU, one forward pass takes 0.71\,s; Stage~1 costs $1.0\times$ a forward pass, gradient feature collection $1.02\times$, and the gate runs on CPU. \mea{} costs $2+2K$ forward-or-backward evaluations: $20.4\times$ a forward pass at the median $K=9$ and $42.9\times$ at $K=20$. Because the gate activates \mea{} on 0.788\% of inputs, the population-average overhead of execution is about $0.16\times$ a forward pass at the median step count. Re-executing the retrospective 52 reproduces every final prediction, while iteration counts match on 29/52 trajectories under BF16 numerical non-determinism; the stored per-step logs are used for step statistics of that cohort.

\FloatBarrier

\subsection{Whole-Head MEA Ablation in Llama}
\label{app:wholehead_mea}
The ablation reruns oracle \mea{} on the same $N=112$ baseline-wrong cohort with the projector $P=UU^\top$ (Top-4) replaced by the identity on the 128-dimensional \headname{} slice; model, float32 precision, numerals, trust radius $\tau=0.10\,|\alpha_{\rm prop}-1|\,\|h_{\rm current}\|_2$, 20-step horizon, and stopping rule are unchanged. Before the ablation, the Top-4 run is re-executed and reproduces the original result exactly (91/112; all 112 final predictions and iteration counts match). The protocol was fixed before the whole-head run.

\ifaclstyle\begin{table*}[!tp]\else\begin{table}[H]\fi
\centering\fontsize{8.5}{10.5}\selectfont
\caption{Oracle \mea{} with and without the Top-4 restriction ($N=112$). Discordance is whole head : Top-4. $B_0$ is the reference budget of Table~\ref{tab:add_cost}; differences are paired means with bootstrap 95\% CIs.}
\label{tab:wholehead_mea}
\begin{tabular}{@{}lrrl@{}}
\toprule
\bfseries Endpoint & \bfseries Top-4 & \bfseries Whole head & \bfseries Paired test \\
\midrule
Target reached & 91 (81.25\%) & 92 (82.14\%) & 10:9, $p=1.00$ \\
Generated answer correct (identity 0/112) & 80 & 88 & 15:7, $p=0.13$ \\
Generated answer $=$ proxy argmax & 89 & 98 & -- \\
Median iterations (all / reached) & 9 / 8 & 8 / 7 & $-0.71$ $[-1.59,0.21]$ \\
Median net displacement$/B_0$ (all / reached) & 0.985 / 0.821 & 0.926 / 0.787 & $-0.26$ $[-0.35,-0.17]$ \\
\bottomrule
\end{tabular}
\ifaclstyle\end{table*}\else\end{table}\fi

Restricting \mea{} to the four sensitivity dimensions does not change reachability or generated-answer correctness. The unrestricted edit reaches its targets with slightly less net displacement, since it can also use directions outside the subspace; the iteration difference is not distinguishable from zero.

\FloatBarrier
\subsection{MEA Replication in IDEFICS}
\label{app:idefics_mea}
\paragraph{Protocol.}
The replication uses \idefics{} at the pinned revision, actuator B2H28, and the fixed raw uncentered Top-4 basis of Appendix~\ref{app:idefics}, in float32. Numeral scores are the complete teacher-forced sequence log-likelihoods of the strings 0--15 used throughout the \idefics{} experiments. \mea{} follows Eq.~\eqref{eq:mea_step} with the target margin over these scores, the shared-offset gradient, a 20-step horizon, and a trust radius $\tau=0.10\,B$ with $B=k\|h_{\rm current}\|_2$. The only selected quantity is $k\in\{0.5,1,2,3\}$, chosen on the 189 baseline-wrong examples of a 200-example development cohort: reachability is 1, 3, 47, and 103 of 189, so $k=3$, the largest grid value, is selected; the fixed-direction control reaches 3/189 at $k=3$. Before development, autograd and central finite differences of the margin agree to a median relative error of 0.0057 at step $0.1\|h\|$ (smaller steps fall below float32 resolution for this head, whose gradient norm is about 0.05), and the first-order prediction of a $0.05B$ step correlates with the realized margin change at $r\geq0.9987$. Free-form generation uses the scorer prompt unchanged, greedy decoding with at most 24 new tokens, and a parser that takes the first digit run or number word after any echoed question; these settings were fixed on development data. The test cohort consists of 500 images newly drawn from the procedural generator of the Llama cohorts, with its collision rule, and has zero seed, pixel, and file overlap with every Llama and \idefics{} cohort. The protocol was fixed before this cohort was evaluated, and it was evaluated once.

\ifaclstyle\begin{table*}[!tp]\else\begin{table}[H]\fi
\centering\fontsize{8.5}{10.5}\selectfont
\caption{Oracle \mea{} in \idefics{} on 470 baseline-wrong examples of the new 500-image cohort, with the Llama oracle cohort for reference. Controls use the realized \mea{} step norms. The activation-aligned control in \idefics{} fixes $P_Uh_0/\|P_Uh_0\|$ at step 0, whereas the Llama control re-evaluates it at every step.}
\label{tab:idefics_mea}
\begin{tabular}{@{}lll@{}}
\toprule
\bfseries Endpoint & \bfseries IDEFICS ($N=470$) & \bfseries Llama ($N=112$) \\
\midrule
\mea{} target reached & \textbf{250 (53.2\%)} & 91 (81.25\%) \\
Fixed direction & 13 (2.8\%); 238:1, $p=5.4\times10^{-70}$ & 17 (15.18\%) \\
Activation-aligned direction & 7 (1.5\%); 243:0, $p=1.4\times10^{-73}$ & 1 (0.89\%) \\
Generated answer correct, edited vs.\ unedited & 218 vs.\ 124; 97:3, $p=2.6\times10^{-25}$ & 80 vs.\ 0 \\
Local-law $r$ at $\beta=0.05/0.10/0.50/1.0/2.0$ & 1.000 / 1.000 / 0.9999 / 0.9995 / 0.9986 & 0.993 / 0.974 / 0.756 / 0.419 / 0.341 \\
Median iterations (all / reached) & 19 / 15 & 9 / 8 \\
\bottomrule
\end{tabular}
\ifaclstyle\end{table*}\else\end{table}\fi

\paragraph{Readouts.}
The unedited model is correct on 30/500 test images under the closed-set proxy. Its free-form answers are more often correct than its proxy prediction, because the teacher-forced scores favor 0 and 1: on development data, the parsed answer matches the proxy argmax on 44/200 examples. Reachability and generated-answer correctness are therefore separate readouts here. On the test cohort, the edited answer equals the final \mea{} proxy prediction on 240/470 examples, and no answer fails to parse.

\paragraph{Local law and displacement.}
Local-law correlations are computed on 200 baseline examples of the test cohort along the projected mean-gradient direction, with bootstrap 95\% CIs inside $[0.998,1.000]$ and regression slopes between 1.001 and 1.021. Edit sizes are not matched across models: at $\beta=2$ the median predicted change of the expected numeral is 0.24, so these values document a wide linear regime of this head rather than a contrast with Llama. Because the trust radius scales with the current-state norm and this head's activation norm is small ($\|h\|\approx0.45$), \idefics{} trajectories travel further than Llama trajectories relative to the unedited state: median net displacement is $6.8B_0$ for reached targets and $14.7B_0$ over all trajectories, with $B_0=3\|h_0\|_2$. \mea{} costs on average 36.3 batched forward and 17.6 backward evaluations per example. The \idefics{} replication uses 6.2 GPU-hours and the whole-head ablation 1.7.

\FloatBarrier
\Needspace{6\baselineskip}
\section{Routing over Direction, Gain, and Geometry}
\label{app:routing}

\subsection{Direction and Gain Candidates}
\label{app:geometry_trajectory_ablation}
The v4 comparison uses the proposal controller and its fixed intervention convention. To separate geometry from gain, we run a development-only common gain-budget audit on the 629 v3 upward-action examples. At reference strength $s=1$, whole-head accuracy is 52.78\% and magnitude-normalized low-rank accuracy is 54.37\%. Independent tuning changes the ordering: whole-head peaks at 64.07\% for $s=2.75$, while MN peaks at 62.80\% for $\beta=1.75$. Gain is therefore a separate calibration variable, distinct from geometry.

A fresh direction-development cohort ($N=500$) then compares activation-aligned MN and sensitivity-aligned SA inside the same fixed Top-4. The projected directions have median cosine 0.7898; 44/500 (8.8\%) are non-positive. On the separately generated prospective cohort, matched-gain accuracy differences SA$-$MN are $+1.8\pp$ at $\beta=0.75$ (9:0 paired discordance, exact $p=0.00391$), $+1.8\pp$ at $\beta=1$ (11:2, $p=0.02246$), and $-1.8\pp$ at $\beta=2$ (0:9, $p=0.00391$). This replicated crossover is the main reason both directions are retained as candidates rather than selecting one globally.

A simpler tempered sensitivity-gain rule is also tested prospectively. Although sensitivity strongly predicts the first gain at which some errors become repairable on development data, the rule reaches 59.6\% on the prospective $N=500$ cohort, below both separately selected MN (61.0\%) and SA (61.6\%). This prospective result motivates candidate utility prediction: the response threshold is informative, and correctness utility is modeled explicitly.

A separate matched-budget $N=500$ audit compares the best observed MN, SA, and WHOLE finite-step accuracies under the same candidate budget. The three peak accuracies are nearly tied, but their optimal gains differ.

\ifaclstyle\begin{table*}[!tp]\else\begin{table}[H]\fi
\centering\fontsize{8.5}{10.5}\selectfont
\caption{Matched-budget peak comparison on the separate $N=500$ geometry audit. Confidence intervals are paired peak-to-peak descriptive intervals.}
\label{tab:geometry_peak_audit}
\begin{tabular}{@{}lrrr@{}}
\toprule
\bfseries Comparison & \bfseries Peak accuracies & \bfseries Difference & \bfseries 95\% CI \\
\midrule
WHOLE $-$ MN & 61.6\% vs 61.4\% & $+0.2\pp$ & $[-1.4,+1.8]\pp$ \\
WHOLE $-$ SA & 61.6\% vs 61.0\% & $+0.6\pp$ & $[-1.0,+2.4]\pp$ \\
MN $-$ SA & 61.4\% vs 61.0\% & $+0.4\pp$ & $[0,+1.0]\pp$ \\
\bottomrule
\end{tabular}
\ifaclstyle\end{table*}\else\end{table}\fi

The routing-relevant $N=160$ upward-proposal development subset provides the full common-grid geometry trajectory. Gains are $0.25,0.5,0.75,1,1.25,1.5,1.75,2,2.25,2.5$ in column order below.

\ifaclstyle\begin{table*}[!tp]\else\begin{table}[H]\fi
\centering\fontsize{8}{9.5}\selectfont
\caption{Matched-gain exact accuracy on the $N=160$ upward-proposal development population. The crossover and non-monotonic tails motivate geometry/gain routing rather than a global geometry claim.}
\label{tab:geometry_trajectory}
\setlength{\tabcolsep}{2.6pt}
\begin{tabular}{@{}lrrrrrrrrrr@{}}
\toprule
& .25 & .50 & .75 & 1.00 & 1.25 & 1.50 & 1.75 & 2.00 & 2.25 & 2.50 \\
\midrule
MN & 36.88 & 40.00 & 45.00 & 54.38 & 56.88 & 59.38 & 61.25 & 63.75 & 63.75 & 60.63 \\
SA & 38.13 & 40.63 & 52.50 & 59.38 & 61.25 & 61.88 & 62.50 & 59.38 & 58.75 & 55.63 \\
WHOLE & 35.00 & 38.75 & 44.38 & 50.63 & 53.13 & 56.88 & 61.88 & 62.50 & 64.38 & 63.13 \\
\bottomrule
\end{tabular}
\ifaclstyle\end{table*}\else\end{table}\fi

\subsection{Candidate Utility Router}
\label{app:utility_router}
\paragraph{Stage 2: Candidate Utility Router.}
For an upward proposal from the Proposal Controller, the
\textbf{Candidate Utility Router} chooses among abstention and a bounded
family of low-rank direction/gain interventions.
The candidate set is
\begin{equation}
\mathcal{C}(x)
=
\{\mathrm{identity}\}
\cup
\left(
\{\mathrm{MN},\mathrm{SA}\}
\times
\mathcal{B}
\right),
\label{eq:candidate_set}
\end{equation}
where
\begin{equation}
\mathcal{B}
=
\{0.5,0.75,1,1.25,1.5,1.75,2\}.
\label{eq:gain_grid}
\end{equation}
The identity candidate applies no second-stage displacement,
$\Delta h_{\mathrm{identity}}=0$.
Thus, each upward proposal has 15 possible second-stage decisions:
identity plus 14 direction/gain interventions.

For each candidate $c\in\mathcal{C}(x)$, a fixed
candidate-conditioned logistic model estimates candidate correctness,
\begin{equation}
Q_\theta(x,c)
=
\Pr\!\left[
\hat y_c(x)=y
\mid
\Phi(x,c)
\right],
\label{eq:utility_score}
\end{equation}
where $\Phi(x,c)$ contains only information available without ground truth
at inference time.

The sample representation contains the 39 fixed Proposal Controller features, the proposal action and proposal score, and six transformed mechanistic features: log projected-gradient norm, log pooled-head norm, log SA sensitivity, projected-gradient/projected-activation cosine, gradient capture, and activation capture. Eight candidate channels encode method, gain, gain squared, and method--gain combinations ($2\times7=14$ low-rank interventions plus identity); interactions with the 47 sample features produce the 431-dimensional logistic design. Training is grouped by sample so candidate rows from one example never split across folds.

The deployed second-stage decision is
\begin{equation}
c^*(x)
=
\arg\max_{c\in\mathcal{C}(x)}
Q_\theta(x,c).
\label{eq:candidate_router}
\end{equation}
Deterministic ties are resolved in favor of identity, then smaller gain, then
MN before SA.

Ground truth is used only to fit and evaluate the Candidate Utility Router;
it is never used to construct inference-time features or to select a
candidate at deployment.
Because Stage~2 is invoked only when $\alpha_{\rm prop}>1$, all evaluation
Candidate Utility Router results are conditional on the corresponding
upward-proposal population.
Accordingly, the final routing experiment evaluates second-stage refinement
of fixed upward proposals rather than end-to-end routing over all examples
and intervention directions.

\paragraph{Candidate Utility Router development and fixed external validation.}
The Candidate Utility Router is developed only on the 160 upward-proposal
examples selected by the Proposal Controller from the direction-development cohort.
Grouped five-fold out-of-fold evaluation keeps all candidate rows from the
same example in the same fold.

The fold-selected global candidate reaches 62.50\% accuracy, whereas the
Candidate Utility Router reaches 67.50\%, a $+5.00\pp$ improvement with
paired bootstrap 95\% CI $[+1.25,+9.375]\pp$ and exact McNemar
$p=0.03857$.

The Utility Router feature-family audit is performed only on grouped OOF development. It is interpretive rather than confirmatory: no ablation is used to refit or modify the external/final policy.

\ifaclstyle\begin{table*}[!tp]\else\begin{table}[H]\fi
\centering\fontsize{8.5}{10.5}\selectfont
\caption{Grouped-OOF Candidate Utility Router feature-family ablation on the $N=160$ development population.}
\label{tab:router_feature_ablation}
\begin{tabular}{@{}lr@{}}
\toprule
\bfseries Feature configuration & \bfseries OOF accuracy \\
\midrule
FULL & 67.50\% \\
NO\_INTERNAL & 66.875\% \\
NO\_OUTPUT\_GEOMETRY & 65.625\% \\
NO\_PROPOSAL & 65.00\% \\
BASELINE+PROPOSAL & 65.00\% \\
OUTPUT\_GEOM+PROPOSAL & 65.00\% \\
INTERNAL\_MECH+PROPOSAL & 63.125\% \\
\bottomrule
\end{tabular}
\ifaclstyle\end{table*}\else\end{table}\fi

A complementary feature-group permutation audit produces accuracy drops of 14.481\pp{} for baseline-state features, 4.525\pp{} for proposal context, 4.223\pp{} for internal-mechanistic features, and 3.948\pp{} for output-geometry features. The baseline state is the strongest individual information source, but every mechanistic/action-context group contributes measurable complementary information to the full representation.

After model fitting and candidate design are fixed, the router is evaluated
on a separately generated prospective cohort containing 156 upward-proposal
examples.
The global \mncsa{} $\beta=2$ comparator reaches 66.67\%, while the
Candidate Utility Router reaches 71.15\%, a $+4.487\pp$ difference with
bootstrap 95\% CI $[+0.641,+8.974]\pp$.
The corresponding exact McNemar test gives $p=0.06543$.
Because this cohort had previously supported a different prospective
sensitivity-only experiment, we treat this stage as \emph{fixed external
validation}, not held-out confirmation.

\ifaclstyle\begin{table*}[!tp]\else\begin{table}[H]\fi
\centering
\fontsize{8.5}{10.5}\selectfont
\caption{Utility Router evidence chain. The final row is the only held-out confirmatory evaluation. Accuracy is measured on the prespecified upward-action population of each stage.}
\label{tab:utility_router_chain_app}
\begin{tabular}{@{}lrrrrl@{}}
\toprule
\bfseries Stage & \bfseries $N$ & \bfseries Global MN & \bfseries Router & \bfseries $\Delta$ & \bfseries Evidence level \\
\midrule
direction-development grouped OOF & 160 & 62.50\% & 67.50\% & $+5.00\pp$ & Development / OOF \\
Prospective fixed external & 156 & 66.67\% & 71.15\% & $+4.487\pp$ & Prespecified external \\
Utility-final held-out & 294 & 60.54\% & 66.33\% & $+5.782\pp$ & Confirmatory \\
\bottomrule
\end{tabular}
\ifaclstyle\end{table*}\else\end{table}\fi

\paragraph{Held-out upward-proposal confirmation.}
Before the final cohort is generated, the Proposal Controller, candidate
family, feature map, Candidate Utility Router, fixed global comparator,
primary endpoint, and statistical test are all fixed.
A new $N=1000$ procedural cohort is then generated with zero seed and exact
image-hash overlap with the earlier procedural, direction-development, and prospective
cohorts.
The Proposal Controller produces 294 upward proposals, defining the
prespecified primary population for the second-stage confirmation.

On these $N=294$ examples, the global \mncsa{} $\beta=2$ comparator
reaches $178/294=60.54\%$, whereas the Candidate Utility Router reaches
$195/294=66.33\%$.
The paired improvement is \textbf{$+5.782\pp$}, with paired bootstrap
95\% CI $[+3.401,+8.503]\pp$ and two-sided exact McNemar
$p=1.53\times10^{-5}$.
The proposal controller's 294 upward examples comprise 12 with $\alpha=1.5$, 24 with $\alpha=2$, and 258 with $\alpha=4$; candidate outcomes are generated for all 14 non-identity candidates and fixed before router evaluation, yielding 4,116 candidate rows.

The improvement reflects both additional repairs and fewer destructive
interventions.
Relative to identity, the global comparator produces 132 repairs and
27 breaks, whereas the Candidate Utility Router produces 138 repairs and
16 breaks.
All 17 paired discordant outcomes favor the Candidate Utility Router
(17 router-only versus 0 comparator-only).

The final policy also does not collapse to one intervention.
Across the 294 upward proposals, it selects identity on 60 examples,
MN on 134, and SA on 100, and uses every nonzero gain in the candidate grid.
Together with the preceding direction--gain crossover, this held-out result
supports the bounded policy claim:
even after a coarse cardinality action has been proposed, deciding
\emph{whether}, \emph{along which direction}, and \emph{how strongly} to
intervene remains an example-dependent utility problem.

\ifaclstyle\begin{table*}[!tp]\else\begin{table}[H]\fi
\centering
\fontsize{8.5}{10.5}\selectfont
\caption{Held-out Utility Router final confirmation on the $N=294$ prespecified upward population.}
\label{tab:utility_router_final_app}
\begin{tabular}{@{}lrrrr@{}}
\toprule
\bfseries Policy & \bfseries Correct & \bfseries Accuracy & \bfseries Repairs & \bfseries Breaks \\
\midrule
Identity & 73 & 24.83\% & -- & -- \\
Fixed global MN $\beta=2$ & 178 & 60.54\% & 132 & 27 \\
Fixed Utility Router & 195 & 66.33\% & 138 & 16 \\
Candidate oracle & 217 & 73.81\% & -- & -- \\
\bottomrule
\end{tabular}
\ifaclstyle\end{table*}\else\end{table}\fi

Relative to the global comparator, the router captures 17 of the 39 available oracle-headroom examples, or 43.6\% of the remaining gap to the candidate oracle. Gain selections are $\beta=0$ (identity) 60, 0.5: 29, 0.75: 7, 1: 16, 1.25: 20, 1.5: 18, 1.75: 22, and 2: 122.

\ifaclstyle\begin{figure*}[!tp]\else\begin{figure}[!htbp]\fi
\centering

\pgfplotsset{uraxis/.style={scale only axis,width=\dimexpr\linewidth-1.25cm\relax,
at={(0,0)},anchor=south west,
height=3.15cm,axis x line*=bottom,axis y line*=left,
axis line style={draw=aromaInk!60,line width=.45pt},tick style={draw=aromaInk!60},
tick align=outside,major tick length=2pt,
tick label style={font=\rmfamily\fontsize{7.5}{9}\selectfont,text=aromaInk},
label style={font=\rmfamily\fontsize{7.5}{9}\selectfont,text=aromaInk},
ymajorgrids=true,grid style={draw=aromaGrid,line width=.35pt},scaled ticks=false,clip=true}}

\begin{minipage}[t]{.48\textwidth}\vspace{0pt}\centering
\begin{tikzpicture}
\begin{axis}[uraxis,ybar,bar width=7pt,xmin=-.6,xmax=2.6,ymin=55,ymax=74,
ylabel={Accuracy (\%)},xtick={0,1,2},xticklabels={OOF,External,Final},xticklabel style={font=\rmfamily\fontsize{7.5}{9}\selectfont},ytick={55,60,65,70},
legend style={font=\rmfamily\fontsize{7.5}{9}\selectfont,draw=none,fill=none,at={(.5,1.03)},anchor=south,legend columns=1,/tikz/every even column/.append style={column sep=5pt}},legend image code/.code={\path[#1] (0pt,-2pt) rectangle (8pt,3pt);}]
\addplot[draw=none,fill=aromaGray,bar shift=-4pt,forget plot] coordinates {(0,62.50)(1,66.67)(2,60.54)};
\addplot[draw=none,fill=aromaGreen,bar shift=4pt,forget plot] coordinates {(0,67.50)(1,71.15)(2,66.33)};
\addlegendimage{area legend,fill=aromaGray,draw=none}
\addlegendentry{Global comparator}
\addlegendimage{area legend,fill=aromaGreen,draw=none}
\addlegendentry{Utility Router}
\node[font=\rmfamily\fontsize{7.5}{9}\selectfont\bfseries,text=aromaGreen!75!black,anchor=south] at (axis cs:2,67.2) {$+5.78$ pp};
\end{axis}

\end{tikzpicture}\par\smallskip
{\rmfamily\fontsize{8}{9.5}\selectfont\bfseries (a) Routing advantage across evaluation stages\par}
\end{minipage}\hfill\begin{minipage}[t]{.48\textwidth}\vspace{0pt}\centering
\begin{tikzpicture}
\begin{axis}[uraxis,ybar,bar width=10pt,xmin=-.6,xmax=1.6,ymin=0,ymax=165,
ylabel={Examples},xtick={0,1},xticklabels={Global MN,Router},ytick={0,50,100,150},
legend style={font=\rmfamily\fontsize{7.5}{9}\selectfont,draw=none,fill=none,at={(.5,1.03)},anchor=south,legend columns=1,/tikz/every even column/.append style={column sep=5pt}},legend image code/.code={\path[#1] (0pt,-2pt) rectangle (8pt,3pt);}]
\addplot[draw=none,fill=aromaGreen,bar shift=-6pt,forget plot,nodes near coords,every node near coord/.append style={font=\rmfamily\fontsize{7.5}{9}\selectfont,text=aromaInk,yshift=1pt}] coordinates {(0,132)(1,138)};
\addplot[draw=none,fill=aromaOrange,bar shift=6pt,forget plot,nodes near coords,every node near coord/.append style={font=\rmfamily\fontsize{7.5}{9}\selectfont,text=aromaInk,yshift=1pt}] coordinates {(0,27)(1,16)};
\addlegendimage{area legend,fill=aromaGreen,draw=none}\addlegendentry{Repairs}
\addlegendimage{area legend,fill=aromaOrange,draw=none}\addlegendentry{Breaks}

\end{axis}

\end{tikzpicture}\par\smallskip
{\rmfamily\fontsize{8}{9.5}\selectfont\bfseries (b) Final repairs increase while breaks fall\par}
\end{minipage}\par\vspace{14pt}\begin{minipage}[t]{.48\textwidth}\vspace{0pt}\centering
\begin{tikzpicture}
\begin{axis}[uraxis,ybar,bar width=15pt,bar shift=0pt,xmin=-.6,xmax=2.6,ymin=0,ymax=150,
ylabel={Selections},xtick={0,1,2},xticklabels={Identity,MN,SA},ytick={0,50,100,150},
nodes near coords,every node near coord/.append style={font=\rmfamily\fontsize{7.5}{9}\selectfont\bfseries,anchor=south,yshift=1pt},legend image code/.code={\path[#1] (0pt,-2pt) rectangle (8pt,3pt);}]
\addplot[draw=none,fill=aromaGray] coordinates {(0,60)};
\addplot[draw=none,fill=aromaBlue] coordinates {(1,134)};
\addplot[draw=none,fill=aromaOrange] coordinates {(2,100)};
\end{axis}

\end{tikzpicture}\par\smallskip
{\rmfamily\fontsize{8}{9.5}\selectfont\bfseries (c) The router uses all three decision types\par}
\end{minipage}\hfill\begin{minipage}[t]{.48\textwidth}\vspace{0pt}\centering
\begin{tikzpicture}
\begin{axis}[uraxis,ybar,bar width=5pt,bar shift=0pt,xmin=-.2,xmax=2.2,ymin=0,ymax=135,
xlabel={Selected $\beta$},ylabel={Selections},xtick={0,.5,.75,1,1.25,1.5,1.75,2},
xticklabels={0,.5,.75,1,1.25,1.5,1.75,2},xticklabel style={font=\rmfamily\fontsize{7.5}{9}\selectfont,rotate=45,anchor=north east},ytick={0,40,80,120},legend image code/.code={\path[#1] (0pt,-2pt) rectangle (8pt,3pt);}]
\addplot[draw=none,fill=aromaBlue] coordinates {(0,60)(.5,29)(.75,7)(1,16)(1.25,20)(1.5,18)(1.75,22)(2,122)};
\end{axis}

\end{tikzpicture}\par\smallskip
{\rmfamily\fontsize{8}{9.5}\selectfont\bfseries (d) Adaptive gain selection\par}
\end{minipage}
\par\smallskip{\rmfamily\fontsize{7.5}{9}\selectfont Paired wins: 17 router-only / 0 global-only\par}
\caption{\textbf{Utility-aware routing confirms adaptive control within the low-rank candidate family.}
\textbf{(a)} The router improves over the fold-selected global comparator in grouped OOF development and the global comparator in external validation and the held-out final population; only the last stage is the primary confirmatory test. \textbf{(b)} On the final $N=294$, the advantage combines six additional repairs with eleven fewer breaks, yielding 17 router-only versus 0 global-only paired wins. \textbf{(c--d)} The learned policy does not collapse to a single method or gain: it selects identity, MN, and SA and uses every nonzero gain in the grid. Final primary effect: $+5.782\pp$, 95\% CI $[+3.401,+8.503]\pp$, exact McNemar $p=1.53\times10^{-5}$.}
\label{fig:utility_router}
\ifaclstyle\end{figure*}
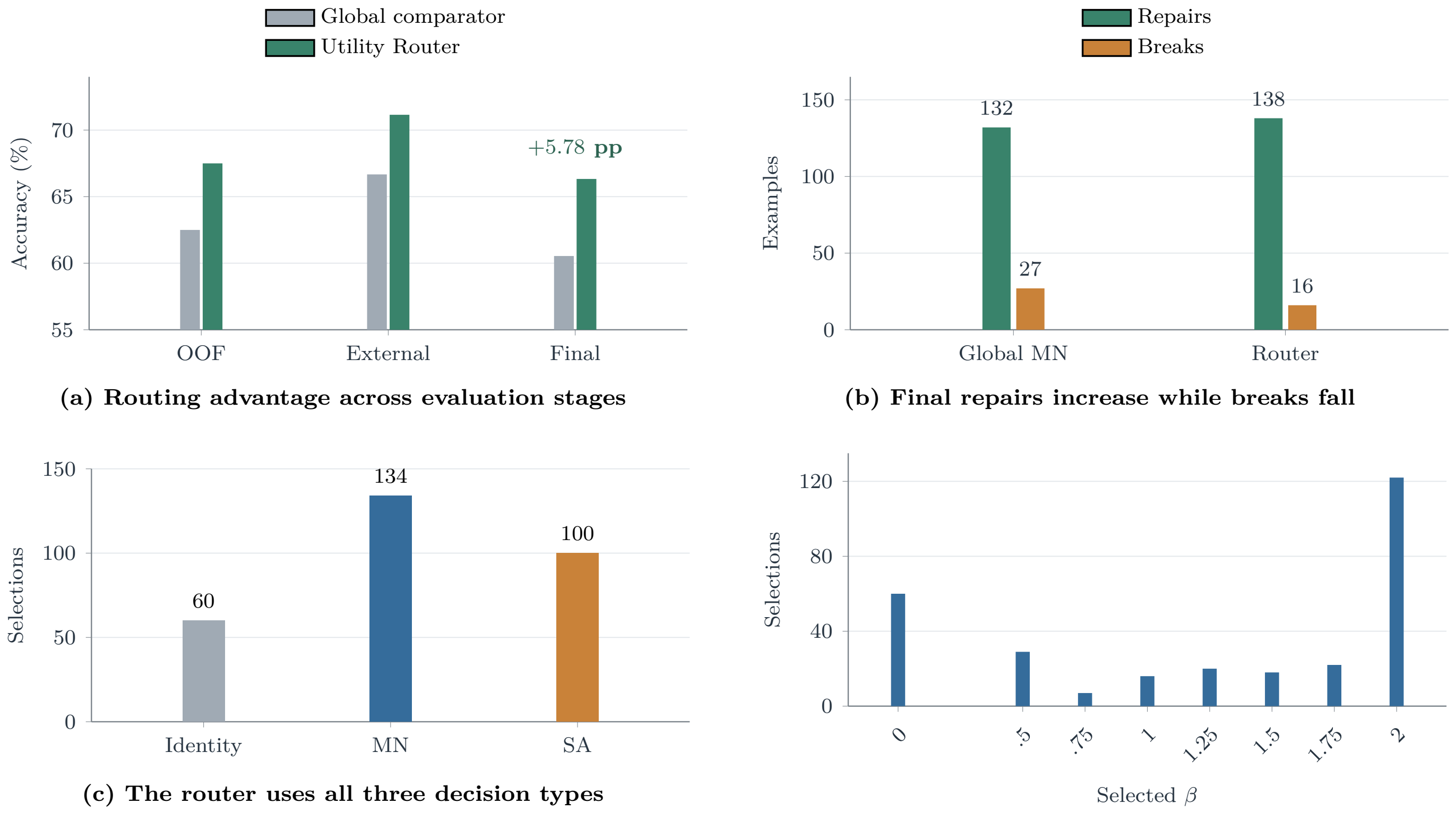\else\end{figure}\fi

\subsection{Whole-Head Robustness and Matched-Budget Geometry Diagnostics}
\label{app:geometry_diagnostics}
The original Utility Router confirmation is intentionally interpreted within its fixed low-rank candidate family.
After that $N=294$ final population had already been observed, we ran a separate diagnostic in which whole-head intervention received its own per-example gain router trained only on direction-development data.
The adaptive whole-head router reaches $208/294=70.75\%$, compared with $195/294=66.33\%$ for the original low-rank router.
The post hoc difference is $+4.422\pp$ with paired bootstrap 95\% CI $[+1.020,+7.823]\pp$ and exact McNemar $p=0.01916$ (20 whole-only versus 7 low-rank-only outcomes).
The adaptive whole policy produces 150 repairs and 15 breaks versus 138 repairs and 16 breaks for the low-rank router.
Because the final population had already been analysed, this comparison is a post hoc robustness diagnostic.

To separate geometry from candidate budget and router capacity, a second post hoc experiment gives MN and whole-head families identical eight-candidate budgets (identity plus seven gains), identical 47 sample features, identical 191-dimensional generic candidate-conditioned router design, and matched displacement norm $g|\alpha-1|\|h\|$.
The result does not yield a stable intrinsic geometry ranking:
\ifaclstyle\begin{table*}[!tp]\else\begin{table}[H]\fi
\centering\fontsize{8.5}{10.5}\selectfont
\caption{Matched-budget MN versus whole-head diagnostic. Positive $\Delta$ denotes Whole$-$MN. Neither population supports a stable geometry ordering.}
\label{tab:matched_budget_geometry}
\begin{tabular}{@{}lrrrrr@{}}
\toprule
\bfseries Population & \bfseries $N$ & \bfseries MN & \bfseries Whole & \bfseries $\Delta$ & \bfseries 95\% CI \\
\midrule
Development OOF & 160 & 66.25\% & 62.50\% & $-3.75\pp$ & $[-8.75,+1.25]$ \\
Final diagnostic & 294 & 65.65\% & 66.67\% & $+1.02\pp$ & $[-1.70,+4.08]$ \\
\bottomrule
\end{tabular}
\ifaclstyle\end{table*}\else\end{table}\fi
Exact McNemar $p$-values are 0.2379 and 0.6476, respectively.
Geometry, gain range, and routing therefore interact, and no single family is intrinsically superior.
This diagnostic directly motivates prospective unified geometry routing.

\subsection{Unified Geometry Router}
\label{app:ugr}
UGR uses the Stage-1 Proposal Controller and invokes Stage~2 only for $\alpha_{\rm prop}>1$.
For each invoked sample, the candidate set is identity plus MN, SA, and WHOLE crossed with gains $\{0.5,0.75,1,1.25,1.5,1.75,2,\allowbreak 2.25,2.5,2.75,3,3.25,3.5,4\}$, giving 43 candidates per sample.
The 47 sample features are the 39 fixed Proposal Controller features, selected proposal action/score, and six mechanistic summaries.
Nine candidate channels plus their interactions yield 479 design features.
All routers use StandardScaler plus L2 LogisticRegression with grouped five-fold splitting by sample.
MN and SA retain the Top-4 directions, with displacement magnitude $\gamma(\alpha_{\rm prop}-1)\|h\|_2$ for gain $\gamma$ in the grid
\begin{equation}
\begin{aligned}
\mathcal{G}=\{&0.5,0.75,1,1.25,1.5,1.75,2,\\
              &2.25,2.5,2.75,3,3.25,3.5,4\}.
\end{aligned}
\label{eq:ugr_gain_grid}
\end{equation}
For WHOLE, the matched finite-gain rule is $\alpha_{\rm whole}=1+\gamma(\alpha_{\rm prop}-1)$, so $\gamma$ controls the same proposal-relative scaling variable across geometry families. The design therefore has $47+9+47\times9=479$ dimensions, and deterministic ties are broken toward IDENTITY, smaller gain, MN, SA, then WHOLE.

\ifaclstyle\begin{table*}[!tp]\else\begin{table}[H]\fi
\centering\fontsize{8.5}{10.5}\selectfont
\caption{UGR development OOF and held-out final results. The selected restricted comparator is whole-head restricted because restricted accuracies tie in development but whole-head has fewer breaks.}
\label{tab:ugr_full}
\begin{tabular}{@{}lrrrrrr@{}}
\toprule
\bfseries Stage / policy & \bfseries $N$ & \bfseries Correct & \bfseries Accuracy & \bfseries Repairs & \bfseries Breaks & \bfseries Oracle \\
\midrule
Dev low-rank restricted & 654 & 450 & 68.81\% & 290 & 37 & 80.58\% \\
Dev whole restricted & 654 & 450 & 68.81\% & 287 & 34 & 80.58\% \\
Dev unified & 654 & 470 & 71.87\% & 305 & 32 & 83.18\% \\
\midrule
Final identity & 633 & 195 & 30.81\% & -- & -- & -- \\
Final whole restricted & 633 & 449 & 70.93\% & 282 & 28 & 80.88\% \\
Final unified & 633 & 463 & \textbf{73.14\%} & 298 & 30 & \textbf{84.36\%} \\
\bottomrule
\end{tabular}
\ifaclstyle\end{table*}\else\end{table}\fi

The development unified-minus-selected-restricted difference is $+3.058\pp$.
On held-out final data, unified-minus-whole is $+2.212\pp$ with 95\% paired bootstrap CI $[+0.316,+4.265]\pp$, exact McNemar $p=0.0385$, and discordance 27 unified-only versus 13 whole-only.
The final selection distribution is WHOLE 311, MN 214, SA 43, and IDENTITY 65. These frequencies are descriptive.
The full-head restricted router selects WHOLE on 549 and IDENTITY on 84.
The final fixed grid is used without post hoc extension. All $633\times43=27{,}219$ candidate outcomes were generated once and checked for sample order, candidate order, and family counts before evaluation; no outcome, feature, router, comparator, or candidate family was modified.

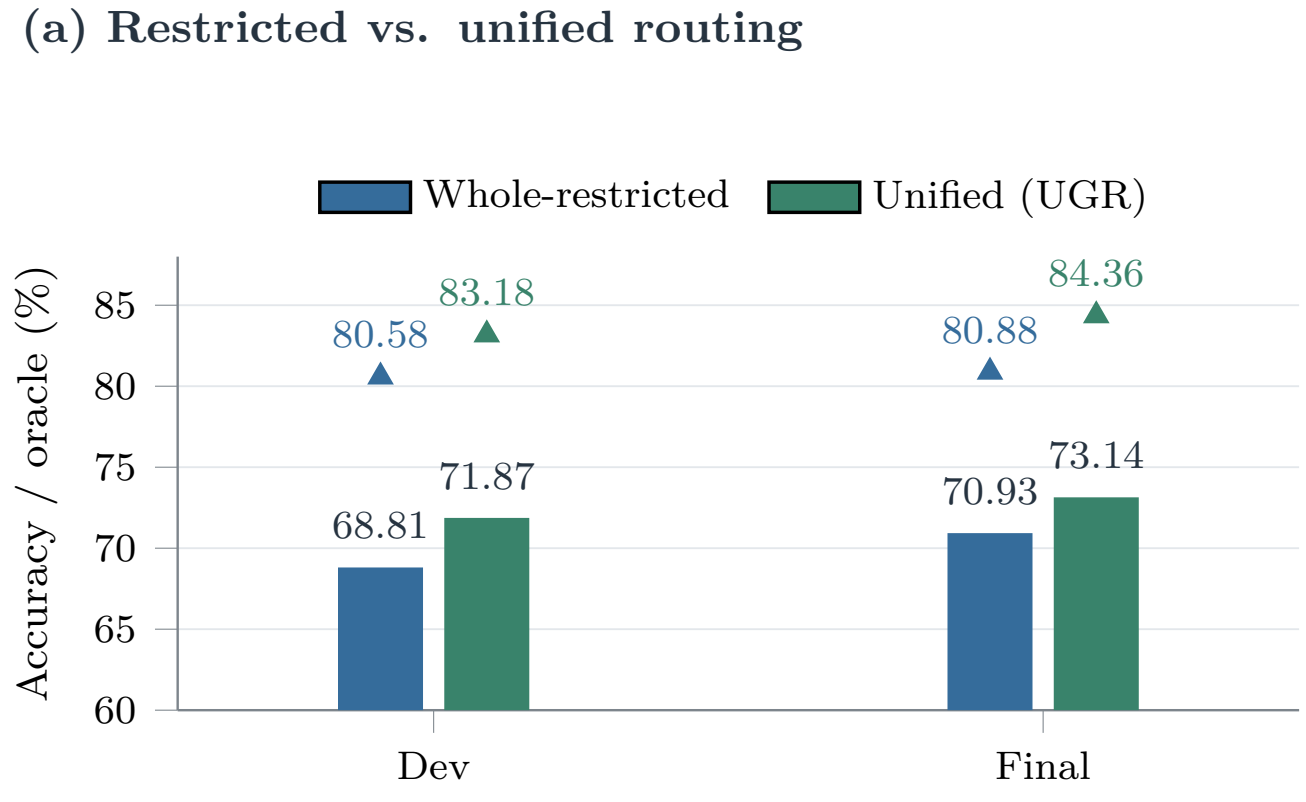
\ifaclstyle\begin{figure*}[!tp]\else\begin{figure}[!htbp]\fi
\centering

\pgfmathsetlengthmacro{\ugrHalf}{0.5*\textwidth}
\pgfmathsetlengthmacro{\ugrWidth}{0.5*\textwidth-1.65cm}
\pgfmathsetlengthmacro{\ugrRight}{\ugrHalf+1.15cm}
\begin{tikzpicture}
\tikzset{ugrtitle/.style={anchor=north west,inner sep=0pt,
  text width=\dimexpr.5\textwidth-8pt\relax,align=left,font=\rmfamily\fontsize{8.5}{10.5}\selectfont\bfseries,text=aromaInk}}
\pgfplotsset{ugraxis/.style={scale only axis,width=\ugrWidth,height=3.0cm,
  anchor=north west,axis x line*=bottom,axis y line*=left,tick align=outside,
  axis line style={draw=aromaInk!60,line width=.45pt},tick style={draw=aromaInk!60},
  tick label style={font=\rmfamily\fontsize{7.5}{9}\selectfont},
  label style={font=\rmfamily\fontsize{8}{9.5}\selectfont},
  ymajorgrids=true,grid style={draw=aromaGrid,line width=.35pt},
  legend style={font=\rmfamily\fontsize{7.5}{9}\selectfont,draw=none,fill=none,
    at={(.5,1.04)},anchor=south,legend columns=2,
    /tikz/every even column/.append style={column sep=5pt}}}}
\node[ugrtitle] at (0,0) {(a) Restricted vs. unified routing};
\node[ugrtitle] at (\ugrHalf,0) {(b) The unified policy uses multiple\\intervention families};
\begin{axis}[ugraxis,at={(1.0cm,-1.65cm)},
  ymin=60,ymax=88,ylabel={Accuracy / oracle (\%)},
  symbolic x coords={Dev,Final},xtick=data,ytick={60,65,70,75,80,85},
  ybar,bar width=16pt,enlarge x limits=0.42,
  nodes near coords,every node near coord/.append style={
    font=\rmfamily\fontsize{7.5}{9}\selectfont,text=aromaInk,anchor=south,yshift=2pt},legend image code/.code={\path[#1] (0pt,-2pt) rectangle (8pt,3pt);}]
\addplot[forget plot,fill=aromaBlue,draw=none,bar shift=-10pt] coordinates {(Dev,68.81)(Final,70.93)};
\addlegendimage{area legend,fill=aromaBlue,draw=none}\addlegendentry{Whole-restricted}
\addplot[forget plot,fill=aromaGreen,draw=none,bar shift=10pt] coordinates {(Dev,71.87)(Final,73.14)};
\addlegendimage{area legend,fill=aromaGreen,draw=none}\addlegendentry{Unified (UGR)}
\addplot[forget plot,only marks,mark=triangle*,mark size=2.5pt,
  draw=aromaBlue,fill=aromaBlue,bar shift=0pt,xshift=-10pt,
  every node near coord/.append style={text=aromaBlue,font=\rmfamily\fontsize{7.5}{9}\selectfont}]
  coordinates {(Dev,80.58)(Final,80.88)};
\addplot[forget plot,only marks,mark=triangle*,mark size=2.5pt,
  draw=aromaGreen,fill=aromaGreen,bar shift=0pt,xshift=10pt,
  every node near coord/.append style={text=aromaGreen,font=\rmfamily\fontsize{7.5}{9}\selectfont}]
  coordinates {(Dev,83.18)(Final,84.36)};
\end{axis}
\begin{axis}[ugraxis,at={(\ugrRight,-1.65cm)},
  ymin=0,ymax=100,ylabel={Selection share (\%)},
  symbolic x coords={Dev-R,Dev-U,Final-R,Final-U},xtick=data,
  xticklabels={Dev-R,Dev-U,Final-R,Final-U},ytick={0,20,40,60,80,100},
  ybar stacked,bar width=21pt,enlarge x limits=0.16,legend style={legend columns=2},legend image code/.code={\path[#1] (0pt,-2pt) rectangle (8pt,3pt);}]
\addplot[forget plot,fill=aromaPurple,draw=none] coordinates {(Dev-R,87.61)(Dev-U,54.89)(Final-R,86.73)(Final-U,49.13)};
\addlegendimage{area legend,fill=aromaPurple,draw=none}\addlegendentry{WHOLE}
\addplot[forget plot,fill=aromaBlue,draw=none] coordinates {(Dev-R,0)(Dev-U,30.43)(Final-R,0)(Final-U,33.81)};
\addlegendimage{area legend,fill=aromaBlue,draw=none}\addlegendentry{MN}
\addplot[forget plot,fill=aromaOrange,draw=none] coordinates {(Dev-R,0)(Dev-U,5.96)(Final-R,0)(Final-U,6.79)};
\addlegendimage{area legend,fill=aromaOrange,draw=none}\addlegendentry{SA}
\addplot[forget plot,fill=aromaGray,draw=none] coordinates {(Dev-R,12.39)(Dev-U,8.72)(Final-R,13.27)(Final-U,10.27)};
\addlegendimage{area legend,fill=aromaGray,draw=none}\addlegendentry{IDENTITY}
\end{axis}
\node[anchor=north west,inner sep=0pt,text width=\dimexpr.5\textwidth-8pt\relax,align=left,
  font=\rmfamily\fontsize{7.5}{9}\selectfont,text=aromaInk] at (0,-5.35cm)
  {Triangles are candidate-oracle upper bounds. Dev $\Delta=+3.06\pp$; final $\Delta=+2.21\pp$.};
\node[anchor=north west,inner sep=0pt,text width=\dimexpr.5\textwidth-8pt\relax,align=left,
  font=\rmfamily\fontsize{7.5}{9}\selectfont,text=aromaInk] at (\ugrHalf,-5.35cm)
  {R/U denote restricted/unified. Selection shares are descriptive: they show that the unified policy does not degenerate to a single family.};
\end{tikzpicture}

\caption{\textbf{Prospective unified geometry routing improves the strongest development-selected restricted policy.}
\textbf{(a)} Development OOF and held-out-final exact accuracy for the prospectively selected full-head restricted comparator and UGR; triangles show candidate-oracle upper bounds. The paired final gain is $+2.212\pp$ (95\% CI $[+0.316,+4.265]\pp$; exact McNemar $p=0.0385$), with 27 unified-only versus 13 restricted-only correct examples. \textbf{(b)} Selection shares show that the unified router uses the prespecified intervention families non-degenerately. These shares are descriptive; the primary claim is the comparison in (a).}
\label{fig:ugr_main}
\ifaclstyle\end{figure*}\else\end{figure}\fi
\FloatBarrier
\Needspace{6\baselineskip}
\section{Natural-Image Transfer and the HoloCount Boundary}
\label{app:natural}

\subsection{Geometry Transfer, Zero-Shot Policy Failure, and Recalibration}
\paragraph{Geometry transfer.}
We first test whether the Llama low-rank control geometry discovered on procedural data remains useful on natural images. On an image-disjoint 4,000-example TallyQA confirmation set, the rank-4 basis and \mncsa{} intervention rule are transferred without natural-domain refitting. Using the same fixed natural-domain Proposal Controller for both whole-head and low-rank actuation, baseline exact-count accuracy is 64.975\%, whole-head reaches 67.075\%, and \mncsa{} reaches 66.375\%. Fixed low-rank geometry thus retains positive natural-domain utility ($+1.40\pp$ over baseline; 95\% CI $[+0.875,+1.925]\pp$; \mncsa{} minus same-run whole-head: $-0.70\pp$, 95\% CI $[-1.15,-0.25]\pp$), establishing that the procedurally discovered geometry transfers to natural images without refitting.

\paragraph{Zero-shot policy transfer.}
On a separate fixed 4,000-example TallyQA set, the synthetic Proposal Controller decreases exact-count accuracy from 64.85\% to 61.875\% ($-2.975\pp$), producing 51 repairs and 170 breaks. Yet a per-example oracle over the same actuator and coarse action set reaches 71.575\% ($+6.725\pp$ over baseline), showing that substantial conditional repair capacity remains even when the transferred policy is harmful. Post hoc diagnostics show that this failure coincides with a reversal in the natural error regime, loss of score--utility alignment, and feature-support shift. Support clipping changes controller decisions but does not recover accuracy, indicating a conditional-utility shift rather than a single preprocessing artifact.

\paragraph{Natural-domain recalibration.}
Keeping the actuator, coarse action set, feature family, model class, and regularization fixed, we relearn only the first-stage utility mapping. After fixing a natural-domain Proposal Controller calibrated on $K=500$ natural examples, evaluation on a new 4,000-example question-, image-, and image-ID-disjoint TallyQA set improves exact-count accuracy from 64.925\% to 67.050\% ($+2.125\pp$; 95\% CI $[+1.50,+2.775]\pp$; exact McNemar $p=9.18\times10^{-11}$), with 130 repairs and 45 breaks. The action distribution is selective: 3,200/4,000 examples (80\%) receive identity. This sparsity is a property of the learned policy.

\ifaclstyle\begin{table*}[!tp]\fi
\noindent\begin{minipage}[t]{.487\linewidth}\vspace{0pt}
\begingroup\captionsetup{type=table}
\centering\fontsize{8.5}{10.5}\selectfont\setlength{\tabcolsep}{3pt}
\caption{Natural fixed actions on the 4,000-example TallyQA development/forensic sample.}
\label{tab:naturalfixed}
\begin{tabularx}{\linewidth}{@{}Yrr@{}}
\toprule
\bfseries Action & \bfseries Accuracy & \multicolumn{1}{P{2.2cm}@{}}{\bfseries Change from baseline} \\
\midrule
Baseline / $\alpha=1$ & 64.85\% & 0.00\pp \\
$\alpha=0$ & 67.20\% & $+2.35\pp$ \\
$\alpha=1.5$ & 63.10\% & $-1.75\pp$ \\
$\alpha=2$ & 62.15\% & $-2.70\pp$ \\
$\alpha=4$ & 57.725\% & $-7.125\pp$ \\
Per-example oracle & 71.575\% & $+6.725\pp$ \\
\bottomrule
\end{tabularx}
\endgroup\par\smallskip
Among the 269 oracle-repairable natural errors, the closest successful action is $\alpha=0$ for 149 cases, $\alpha=1.5$ for 19, $\alpha=2$ for 32, and $\alpha=4$ for 69.
The synthetic controller captures only 51/269 (18.96\%) of these opportunities.
\end{minipage}\hfill\begin{minipage}[t]{.487\linewidth}\vspace{0pt}
\begingroup\captionsetup{type=table}
\centering\fontsize{8.5}{10.5}\selectfont\setlength{\tabcolsep}{3pt}
\caption{Controller behavior across domains.}
\label{tab:domains}
\begin{tabularx}{\linewidth}{@{}YYYY@{}}
\toprule
\bfseries Domain & \bfseries Accuracy change & \bfseries Intervention rate & \bfseries Score--utility Spearman \\
\midrule
v2 OOF development & $+8.30\pp$ & 33.30\% & $\approx+0.409$ \\
v3 fixed confirmation & $+8.10\pp$ & 32.65\% & $\approx+0.411$ \\
TallyQA fixed zero-shot & $-2.975\pp$ & 65.275\% & $\approx-0.0667$ \\
\bottomrule
\end{tabularx}
\endgroup\par\smallskip
\end{minipage}\par
\ifaclstyle\end{table*}\fi

\ifaclstyle\begin{table*}[!tp]\else\begin{table}[H]\fi
\centering
\begin{minipage}[t]{0.72\textwidth}
\vspace{0pt}
\captionsetup{justification=raggedright}
\setlength{\tabcolsep}{3pt}
\centering\fontsize{8.5}{10.5}\selectfont
\caption{Final natural confirmation by subset.}
\label{tab:finalsubset}
\vspace{2pt}
\begin{tabular*}{\linewidth}{@{\extracolsep{\fill}}lrrrrrr@{}}
\toprule
\bfseries Subset & \bfseries $N$ & \bfseries Baseline & \bfseries Post & \bfseries $\Delta$ & \bfseries Repairs & \bfseries Breaks \\
\midrule
Simple & 2000 & 78.05\% & 79.15\% & $+1.10\pp$ & 49 & 27 \\
Complex & 2000 & 51.80\% & 54.95\% & $+3.15\pp$ & 81 & 18 \\
Overall & 4000 & 64.925\% & 67.050\% & $+2.125\pp$ & 130 & 45 \\
\bottomrule
\end{tabular*}
\end{minipage}\hfill
\begin{minipage}[t]{0.25\textwidth}
\vspace{0pt}
\captionsetup{justification=raggedright}
\setlength{\tabcolsep}{3pt}
\centering\fontsize{8.5}{10.5}\selectfont
\caption{Final natural-controller action distribution.}
\label{tab:finalactions}
\begin{tabular*}{\linewidth}{@{\extracolsep{\fill}}rrr@{}}
\toprule
\bfseries $\alpha$ & \bfseries Count & \bfseries Fraction \\
\midrule
0 & 508 & 12.70\% \\
1 & 3200 & 80.00\% \\
1.5 & 32 & 0.80\% \\
2 & 122 & 3.05\% \\
4 & 138 & 3.45\% \\
\bottomrule
\end{tabular*}
\end{minipage}
\ifaclstyle\end{table*}\else\end{table}\fi

\ifaclstyle\begin{figure*}[!tp]\else\begin{figure}[!htbp]\fi
\centering
\pgfmathsetlengthmacro{\trHalf}{0.5*\textwidth}
\pgfmathsetlengthmacro{\trWidth}{0.5*\textwidth-1.72cm}
\pgfmathsetlengthmacro{\trRight}{\trHalf+1.12cm}
\ifaclstyle\resizebox{\linewidth}{!}{\fi
\begin{tikzpicture}
\pgfplotsset{traxis/.style={scale only axis,width=\trWidth,height=2.20cm,anchor=north west,
axis x line*=bottom,axis y line*=left,axis line style={draw=aromaInk!60,line width=.45pt},tick style={draw=aromaInk!60,line width=.4pt},
tick align=outside,major tick length=2pt,tick label style={font=\rmfamily\fontsize{7.5}{9}\selectfont,text=aromaInk},
label style={font=\rmfamily\fontsize{7.5}{9}\selectfont,text=aromaInk},ymajorgrids=true,grid style={draw=aromaGrid,line width=.35pt},scaled ticks=false,clip=true}}
\tikzset{trtitle/.style={anchor=west,inner sep=0pt,text=aromaInk,font=\rmfamily\fontsize{8}{9.5}\selectfont\bfseries}}

\node[trtitle] at (0,0) {(a) Fixed geometry on natural TallyQA};
\begin{axis}[traxis,at={(1.0cm,-.70cm)},ybar,bar width=22pt,xmin=-.6,xmax=2.6,ymin=63,ymax=68,
ylabel={Accuracy (\%)},xtick={0,1,2},xticklabels={Baseline,Whole,MN-CSA},ytick={63,64,65,66,67,68},
nodes near coords,every node near coord/.append style={font=\rmfamily\fontsize{7.5}{9}\selectfont\bfseries,anchor=south,yshift=1pt},legend image code/.code={\path[#1] (0pt,-2pt) rectangle (8pt,3pt);}]
\addplot[draw=none,fill=aromaGray,bar shift=0pt] coordinates {(0,64.975)};
\addplot[draw=none,fill=aromaBlue,bar shift=0pt] coordinates {(1,67.075)};
\addplot[draw=none,fill=aromaGreen,bar shift=0pt] coordinates {(2,66.375)};
\end{axis}

\node[trtitle,align=left,text width=8.0cm] at (\trHalf,0) {(b) Zero-shot policy failure,\\actuator headroom remains};
\begin{axis}[traxis,at={(\trRight,-.70cm)},ybar,bar width=22pt,xmin=-.6,xmax=2.6,ymin=58,ymax=74,
ylabel={Accuracy (\%)},xtick={0,1,2},xticklabels={Baseline,Synthetic router,Action oracle},xticklabel style={font=\rmfamily\fontsize{7.5}{9}\selectfont,align=center},
ytick={58,62,66,70,74},nodes near coords,every node near coord/.append style={font=\rmfamily\fontsize{7.5}{9}\selectfont\bfseries,anchor=south,yshift=1pt},legend image code/.code={\path[#1] (0pt,-2pt) rectangle (8pt,3pt);}]
\addplot[draw=none,fill=aromaGray,bar shift=0pt] coordinates {(0,64.85)};
\addplot[draw=none,fill=aromaOrange,bar shift=0pt] coordinates {(1,61.875)};
\addplot[draw=none,fill=aromaGreen,bar shift=0pt] coordinates {(2,71.575)};
\end{axis}

\node[trtitle] at (0,-3.75cm) {(c) Sample-efficient natural recalibration};
\begin{axis}[traxis,at={(1.0cm,-4.27cm)},xmode=log,log basis x=10,xmin=40,xmax=4000,ymin=0,ymax=2.8,
xlabel={Calibration size $K$},ylabel={Mean gain (pp)},xtick={50,100,250,500,1000,2000,3200},
xticklabels={50,100,250,500,1k,2k,3.2k},xticklabel style={font=\rmfamily\fontsize{7.5}{9}\selectfont,rotate=25,anchor=north east},ytick={0,1,2}]
\addplot[color=aromaBlue,line width=.9pt,mark=*,mark size=2pt] coordinates {(50,.330)(100,1.180)(250,1.870)(500,2.120)(1000,2.240)(2000,2.325)(3200,2.300)};
\draw[densely dashed,aromaGreen,line width=.55pt] (axis cs:500,0)--(axis cs:500,2.8);
\node[font=\rmfamily\fontsize{7.5}{9}\selectfont\bfseries,text=aromaGreen!70!black,anchor=south west] at (axis cs:500,.25) {$K=500$};
\end{axis}

\node[trtitle] at (\trHalf,-3.75cm) {(d) HoloCount external boundary};
\begin{axis}[traxis,at={(\trRight,-4.27cm)},ybar,bar width=22pt,xmin=-.6,xmax=2.6,ymin=42,ymax=62,
ylabel={Accuracy (\%)},xtick={0,1,2},xticklabels={Baseline,Transferred,Oracle},ytick={42,46,50,54,58,62},
nodes near coords,every node near coord/.append style={font=\rmfamily\fontsize{7.5}{9}\selectfont\bfseries,anchor=south,yshift=1pt},legend image code/.code={\path[#1] (0pt,-2pt) rectangle (8pt,3pt);}]
\addplot[draw=none,fill=aromaGray,bar shift=0pt] coordinates {(0,46.694)};
\addplot[draw=none,fill=aromaPurple,bar shift=0pt] coordinates {(1,46.653)};
\addplot[draw=none,fill=aromaGreen,bar shift=0pt] coordinates {(2,58.952)};
\end{axis}
\end{tikzpicture}%
\ifaclstyle}\fi
\caption{\textbf{Transfer separates geometry from deployment policy.}
(a) Llama rank-4 geometry retains positive natural-domain utility without refitting.
(b) The synthetic router fails zero-shot on a separate TallyQA set even while the same actuator/action family retains large oracle headroom.
(c) Relearning only the lightweight utility map is sample-efficient; $K=500$ is fixed as an efficiency elbow before a new confirmation set.
(d) HoloCount preserves large oracle headroom while the transferred policy is neutral, marking the frontier for richer routing and readout.}
\label{fig:transfer_layers}
\ifaclstyle\end{figure*}
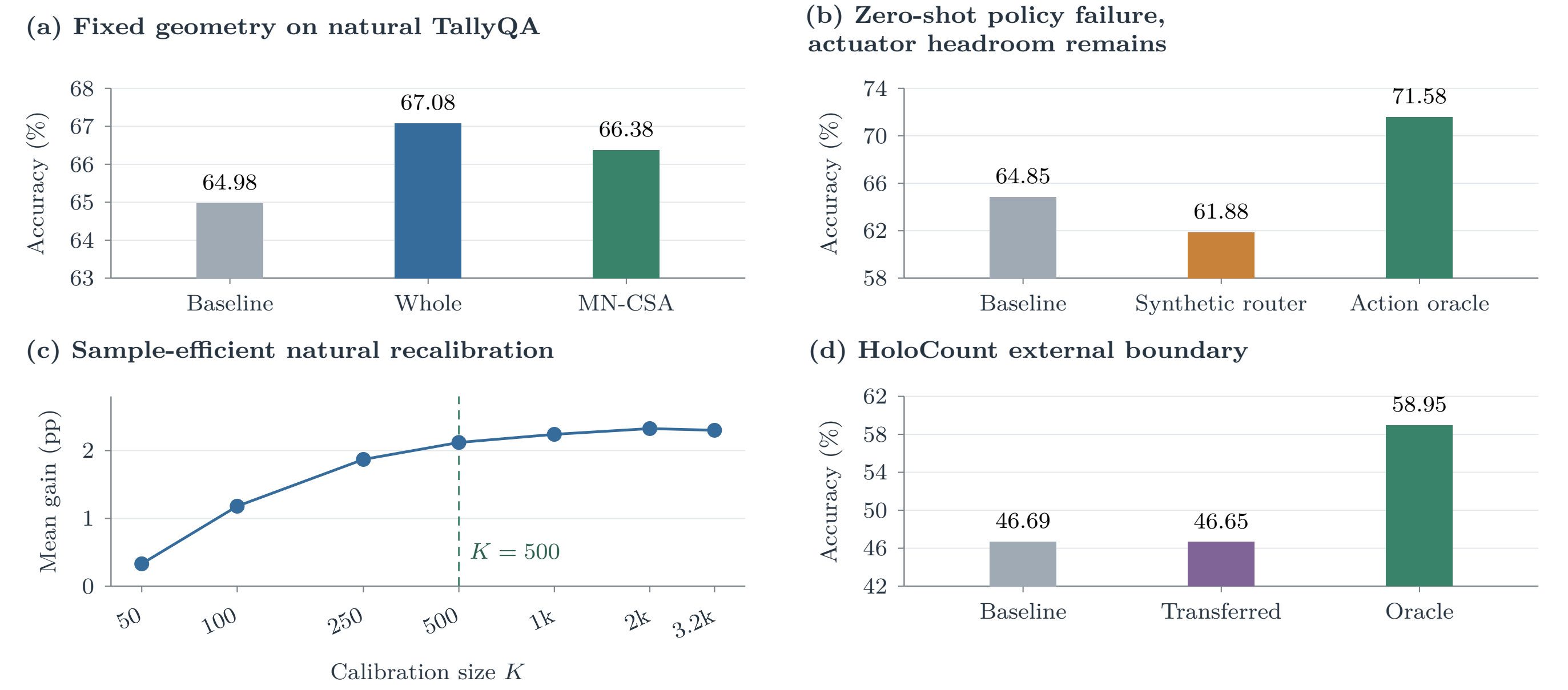\else\end{figure}\fi

\ifaclstyle\begin{figure*}[!tp]\else\begin{figure}[!htbp]\fi
\centering

\pgfmathsetlengthmacro{\fHalf}{0.5*\textwidth}
\pgfmathsetlengthmacro{\fWidth}{0.5*\textwidth-1.55cm}
\pgfmathsetlengthmacro{\fRightAxis}{\fHalf+1.15cm}
\begin{tikzpicture}
\pgfplotsset{
  forensic/.style={
    scale only axis,width=\fWidth,height=2.65cm,
    anchor=north west,
    axis background/.style={fill=white},
    axis x line*=bottom,axis y line*=left,
    axis line style={draw=aromaInk!60,line width=.45pt},
    tick style={draw=aromaInk!60,line width=.4pt},
    tick align=outside,major tick length=2pt,
    tick label style={font=\rmfamily\fontsize{7.5}{9}\selectfont,text=aromaInk},
    label style={font=\rmfamily\fontsize{8}{9.5}\selectfont,text=aromaInk},
    ymajorgrids=true,xmajorgrids=false,
    grid style={draw=aromaGrid,line width=.35pt},
    scaled ticks=false,clip=true,
    legend style={
      font=\rmfamily\fontsize{7.5}{9}\selectfont,
      draw=none,fill=none,inner sep=0pt,
      at={(.5,1.035)},anchor=south,legend columns=2,
      /tikz/every even column/.append style={column sep=9pt}
    }
  }
}
\tikzset{
  ftitle/.style={anchor=west,inner sep=0pt,text=aromaInk,
    font=\rmfamily\fontsize{9}{11}\selectfont\bfseries},
  fvalue/.style={inner sep=1pt,text=aromaInk,
    font=\rmfamily\fontsize{7.5}{9}\selectfont\bfseries}
}

\node[ftitle] at (0,0) {(a) Performance reversal};
\begin{axis}[
  forensic,at={(1.10cm,-.80cm)},
  ybar,bar width=23pt,bar shift=0pt,
  xmin=.4,xmax=2.6,ymin=-5,ymax=11,
  ylabel={Accuracy change (pp)},
  xtick={1,2},xticklabels={Procedural v3,Natural zero-shot},
  ytick={-5,0,5,10}
,legend image code/.code={\path[#1] (0pt,-2pt) rectangle (8pt,3pt);}]
\addplot[draw=none,fill=aromaBlue] coordinates {(1,8.10)};
\addplot[draw=none,fill=aromaOrange] coordinates {(2,-2.975)};
\draw[densely dashed,aromaInk!55,line width=.5pt]
  (axis cs:.4,0)--(axis cs:2.6,0);
\node[fvalue,anchor=south,yshift=3pt]
  at (axis cs:1,8.10) {$+8.10$};
\node[fvalue,anchor=north,yshift=-2pt]
  at (axis cs:2,-2.975) {$-2.975$};
\end{axis}

\node[ftitle] at (\fHalf,0) {(b) Error-direction shift};
\begin{axis}[
  forensic,name=forensicErrors,at={(\fRightAxis,-.80cm)},
  ybar stacked,bar width=34pt,clip mode=individual,
  xmin=.4,xmax=2.6,ymin=0,ymax=100,
  ylabel={Baseline errors (\%)},
  xtick={1,2},xticklabels={Synthetic v2,Natural TallyQA},
  ytick={0,25,50,75,100}
,legend image code/.code={\path[#1] (0pt,-2pt) rectangle (8pt,3pt);}]
\addplot[draw=none,fill=aromaBlue]
  coordinates {(1,93.76) (2,21.12)};
\addplot[draw=none,fill=aromaOrange]
  coordinates {(1,6.24) (2,78.88)};
\legend{Undercount,Overcount}
\coordinate (forensicLabel0) at (axis cs:1,46.88);
\coordinate (forensicLabel1) at (axis cs:2,10.56);
\coordinate (forensicLabel2) at (axis cs:2,60.56);
\node[fvalue,anchor=west,xshift=19pt]
  at (axis cs:1,96.88) {6.24};
\end{axis}
\node[fvalue,text=white] at (forensicLabel0) {93.76};
\node[fvalue,text=white] at (forensicLabel1) {21.12};
\node[fvalue,text=white] at (forensicLabel2) {78.88};

\node[ftitle] at (0,-4.25cm) {(c) Score--utility alignment};
\begin{axis}[
  forensic,at={(1.10cm,-5.10cm)},
  ybar,bar width=20pt,bar shift=0pt,
  xmin=.45,xmax=3.55,ymin=-.19,ymax=.52,
  ylabel={Spearman $\rho$},
  xtick={1,2,3},xticklabels={v2 OOF,v3 fixed,TallyQA},
  ytick={-.1,0,.2,.4},
  yticklabels={$-0.1$,$0$,$0.2$,$0.4$}
,legend image code/.code={\path[#1] (0pt,-2pt) rectangle (8pt,3pt);}]
\addplot[draw=none,fill=aromaGreen]
  coordinates {(1,.409) (2,.411)};
\addplot[draw=none,fill=aromaOrange]
  coordinates {(3,-.0667)};
\draw[densely dashed,aromaInk!55,line width=.5pt]
  (axis cs:.45,0)--(axis cs:3.55,0);
\node[fvalue,anchor=south,yshift=3pt]
  at (axis cs:1,.409) {0.409};
\node[fvalue,anchor=south,yshift=3pt]
  at (axis cs:2,.411) {0.411};
\node[fvalue,anchor=north,yshift=-2pt]
  at (axis cs:3,-.0667) {$-0.067$};
\end{axis}

\node[ftitle] at (\fHalf,-4.25cm)
  {(d) Support clipping};
\begin{axis}[
  forensic,at={(\fRightAxis,-5.10cm)},
  xmin=.45,xmax=2.55,ymin=60,ymax=68,
  ylabel={Percent (\%)},
  xtick={1,2},xticklabels={Post accuracy,Intervention rate},
  ytick={60,62,64,66,68}
]
\addplot[only marks,color=aromaGrayDark,mark=square*,mark size=2.7pt]
  coordinates {(.80,61.875) (1.80,65.275)};
\addplot[only marks,color=aromaPurple,mark=*,mark size=2.7pt]
  coordinates {(1.20,61.875) (2.20,63.275)};
\legend{Unclipped,Clipped}
\node[fvalue,anchor=south,yshift=4pt]
  at (axis cs:.80,61.875) {61.875};
\node[fvalue,anchor=north,yshift=-5pt]
  at (axis cs:1.20,61.875) {61.875};
\node[fvalue,anchor=south,yshift=4pt]
  at (axis cs:1.80,65.275) {65.275};
\node[fvalue,anchor=south,yshift=4pt]
  at (axis cs:2.20,63.275) {63.275};
\end{axis}
\end{tikzpicture}

\par\vspace{2pt}
{\rmfamily\fontsize{8}{9.5}\selectfont
Support clipping: 39.825\% of samples clipped;
82.95\% policy agreement with the router.\par}

\caption{\textbf{Diagnosing zero-shot policy transfer.}
\textbf{(a)} The synthetic router improves procedural v3 accuracy by $+8.10\pp$ but changes zero-shot \tally{} accuracy by $-2.975\pp$. The post hoc natural action oracle reaches 71.575\% ($+6.725\pp$ over the 64.85\% baseline), indicating retained actuator headroom.
\textbf{(b)} Baseline errors shift from 93.76\% undercounts and 6.24\% overcounts on synthetic v2 to 21.12\% undercounts and 78.88\% overcounts on natural \tally{}.
\textbf{(c)} Selected-score/realized-utility Spearman correlation is positive on v2 and v3 ($\rho\approx0.409$ and $0.411$) but near zero and slightly negative on natural data ($\rho\approx-0.0667$).
\textbf{(d)} Without fitting or tuning parameters, support clipping maps standardized features to empirical v2 training support. Clipping affects 39.825\% of samples; accuracy remains 61.875\%, while intervention rate changes from 65.275\% to 63.275\%. Support violation contributes to extreme extrapolation but does not explain the accuracy decrease, indicating a conditional-utility shift.}
\label{fig:supp_forensics}
\ifaclstyle\end{figure*}\else\end{figure}\fi

\ifaclstyle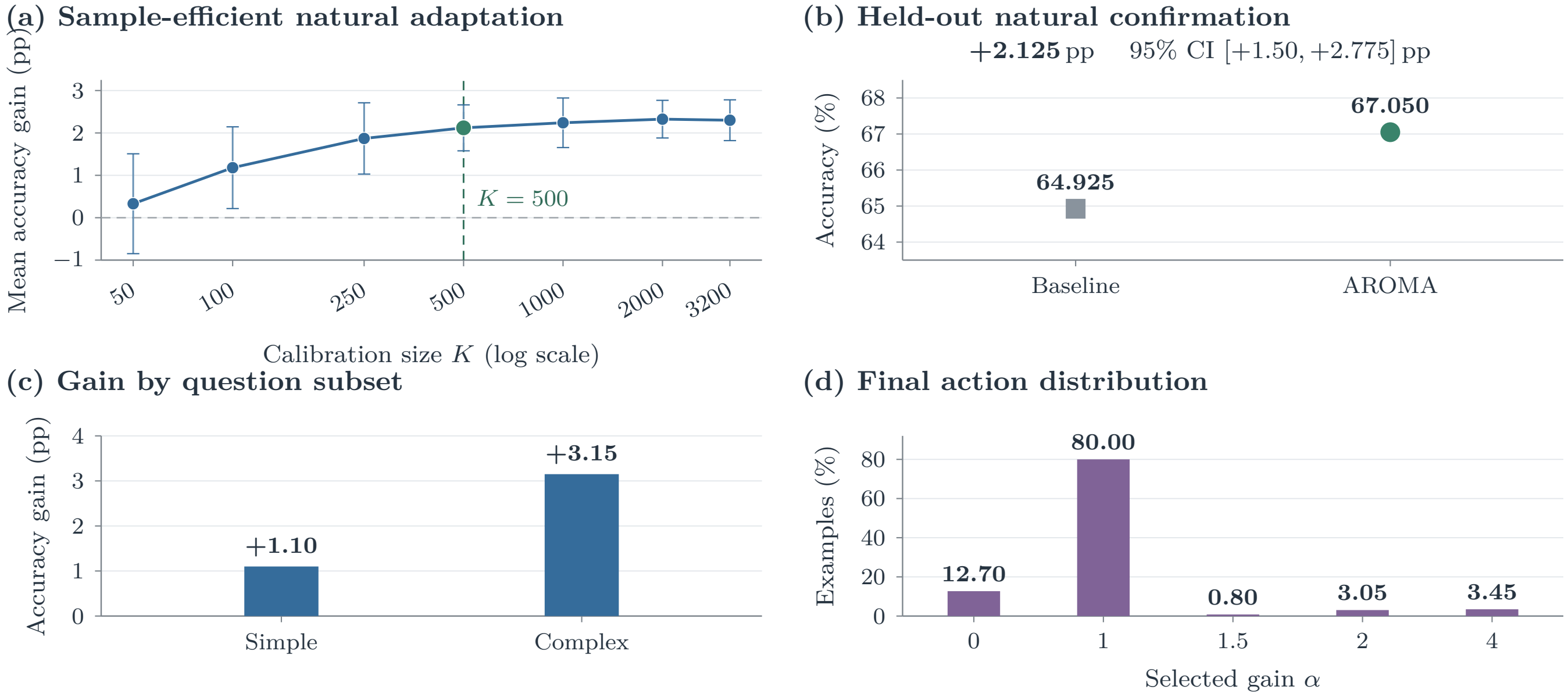
\begin{figure*}[!tp]\else\begin{figure}[!htbp]\fi
\centering

\pgfmathsetlengthmacro{\natHalf}{0.5*\textwidth}
\pgfmathsetlengthmacro{\natWidth}{0.5*\textwidth-1.55cm}
\pgfmathsetlengthmacro{\natRight}{\natHalf+1.15cm}

\begin{tikzpicture}

\pgfplotsset{
    natural axis/.style={
        scale only axis,
        width=\natWidth,
        height=2.05cm,
        anchor=north west,
        axis background/.style={fill=white},
        axis x line*=bottom,
        axis y line*=left,
        axis line style={
            draw=aromaInk!60,
            line width=0.45pt
        },
        tick style={
            draw=aromaInk!60,
            line width=0.4pt
        },
        tick align=outside,
        major tick length=2pt,
        tick label style={
            font=\rmfamily\fontsize{7.5}{9}\selectfont,
            text=aromaInk
        },
        label style={
            font=\rmfamily\fontsize{8}{9.5}\selectfont,
            text=aromaInk
        },
        ymajorgrids=true,
        xmajorgrids=false,
        grid style={
            draw=aromaGrid,
            line width=0.35pt
        },
        scaled ticks=false,
        clip=true
    }
}

\tikzset{
    nat title/.style={
        anchor=west,
        inner sep=0pt,
        text=aromaInk,
        font=\rmfamily\fontsize{9}{11}\selectfont\bfseries
    },
    nat value/.style={
        inner sep=1pt,
        text=aromaInk,
        font=\rmfamily\fontsize{7.5}{9}\selectfont\bfseries
    }
}

\node[nat title] at (0,0)
    {(a) Sample-efficient natural adaptation};

\begin{axis}[
    natural axis,
    at={(1.10cm,-0.70cm)},
    xmode=log,
    log basis x=10,
    xmin=40,
    xmax=4000,
    ymin=-1.0,
    ymax=3.25,
    xlabel={Calibration size $K$ (log scale)},
    ylabel={Mean accuracy gain (pp)},
    xtick={50,100,250,500,1000,2000,3200},
    xticklabels={50,100,250,500,1000,2000,3200},
    xticklabel style={
        font=\rmfamily\fontsize{7.5}{9}\selectfont,
        text=aromaInk,
        rotate=30,
        anchor=north east
    },
    minor x tick num=0,
    ytick={-1,0,1,2,3}
]

\draw[
    densely dashed,
    aromaInk!45,
    line width=0.45pt
]
(axis cs:40,0) -- (axis cs:4000,0);

\draw[
    densely dashed,
    aromaGreen!85!black,
    line width=0.55pt
]
(axis cs:500,-1) -- (axis cs:500,3.25);

\addplot[
    color=aromaBlue,
    line width=0.9pt,
    mark=*,
    mark size=2pt,
    mark options={
        fill=aromaBlue,
        draw=white,
        line width=0.3pt
    },
    error bars/.cd,
    y dir=both,
    y explicit,
    error bar style={
        draw=aromaBlue!75,
        line width=0.55pt
    }
]
coordinates {
    (50,0.330)   +- (0,1.178)
    (100,1.180)  +- (0,0.964)
    (250,1.870)  +- (0,0.841)
    (500,2.120)  +- (0,0.542)
    (1000,2.240) +- (0,0.585)
    (2000,2.325) +- (0,0.444)
    (3200,2.300) +- (0,0.481)
};

\addplot[
    only marks,
    mark=*,
    mark size=2.6pt,
    color=aromaGreen,
    mark options={
        fill=aromaGreen,
        draw=white,
        line width=0.4pt
    }
]
coordinates {(500,2.120)};

\node[
    nat value,
    text=aromaGreen!80!black,
    anchor=south west,
    xshift=3pt
] at (axis cs:500,0.20) {$K=500$};

\end{axis}

\node[nat title] at (\natHalf,0)
    {(b) Held-out natural confirmation};

\node[
    anchor=center,
    inner sep=0pt,
    text=aromaInk,
    font=\rmfamily\fontsize{8}{9.5}\selectfont
] at ({\natHalf+0.25*\textwidth},-0.38cm) {
    $\boldsymbol{+2.125}\,\mathrm{pp}$
    \quad 95\% CI $[+1.50,+2.775]\,\mathrm{pp}$
};

\begin{axis}[
    natural axis,
    at={(\natRight,-0.70cm)},
    xmin=0.45,
    xmax=2.55,
    ymin=63.5,
    ymax=68.5,
    ylabel={Accuracy (\%)},
    xtick={1,2},
    xticklabels={Baseline,AROMA},
    ytick={64,65,66,67,68}
]

\addplot[
    only marks,
    mark=square*,
    mark size=3pt,
    color=aromaGrayMid
]
coordinates {(1,64.925)};

\addplot[
    only marks,
    mark=*,
    mark size=3pt,
    color=aromaGreen
]
coordinates {(2,67.050)};

\node[
    nat value,
    anchor=south,
    yshift=5pt
] at (axis cs:1,64.925) {64.925};

\node[
    nat value,
    anchor=south,
    yshift=5pt
] at (axis cs:2,67.050) {67.050};

\end{axis}

\node[nat title] at (0,-4.15cm)
    {(c) Gain by question subset};

\begin{axis}[
    natural axis,
    at={(1.10cm,-4.75cm)},
    ybar,
    bar width=24pt,
    bar shift=0pt,
    xmin=0.4,
    xmax=2.6,
    ymin=0,
    ymax=4,
    ylabel={Accuracy gain (pp)},
    xtick={1,2},
    xticklabels={Simple,Complex},
    ytick={0,1,2,3,4},
    nodes near coords,
    point meta=explicit symbolic,
    every node near coord/.append style={
        font=\rmfamily\fontsize{7.5}{9}\selectfont\bfseries,
        text=aromaInk,
        anchor=south,
        yshift=3pt,
        inner sep=0pt
    }
,legend image code/.code={\path[#1] (0pt,-2pt) rectangle (8pt,3pt);}]

\addplot[
    draw=none,
    fill=aromaBlue
]
coordinates {
    (1,1.10) [+1.10]
    (2,3.15) [+3.15]
};

\end{axis}

\node[nat title] at (\natHalf,-4.15cm)
    {(d) Final action distribution};

\begin{axis}[
    natural axis,
    at={(\natRight,-4.75cm)},
    ybar,
    bar width=17pt,
    bar shift=0pt,
    xmin=-0.55,
    xmax=4.55,
    ymin=0,
    ymax=92,
    ylabel={Examples (\%)},
    xlabel={Selected gain $\alpha$},
    xtick={0,1,2,3,4},
    xticklabels={$0$,$1$,$1.5$,$2$,$4$},
    ytick={0,20,40,60,80},
    nodes near coords,
    point meta=explicit symbolic,
    every node near coord/.append style={
        font=\rmfamily\fontsize{7.5}{9}\selectfont\bfseries,
        text=aromaInk,
        anchor=south,
        yshift=3pt,
        inner sep=0pt
    }
,legend image code/.code={\path[#1] (0pt,-2pt) rectangle (8pt,3pt);}]

\addplot[
    draw=none,
    fill=aromaPurple
]
coordinates {
    (0,12.70) [12.70]
    (1,80.00) [80.00]
    (2, 0.80) [0.80]
    (3, 3.05) [3.05]
    (4, 3.45) [3.45]
};

\end{axis}
\end{tikzpicture}

\caption{\textbf{Lightweight natural-domain calibration restores routing utility.}
\textbf{(a)} With the actuator and routing model family fixed, mean development gain rises from $+0.33\pp$ at $K=50$ to $+2.12\pp$ at $K=500$ and then largely saturates. The selected $K=500$ is a sample-efficiency elbow, not the maximum-gain point; error bars denote across-run standard deviations.
\textbf{(b)} Evaluation on a new 4,000-example image-disjoint \tally{} set improves accuracy from 64.925\% to 67.050\% ($+2.125\pp$; paired 95\% CI $[+1.50,+2.775]\pp$), with 130 repairs, 45 breaks, and $+85$ net repairs (exact McNemar $p=9.18\times10^{-11}$).
\textbf{(c)} Both prespecified subsets improve: Simple by $+1.10\pp$ and Complex by $+3.15\pp$.
\textbf{(d)} The policy retains identity ($\alpha=1$) for 80.0\% of examples and applies non-identity actions to 20.0\%.}
\label{fig:supp_naturaladapt}

\ifaclstyle\end{figure*}\else\end{figure}\fi

\subsection{Hierarchical Natural UGR}
\label{app:natural_ugr}
Natural Stage-2 development begins from the $N=292$ upward proposals selected by the natural-domain Proposal Controller. The candidate geometry/gain space is identical in concept to procedural UGR, but the policy architecture is treated as a domain-specific development choice. Table~\ref{tab:natural_ugr_dev} records the sequence that led to the final hierarchical factorization.

For the selected hierarchical design, Whole produces 12 repairs and 0 breaks and Unified 16 repairs and 0 breaks on development. The gate threshold is 0.5; logistic models use $C=1$ and the 47-D GT-free sample representation. These choices, the Whole/Unified candidate families, and the complete 14-gain grid are fixed before final-cohort evaluation.

The held-out final cohort contains 8,000 TallyQA examples, balanced 4,000 simple/4,000 complex, with zero question, image-ID, and pixel overlap against the earlier natural data. The Proposal Controller emits 990 proposals at $\alpha=0$, 6,379 at $\alpha=1$, 64 at $\alpha=1.5$, 271 at $\alpha=2$, and 296 at $\alpha=4$, yielding the prespecified upward Stage-2 population $N=631$. For every Stage-2 example, all 43 candidates (IDENTITY plus 14 each for MN, SA, and WHOLE) are executed, for 27,133 candidate outcomes.

\ifaclstyle\begin{table*}[!tp]\else\begin{table}[H]\fi
\centering
\begin{minipage}[t]{0.487\linewidth}\vspace{0pt}
\centering\fontsize{8.5}{10.5}\selectfont
\setlength{\tabcolsep}{2.6pt}
\begin{minipage}[t][4\baselineskip][t]{\linewidth}
\caption{Natural UGR Stage-2 development. Flat and pairwise variants are retained as negative/near-neutral development results; only the hierarchical gate-plus-ranker is carried into the final evaluation.}
\label{tab:natural_ugr_dev}
\end{minipage}\par
\begin{tabularx}{\linewidth}{@{}Yrrrr@{}}
\toprule
\bfseries Variant & \bfseries $N$ & \bfseries Whole & \bfseries Unified & \bfseries U--W \\
\midrule
Flat candidate classifier & 292 & 38.014\% & 35.959\% & $-2.055\pp$ \\
Pairwise ranker & 292 & 36.986\% & 36.301\% & $-0.685\pp$ \\
Hierarchical gate + ranker & 292 & 34.247\% & 35.616\% & $+1.370\pp$ \\
\bottomrule
\end{tabularx}
\end{minipage}\hfill
\begin{minipage}[t]{0.487\linewidth}\vspace{0pt}
\centering\fontsize{8.5}{10.5}\selectfont
\setlength{\tabcolsep}{2.6pt}
\begin{minipage}[t][4\baselineskip][t]{\linewidth}
\caption{Natural hierarchical UGR held-out final result on the prespecified $N=631$ Stage-2 population. Oracles are computed from the exhaustive candidate evaluation.}
\label{tab:natural_ugr_final}
\end{minipage}\par
\begin{tabularx}{\linewidth}{@{}Yrrrrr@{}}
\toprule
\bfseries Policy & \bfseries Correct & \bfseries Accuracy & \bfseries Repairs & \bfseries Breaks & \bfseries Oracle \\
\midrule
Identity & 201 & 31.854\% & -- & -- & -- \\
Hierarchical Whole & 211 & 33.439\% & 16 & 6 & 43.106\% \\
Hierarchical Unified & 223 & 35.341\% & 26 & 4 & 49.287\% \\
\bottomrule
\end{tabularx}
\end{minipage}
\ifaclstyle\end{table*}\else\end{table}\fi

Unified minus Whole is $+1.902\pp$ with paired bootstrap 95\% CI $[+0.792,+3.170]\pp$, exact two-sided McNemar $p=0.00418$, and paired discordance 14 Unified-only versus 2 Whole-only. Whole intervenes on 48/631 (7.61\%) and Unified on 52/631 (8.24\%). Unified selections are IDENTITY 579, WHOLE 27, MN 20, and SA 5; its selected gains are 0.5 on one intervention, 3.25 on one, and 4.0 on 50.

\ifaclstyle\begin{table*}[!tp]\else\begin{table}[H]\fi
\centering\fontsize{8.5}{10.5}\selectfont
\caption{Natural final candidate-oracle capacity and subgroup routing results. Subgroup differences are descriptive.}
\label{tab:natural_ugr_oracles_subgroups}
\begin{tabular}{@{}lrr@{}}
\toprule
\bfseries Oracle / subgroup & \bfseries Value & \bfseries Difference \\
\midrule
MN oracle & 44.216\% & $+12.361\pp$ vs identity \\
SA oracle & 40.095\% & $+8.241\pp$ vs identity \\
Low-rank oracle (MN/SA) & 46.751\% & $+14.897\pp$ vs identity \\
Whole oracle & 43.106\% & $+11.252\pp$ vs identity \\
Unified oracle & 49.287\% & $+17.433\pp$ vs identity \\
\midrule
Complex ($N=450$) & 32.444\% $\rightarrow$ 34.889\% & $+2.444\pp$ \\
Simple ($N=181$) & 35.912\% $\rightarrow$ 36.464\% & $+0.552\pp$ \\
\bottomrule
\end{tabular}
\ifaclstyle\end{table*}\else\end{table}\fi

Because the Whole and Unified pipelines are identical on the remaining 7,369 examples, the final Stage-2 advantage corresponds mechanically to a $12/8000=+0.150\pp$ pairwise difference over the full cohort. We retain the conditional $N=631$ comparison as the primary endpoint because the scientific question is which geometry/gain should refine an already fixed upward proposal.

\paragraph{Action-space headroom.}
The exhaustive candidate evaluation separates available capacity from learned realization. The Whole oracle reaches 272/631=43.106\%, whereas the Unified oracle reaches 311/631=49.287\%, a $+6.181\pp$ capacity advantage. Individual-family oracles are 44.216\% for MN, 40.095\% for SA, and 46.751\% for the combined low-rank family. No single family is universal on natural images; the evidence supports conditional access to complementary actions. The learned router realizes $+1.902\pp$ of this incremental Unified-vs-Whole oracle gap without changing the candidate set.

\paragraph{Hardware numerical check.}
Re-executing the 631-example population on a different GPU type reproduced 630/631 baseline predictions (one change, 0.158\%) and 629/631 Stage-1 actions, with both action changes at decision boundaries. The Stage-1 population, proposal actions, baseline reference, and both routers were kept unchanged; this is a hardware precision effect, not a result-dependent policy change.

\ifaclstyle\begin{figure*}[!tp]\else\begin{figure}[!htbp]\fi
\centering
\begin{minipage}[t]{0.48\textwidth}
\centering
\begin{tikzpicture}
\begin{axis}[
scale only axis,width=\linewidth-1.24cm,height=3.10cm,
axis x line*=bottom,axis y line*=left,enlarge x limits=0.30,
clip=false,tick align=outside,axis line style={draw=black!55,line width=.45pt},
ybar,bar shift=0pt,bar width=14pt,ymin=28,ymax=38,
ylabel={Accuracy (\%)},symbolic x coords={ID,Whole,Unified},xtick={ID,Whole,Unified},
xticklabel style={font=\rmfamily\fontsize{7.5}{9}\selectfont},tick label style={font=\rmfamily\fontsize{7.5}{9}\selectfont},label style={font=\rmfamily\fontsize{7.5}{9}\selectfont},
ymajorgrids=true,grid style={draw=aromaRule!30},nodes near coords,
every node near coord/.append style={font=\rmfamily\fontsize{7.5}{9}\selectfont\bfseries,anchor=south},legend image code/.code={\path[#1] (0pt,-2pt) rectangle (8pt,3pt);}]
\addplot[draw=none,fill=aromaGray] coordinates {(ID,31.854)};
\addplot[draw=none,fill=aromaBlue] coordinates {(Whole,33.439)};
\addplot[draw=none,fill=aromaGreen] coordinates {(Unified,35.341)};
\end{axis}
\end{tikzpicture}
\par\vspace{4pt}{\rmfamily\fontsize{7.5}{9}\selectfont\textbf{(a) Final accuracy}\\$\Delta_{U-W}=+1.902\pp$; $p=0.00418$}
\end{minipage}\hfill
\begin{minipage}[t]{0.48\textwidth}
\centering
\begin{tikzpicture}
\begin{axis}[
scale only axis,width=\linewidth-1.24cm,height=3.10cm,
axis x line*=bottom,axis y line*=left,enlarge x limits=0.30,
clip=false,tick align=outside,axis line style={draw=black!55,line width=.45pt},
ybar,bar shift=0pt,bar width=14pt,ymin=38,ymax=52,
ylabel={Oracle (\%)},symbolic x coords={Whole,Unified},xtick={Whole,Unified},
xticklabel style={font=\rmfamily\fontsize{7.5}{9}\selectfont},tick label style={font=\rmfamily\fontsize{7.5}{9}\selectfont},label style={font=\rmfamily\fontsize{7.5}{9}\selectfont},
ymajorgrids=true,grid style={draw=aromaRule!30},nodes near coords,
every node near coord/.append style={font=\rmfamily\fontsize{7.5}{9}\selectfont\bfseries,anchor=south},legend image code/.code={\path[#1] (0pt,-2pt) rectangle (8pt,3pt);}]
\addplot[draw=none,fill=aromaBlue] coordinates {(Whole,43.106)};
\addplot[draw=none,fill=aromaGreen] coordinates {(Unified,49.287)};
\end{axis}
\end{tikzpicture}
\par\vspace{4pt}{\rmfamily\fontsize{7.5}{9}\selectfont\textbf{(b) Candidate capacity}\\Unified oracle $+6.181\pp$}
\end{minipage}\par\vspace{14pt}
\begin{minipage}[t]{0.48\textwidth}
\centering
\begin{tikzpicture}
\begin{axis}[
scale only axis,width=\linewidth-1.24cm,height=3.10cm,
axis x line*=bottom,axis y line*=left,enlarge x limits=0.30,
clip=false,tick align=outside,axis line style={draw=black!55,line width=.45pt},
ybar,bar width=14pt,ymin=0,ymax=30,
ylabel={Examples},symbolic x coords={Whole,Unified},xtick={Whole,Unified},
xticklabel style={font=\rmfamily\fontsize{7.5}{9}\selectfont},tick label style={font=\rmfamily\fontsize{7.5}{9}\selectfont},label style={font=\rmfamily\fontsize{7.5}{9}\selectfont},
ymajorgrids=true,grid style={draw=aromaRule!30},legend style={font=\rmfamily\fontsize{7.5}{9}\selectfont,draw=none,at={(0.5,1.04)},anchor=south,legend columns=2},nodes near coords,
every node near coord/.append style={font=\rmfamily\fontsize{7.5}{9}\selectfont\bfseries,anchor=south},legend image code/.code={\path[#1] (0pt,-2pt) rectangle (8pt,3pt);}]
\addplot[draw=none,fill=aromaGreen] coordinates {(Whole,16) (Unified,26)};
\addplot[draw=none,fill=aromaOrange] coordinates {(Whole,6) (Unified,4)};
\legend{Repairs,Breaks}
\end{axis}
\end{tikzpicture}
\par\vspace{4pt}{\rmfamily\fontsize{7.5}{9}\selectfont\textbf{(c) Repair--break profile}\\more repairs, fewer breaks}
\end{minipage}\hfill
\begin{minipage}[t]{0.48\textwidth}
\centering
\begin{tikzpicture}
\begin{axis}[
scale only axis,width=\linewidth-1.24cm,height=3.10cm,
axis x line*=bottom,axis y line*=left,enlarge x limits=0.30,
clip=false,tick align=outside,axis line style={draw=black!55,line width=.45pt},
ybar,bar shift=0pt,bar width=14pt,ymin=0,ymax=30,
ylabel={Selections},symbolic x coords={WHOLE,MN,SA},xtick={WHOLE,MN,SA},
xticklabel style={font=\rmfamily\fontsize{7.5}{9}\selectfont},tick label style={font=\rmfamily\fontsize{7.5}{9}\selectfont},label style={font=\rmfamily\fontsize{7.5}{9}\selectfont},
ymajorgrids=true,grid style={draw=aromaRule!30},nodes near coords,
every node near coord/.append style={font=\rmfamily\fontsize{7.5}{9}\selectfont\bfseries,anchor=south},legend image code/.code={\path[#1] (0pt,-2pt) rectangle (8pt,3pt);}]
\addplot[draw=none,fill=aromaPurple] coordinates {(WHOLE,27)};
\addplot[draw=none,fill=aromaBlue] coordinates {(MN,20)};
\addplot[draw=none,fill=aromaOrange] coordinates {(SA,5)};
\end{axis}
\end{tikzpicture}
\par\vspace{4pt}{\rmfamily\fontsize{7.5}{9}\selectfont\textbf{(d) Unified interventions}\\25/52 use MN or SA}
\end{minipage}
\caption{\textbf{Natural hierarchical UGR converts complementary action-space capacity into a held-out gain.} On the prespecified $N=631$ Stage-2 population from a new 8,000-example TallyQA cohort, fixed Unified routing improves over fixed Whole-only routing by $+1.902\pp$ (95\% CI $[+0.792,+3.170]\pp$; exact McNemar $p=0.00418$; 14:2 paired discordance). The unified candidate oracle adds $+6.181\pp$ over the Whole oracle. The realized policy is sparse (52/631 interventions), increases repairs while reducing breaks, and does not collapse to WHOLE: 48.1\% of its non-identity selections use MN or SA.}
\label{fig:natural_ugr_v2}
\ifaclstyle\end{figure*}
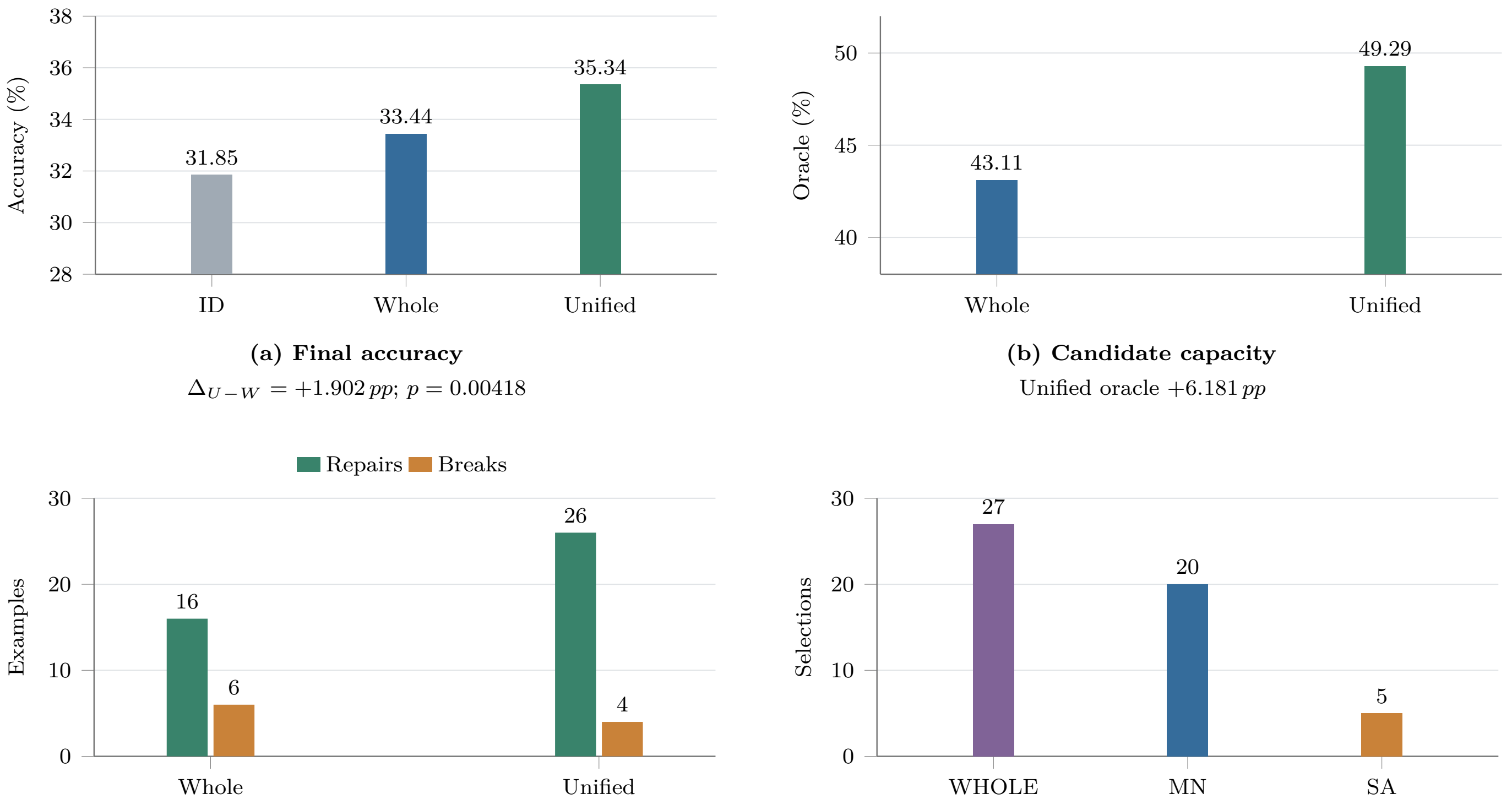\else\end{figure}\fi

\subsection{HoloCount External Boundary}
The HoloCount external run contains 2,480 questions and 2,056 unique image hashes over 20 subsets.
The TallyQA-adapted controller changes 46.69\% baseline accuracy to 46.65\% ($-0.04\pp$), with 72 repairs and 73 breaks.
The action oracle reaches 58.95\% ($+12.26\pp$) and identifies 304 repairable baseline errors.
Fixed-action changes are $-0.85\pp$ for $\alpha=0$, $+0.04\pp$ for $\alpha=1.5$, $-0.65\pp$ for $\alpha=2$, and $-3.59\pp$ for $\alpha=4$.
Grouped OOF training with the same 39-feature StandardScaler$\rightarrow$Ridge family also yields approximately null gain and near-zero score--utility alignment.
HoloCount therefore marks the frontier for routing and readout: the actuator/action family retains $+12.26\pp$ of conditional repair capacity, a concrete target for representations beyond the tested linear family.
\Needspace{6\baselineskip}
\section{Target Selection and Deployment}
\label{app:target}

This section documents the complete targeted-deployment chain (Gap~3) on the same fixed actuator and Top-4 basis: the upward-only target rule, development of the runner-up rule, a retrospective evaluation, the prospective confirmation on a new natural cohort, full-population accounting, and the external, output-space, and gate controls. Evidence levels are stated for each.

\subsection{Deployable Target-Policy Timeline}
The first deployable target rule was the upward-only rule
\[
 t_{\rm up}=\arg\max_{n>b}p(n),
\]
with eight natural abstentions at the maximum numeral. On 623 targetable examples, \mea{} reaches 417 assigned targets (66.93\%) with zero projected-gradient abstentions. The forced-deployment replay nevertheless changes accuracy by $-18.542$ percentage points, with 17 repairs against 134 breaks. Retaining the old fixed Unified gate reduces the intervention count to 52/631 and breaks to four, but the upward-only controller produces only one repair. In the gated cohort, $t_{\rm up}$ equals ground truth on only 1/48 targetable examples, and on none of the 24 targetable reference-router repairs.

The runner-up rule in Eq.~\eqref{eq:runnerup_target} was then evaluated on development OOF data before the runner-up controller was run. Under the unified hierarchical family, 30/292 development examples are gated decisions; the runner-up target matches ground truth on 24/30=80.0\% and on 15/16=93.75\% of reference-router repairs. For the Whole-only hierarchical family, the corresponding rates are 17/21=80.95\% and 10/12=83.33\%. Moreover, 28/30 development gated errors require a \emph{downward} correction even though Stage~2 was invoked by an upward proposal. On this basis the runner-up rule was fixed.

\subsection{Retrospective Evaluation of Runner-up Targeting}
The runner-up controller leaves the hierarchical \ugr{} apply/no-op gate fixed and executes \mea{} only on the 52 gated examples. The remaining 579 retain the identity prediction. Table~\ref{tab:mea_retrospective} records the paired end-to-end result.

\ifaclstyle\begin{table*}[!tp]\else\begin{table}[H]\fi
\centering\fontsize{8.5}{10.5}\selectfont
\caption{Retrospective final-population targeted-deployment comparison on the $N=631$ natural Stage-2 cohort.}
\label{tab:mea_retrospective}
\begin{tabular}{@{}lrrrrr@{}}
\toprule
\bfseries Policy & \bfseries Correct & \bfseries Accuracy & \bfseries $\Delta$ vs ID & \bfseries Repairs & \bfseries Breaks \\
\midrule
Identity & 201 & 31.854\% & -- & -- & -- \\
Reference \ugr{} & 223 & 35.341\% & $+3.487\pp$ & 26 & 4 \\
Runner-up + MEA & 234 & 37.084\% & $+5.230\pp$ & 38 & 5 \\
\bottomrule
\end{tabular}
\ifaclstyle\end{table*}\else\end{table}\fi

Versus identity, runner-up + \mea{} has exact McNemar $p=2.50\times10^{-7}$. Directly against the reference \ugr{}, 15 examples are correct only under runner-up + \mea{} and four only under the reference \ugr{}, giving $+1.743\pp$ and exact paired $p=0.01921$. On the 52 gated examples, the deploy target equals ground truth on 37/52=71.15\%, 50/52=96.15\% assigned targets are reached, all 37/37 correct targets are reached, there are zero projected-gradient abstentions, median iterations are 7, and mean iterations are 7.365.

\subsection{Prospective Confirmation on a New Natural Cohort}
After the runner-up rule is fixed, a separate $N=7{,}992$ natural cohort is finalized before outcome execution, with 3,996 simple and 3,996 complex questions, zero historical question/image-ID/pixel overlap, and zero within-cohort pixel duplicates. Stage~1 yields 630 upward proposals. The unchanged hierarchical gate fires on 63; both the runner-up target plan and the reference-\ugr{} method/gain plan are fixed before the 63 outcome pairs are executed.

\ifaclstyle\begin{table*}[!tp]\else\begin{table}[H]\fi
\centering\fontsize{8.5}{10.5}\selectfont
\caption{Held-out prospective targeted-deployment confirmation on the $N=630$ natural Stage-2 population.}
\label{tab:mea_prospective}
\begin{tabular}{@{}lrrrrr@{}}
\toprule
\bfseries Policy & \bfseries Correct & \bfseries Accuracy & \bfseries $\Delta$ vs ID & \bfseries Repairs & \bfseries Breaks \\
\midrule
Identity & 197 & 31.270\% & -- & -- & -- \\
Reference \ugr{} & 217 & 34.444\% & $+3.175\pp$ & 23 & 3 \\
Runner-up + MEA & 244 & 38.730\% & $+7.460\pp$ & 51 & 4 \\
\bottomrule
\end{tabular}
\ifaclstyle\end{table*}\else\end{table}\fi

Directly against the reference \ugr{}, the prospective controller is uniquely correct on 29 examples and the reference \ugr{} on two, yielding $+4.286\pp$ and exact two-sided McNemar $p=4.63\times10^{-7}$. The runner-up target equals ground truth on 51/63=80.95\% of gated examples; 61/63=96.83\% assigned targets are reached, including all 51/51 correct targets. Median/mean MEA iterations are 9/8.238, projected-gradient abstentions are zero, and the re-executed baseline matches the original baseline on all 63 gated examples.
The unmodified model attains 65.153\% accuracy on the full cohort, and the gate fires on 63/630=10.0\% of the upward proposals. Relative to identity, the reference \ugr{} nets $+20$ examples and targeted \mea{} nets $+47$. The prospective effect is larger than the earlier retrospective $+1.743\pp$ estimate; the cross-cohort difference is descriptive.

\paragraph{Estimands and comparator fairness.}
The final comparison of the reference \ugr{} with \mea{} estimates \emph{deployment utility under two fixed post-gate policies}, each at its own realized intervention size. Both policies receive the same fixed gate and therefore act on the identical gated set, but the reference \ugr{} executes its fixed ranked geometry/gain action whereas MEA executes a fixed explicit target with iterative re-linearization. Their realized intervention paths can consequently differ by design. Questions about whether re-linearization itself is useful under matched intervention magnitude are answered separately by the oracle-MEA controls in Sec.~\ref{sec:res_mea}, where fixed-direction and activation-aligned baselines receive the exact realized MEA step norms. The prospective $+4.286\pp$ result is therefore a policy-level utility improvement; efficiency per unit intervention norm is measured by the matched-budget controls, and compute is reported in Appendix~\ref{app:add_cost}.

As noted in \S\ref{sec:routing}, $\alpha_{\rm prop}>1$ triggers Stage~2 and sets a reference budget, and the final prediction may move in either direction; on natural data, most development gated errors require a downward correction.

\subsection{What Is Confirmed and What Is Reconstructed}
The prespecified primary targeted-deployment comparison remains runner-up + \mea{} versus the reference \ugr{} on the same prespecified $N=630$ upward-proposal Stage-2 population. Its paired outcomes (244 versus 217 correct, 29 MEA-only versus two reference-only, exact $p=4.63\times10^{-7}$) are unaffected by the supplementary analyses. The full-cohort reconstruction is \emph{derived after outcome execution} from the identical locked per-example outputs, and the external controls below were developed after the prospective Stage-2 outcomes were available.

\paragraph{Identity-fallback estimand and integrity.}
Let $\mathcal{P}$ denote the $N=7{,}992$ held-out natural cohort; $\mathcal{U}$ its $N=630$ fixed upward-proposal branch, and $\mathcal{A}\subset\mathcal{U}$ the $N=63$ fixed gated set. Let $b_i$ be the stored unmodified model prediction and $m_i$ the prespecified \mea{} prediction on the gated set. We define an \emph{identity-fallback selective intervention policy} on $\mathcal{P}$,
\begin{equation}
\widehat{y}_i^{\mathrm{sel}}=\begin{cases}
m_i,&i\in\mathcal{A},\\
b_i,&i\notin\mathcal{A}.
\end{cases}
\label{eq:identity_fallback_reconstruction}
\end{equation}
This definition retains identity for all non-gated examples, including non-upward Stage-1 proposals. The $\mathcal{U}$ cohort exactly matches the 630 upward proposals in Stage~1, all 630 Stage-1/Stage-2 identity predictions agree, and the Stage-2 runner-up + \mea{} prediction equals identity on all 567 non-gated examples. Therefore the full-population correct count is uniquely determined by the stored predictions:
\begin{equation}
\begin{aligned}
C_{\mathrm{sel}}&=5207-197+244=5254,\\
\Delta_{\mathrm{sel-ID}}&=100\cdot\frac{47}{7992}=+0.59\pp.
\end{aligned}
\label{eq:full_pop_accounting}
\end{equation}
The active gate executes 63/7,992 interventions (0.788\% of the population). Relative to identity, \mea{} yields 51 repairs and four breaks, or 47 net correct responses, and exact two-sided paired McNemar $p=2.05\times10^{-11}$ from the 51:4 discordance. This inference concerns selective prediction accuracy; compute cost is reported in Appendix~\ref{app:add_cost}.

\subsection{Matched-Gate Output Control}
The gated population is strongly enriched for zero-count questions: 51/63 (80.95\%) true answers are zero and 12/63 are nonzero. As an explicit simple control, retain the same AROMA gate and the same identity predictions on non-gated examples, but emit the numeral zero on the gated set. This policy is correct on 51/63 gated examples and 5,253/7,992 full-cohort examples (65.728\%), compared with 52/63 and 5,254/7,992 (65.741\%) for targeted \mea{}. There are five \mea{}-only and four zero-only correct outcomes (exact two-sided McNemar $p=1$). On the gated set, \mea{} is correct on 47/51 true-zero examples and 5/12 nonzero examples, whereas the zero policy is correct on 51/51 and 0/12. The deployment gain over identity is concentrated in zero-count repair, and the matched-budget and oracle-target experiments evaluate the control law on endpoints independent of this target prior.

\ifaclstyle\begin{table*}[!tp]\else\begin{table}[H]\fi
\centering\fontsize{8.5}{10.5}\selectfont
\caption{Audited identity-fallback policies across $N=7{,}992$ and the same fixed gated set. Gate-conditional zero, gate-conditional runner-up emission, and global emit-zero are post hoc stress controls computed from stored outputs.}
\label{tab:fullpop_audit_appendix}
\begin{tabular}{@{}lrrr@{}}
\toprule
\bfseries Prediction policy & \bfseries Gated correct & \bfseries Full correct & \bfseries Full accuracy \\
\midrule
Global emit-zero (no gate) & -- & 1,264 & 15.82\% \\
Unmodified identity & 5/63 & 5,207 & 65.15\% \\
Reference \ugr{} with matched gate & 25/63 & 5,227 & 65.40\% \\
Gate-conditional zero & 51/63 & 5,253 & 65.73\% \\
Gate-conditional runner-up emission & 51/63 & 5,253 & 65.73\% \\
Runner-up + MEA & 52/63 & 5,254 & 65.74\% \\
\bottomrule
\end{tabular}
\ifaclstyle\end{table*}\else\end{table}\fi

\subsection{External Steering Controls}
\label{app:external_controls}
GrAInS \citep{nguyen2025grains} derives steering directions from gradient-based attributions, token ablation, and representation contrasts. We study a \emph{counting-adapted GrAInS-inspired} method and a separate \emph{minimal multimodal compatibility port}; both are task-adapted implementations for counting on \modelname{}. All comparisons use the identical fixed Stage-2 question set, and both external configurations are fixed without using test ground truth as an inference feature. Their development and baseline selection followed AROMA's prospective evaluation, and they are reported as supplemental comparisons.

\paragraph{Counting-adapted GrAInS-inspired steering.}
The earlier adapted implementation uses separate 40-example fitting and 48-example calibration development splits, task-adapted numerical contrastive objectives and token ablation, and a calibrated steering direction/gain on the counting task. Under the same $\mathcal{A}$ intervention set, it predicts 33/63 correctly, compared with \mea{}'s 52/63. With identical identity fallback on non-gated examples, the GrAInS-inspired policy is correct on 225/630 (35.714\%) and targeted \mea{} on 244/630 (38.730\%): a $+3.016\pp$ matched-gate difference, 23 MEA-only versus four GrAInS-only correct outcomes, exact two-sided McNemar $p=3.11\times10^{-4}$. When the adapted method instead steers \emph{all} 630 Stage-2 examples, it reaches 247/630 (39.206\%), versus \mea{}'s 244/630 (38.730\%); the direct global-policy paired discordance is 33 MEA-only versus 36 GrAInS-only ($p=0.810$).

\paragraph{Image-preserving GrAInS minimal port.}
A second method follows the source-level attribution--ablation--representation-difference design more closely while adapting its image-conditioned forward pass to \modelname{}. Starting from the public GrAInS implementation, the compatibility port uses zero-embedding-baseline contrastive Integrated Gradients with 20 Gauss--Legendre steps, separate positive/negative signed token attribution, PAD/UNK token ablation, L31 representation differences, and principal-component analysis of 80 difference vectors constructed from 40 isolated development pairs. Before fitting, we set Top-$K=3$ (the official function-supported default), because the example-script $K=50$ would ablate every eligible text token in the short counting prompts; the median eligible-token ablation rate becomes 13.64\%. A disjoint 48-example calibration split selects $\alpha=0.3$ over $\alpha=1$ by the predeclared smaller-gain tie break (both 22/48 correct), while keeping direction, layer, attribution, and selection rule fixed. GrAInS-style sequence-norm-preserving steering is retained. On the Stage-2 set, the calibrated port reaches 201/630 (31.905\%), or 198/630 (31.429\%) under the same gate and BF16 identity fallback, with 6/63 correct on the gated set. Its own FP32 identity attains 198/630 (31.429\%); calibrated steering versus that FP32 identity has exact paired $p=0.581$. The uncalibrated demonstration $\alpha=10$ reaches 92/630 (14.603\%). The FP32 port and BF16 AROMA identity disagree on 17/630 discrete outputs.

\paragraph{CAA-style correctness-contrast steering.}
A third comparator follows contrastive activation addition \citep{rimsky2024caa} at the same site as \mncsa{} (the L18H13 output-projection input, applied to all tokens). Its direction contrasts pooled activations of correct and incorrect, count-matched natural development examples ($N=292$ Stage-2 development population, five-fold out-of-fold selection over the 30 development gated rows) and is applied as $\Delta h=2.0(\alpha-1)\|h\|_2\,d$ on the 63 gated examples with identity elsewhere. It reaches 207/630, below both targeted \mea{} (244/630; 4:41, exact $p=9.3\times10^{-9}$) and the reference \ugr{} (217/630; 6:16, $p=0.052$).

\ifaclstyle\begin{table*}[!tp]\else\begin{table}[H]\fi
\centering\fontsize{8.5}{10.5}\selectfont
\caption{Supplemental external steering comparisons on the fixed $N=630$ Stage-2 cohort. ``Same-gate'' always retains the BF16 identity prediction on the 567 non-gated examples, whereas ``global'' steers all 630. The gated cohort (63 samples) is chosen by AROMA's gate, not by an external-method-independent randomization.}
\label{tab:grains_external_controls}
\begin{tabular}{@{}lrr@{}}
\toprule
\bfseries Policy / intervention scope & \bfseries Global correct / 630 & \bfseries Gated correct / 63 \\
\midrule
Identity (BF16, no steering) & 197 (31.270\%) & 5 (7.937\%) \\
Reference \ugr{} & 217 (34.444\%) & 25 (39.683\%) \\
\textbf{AROMA runner-up + MEA} & \textbf{244 (38.730\%)} & \textbf{52 (82.540\%)} \\
GrAInS-inspired, counting-adapted, same-gate & 225 (35.714\%) & 33 (52.381\%) \\
GrAInS-inspired, counting-adapted, global & 247 (39.206\%) & 33 (52.381\%) \\
GrAInS minimal port, same-gate & 198 (31.429\%) & 6 (9.524\%) \\
GrAInS minimal port, global ($\alpha=.3$) & 201 (31.905\%) & 6 (9.524\%) \\
GrAInS minimal port, global ($\alpha=10$) & 92 (14.603\%) & 3 (4.762\%) \\
CAA-style correctness-contrast steering, same-gate & 207 (32.857\%) & 15 (23.810\%) \\
Runner-up emission without edit, same-gate & 243 (38.571\%) & 51 (80.952\%) \\
Confidence gate (matched firing rate) + runner-up emission & 198 (31.429\%) & -- \\
Always zero, Stage-2 global & 307 (48.730\%) & 51 (80.952\%) \\
\bottomrule
\end{tabular}
\ifaclstyle\end{table*}\else\end{table}\fi

\paragraph{Interpretive hierarchy.}
The strictest prospective deployment evidence is \mea{} versus the reference \ugr{} policy on the held-out Stage-2 cohort. The oracle-target and matched-budget experiments identify the value of iterative re-linearization given supplied targets and displacement norms. The GrAInS-family comparisons show that fixed-direction steering, adaptation, and deployment scope yield very different accuracies. The gate-conditional zero control shows that, on this cohort, target construction and gating are the operative deployment variables. This separation of \emph{control capacity}, \emph{target construction}, and \emph{deployment policy} is the scientific interpretation of AROMA.

\subsection{Output-Space and Gate Controls}
\label{app:add_output}
Table~\ref{tab:add_output} recomputes every deployment policy from the per-example outputs. The confidence gate replaces the learned gate with a threshold on $\max_n p(n\mid x)$, set on the natural Stage-2 development population ($N=292$) to fire on 10\% of examples before evaluation (it fires on 10.8\% of the 630).

\ifaclstyle\begin{table*}[!tp]\else\begin{table}[H]\fi
\centering\fontsize{8.5}{10.5}\selectfont
\caption{Deployment policies on the gated set, the Stage-2 population, and the full natural cohort. Non-gated examples keep the identity prediction unless the policy is global. Discordance is policy-only : \mea{}-only on the 630.}
\label{tab:add_output}
\setlength{\tabcolsep}{3.5pt}
\begin{tabular}{@{}lrrrrr@{}}
\toprule
\bfseries Policy & \bfseries Gated /63 & \bfseries Stage-2 /630 & \bfseries Full /7,992 & \bfseries Disc. & \bfseries $p$ \\
\midrule
Identity & 5 & 197 & 5,207 & 4:51 & $2.0\times10^{-11}$ \\
Reference \ugr{} (fixed) & 25 & 217 & 5,227 & 2:29 & $4.6\times10^{-7}$ \\
\textbf{Runner-up + \mea{} (fixed)} & \textbf{52} & \textbf{244} & \textbf{5,254} & -- & -- \\
Runner-up emission, no edit & 51 & 243 & 5,253 & 0:1 & 1.0 \\
Gate-conditional zero & 51 & 243 & 5,253 & 4:5 & 1.0 \\
GrAInS-inspired (counting-adapted), same gate & 33 & 225 & 5,235 & 4:23 & $3.1\times10^{-4}$ \\
GrAInS minimal port, same gate & 6 & 198 & -- & -- & -- \\
Confidence gate + runner-up emission & -- & 198 & -- & -- & -- \\
Global emit-zero & -- & 307 & 1,264 & 130:67 & $8.5\times10^{-6}$ \\
\bottomrule
\end{tabular}
\ifaclstyle\end{table*}\else\end{table}\fi

The learned gate fires on 63 examples, of which 51 (81\%) have true count zero, compared with 48.7\% of the Stage-2 population and 15.8\% of the full cohort. The runner-up target is zero on 51 of the 63 rows, 47 of which have true count zero and 4 a non-zero count; runner-up emission is correct on those 47 zero-count rows and on 4 non-zero rows, and \mea{} is correct on 47/51 zero-count and 5/12 non-zero rows. Stratifying by ground truth places the net gain over identity on zero-count questions. Global emit-zero exploits the zero enrichment of the Stage-2 population but collapses on the full cohort, and the matched-rate confidence gate falls to 198/630, so the gain depends on what the learned gate selects rather than on abstaining under low confidence. The GT-in-top-2 oracle on the gated set, an upper bound for runner-up-style targets, is 248/630.

\subsection{Gate Recall and Non-Learned Gates}
\label{app:add_gate}
The 7,992 cohort contains 638 zero-count overcounts (true count zero, identity prediction non-zero). Only 177 fall in the Stage-2 branch; the other 461 lie in examples for which fixed Stage~1 proposes $\alpha=0$ (134) or identity (327), where the second-stage gate cannot fire. The learned gate selects 51 overcounts (recall 28.8\% within the branch, 8.0\% overall), and \mea{} repairs 47 of them.

Two non-learned gates were specified in advance and evaluated once (protocol fixed before evaluation; thresholds and the noun-phrase parser fixed on the $N=292$ natural development population). The p(0) gate fires when the identity prediction is non-zero and $p(0\mid x)>0.180$, a threshold that matches the learned gate's 30/292 development firing count; on development data 24 of its 30 firings coincide with the learned gate. The existence gate asks the frozen model ``Is there any $\langle$object$\rangle$ in the image? Answer yes or no.'', following the polling format of POPE \citep{li2023pope}, with the object phrase parsed from the question (11/7,992 parse failures) and fires on ``no''. For each gate, the primary configuration emits the runner-up numeral; emitting zero and applying the gate to all 7,992 inputs are secondary configurations without multiplicity correction. A third gate, $p(0\mid x)>0.848$ without the non-zero condition, fires only where identity already answers zero and is uninformative.

\ifaclstyle\begin{table*}[!tp]\else\begin{table}[H]\fi
\centering\fontsize{8.5}{10.5}\selectfont
\caption{Gate controls on the Stage-2 population and, for secondary configurations, on the full cohort. Precision is the fraction of fired rows with true count zero. Discordance is policy-only : \mea{}-only; $p$ is exact McNemar against runner-up + \mea{}.}
\label{tab:add_gate}
\setlength{\tabcolsep}{3pt}
\begin{tabular}{@{}lrrrrrrr@{}}
\toprule
& \multicolumn{4}{c}{\bfseries Stage-2 population ($N=630$)} & \multicolumn{3}{c}{\bfseries Full cohort ($N=7{,}992$)} \\
\cmidrule(lr){2-5}\cmidrule(l){6-8}
\bfseries Gate + output & \bfseries Fires & \bfseries Prec. & \bfseries Correct & \bfseries Disc., $p$ & \bfseries Fires & \bfseries Correct & \bfseries Disc., $p$ \\
\midrule
Learned gate + \mea{} (fixed) & 63 & 0.810 & 244 & -- & 63 & 5,254 & -- \\
Identity & 0 & -- & 197 & -- & 0 & 5,207 & -- \\
Learned gate + runner-up & 63 & 0.810 & 243 & 0:1, 1.0 & 63 & 5,253 & 0:1, 1.0 \\
\midrule
p(0) gate + runner-up (primary) & 68 & 0.868 & 251 & 19:12, 0.28 & 143 & 5,317 & 79:16, $3.5\times10^{-11}$ \\
p(0) gate + zero & 68 & 0.868 & 252 & 20:12, 0.22 & 143 & 5,326 & 88:16, $3.2\times10^{-13}$ \\
Existence gate + runner-up (primary) & 174 & 0.839 & 151 & 34:127, $7.9\times10^{-14}$ & 699 & 4,885 & 110:479, $8\times10^{-56}$ \\
Existence gate + zero & 174 & 0.839 & 253 & 57:48, 0.44 & 699 & 5,311 & 185:128, $1.5\times10^{-3}$ \\
\bottomrule
\end{tabular}
\ifaclstyle\end{table*}\else\end{table}\fi

On the Stage-2 branch the p(0) gate is statistically indistinguishable from the learned gate, while the existence question paired with runner-up emission fails because the runner-up of a confident non-zero count is rarely zero. Over the full cohort, gates that are not restricted to the Stage-2 branch reach overcounts the learned gate cannot see. These post hoc comparisons leave the prospective endpoint unchanged and show that the learned gate recovers a zero-count signal already present in the model's own numeral distribution, and that a broader gate is the natural route to larger population-level gains with the same actuator.

\FloatBarrier
\Needspace{6\baselineskip}
\section{Off-Target Stress Test and Finite-Step Response Diagnostics}
\label{app:offtarget}

\subsection{GQA Off-Target Stress Test}
\label{app:gqa_offtarget}
The counting experiments establish target-task controllability, but an inference-time intervention should also be examined for collateral changes outside the behavior used to discover it. We therefore fix a non-counting GQA stress test on the first 1,000 baseline-correct examples in candidate order and compare L18H13 MN against WHOLE under seven matched finite displacement budgets. Because the population is baseline-correct by construction, the endpoint is \emph{break burden}: the fraction of geometry--budget conditions that turn a correct non-counting answer into an error.

Across the 7,000 matched conditions per family, WHOLE produces 101 breaks (1.44\%) and MN produces 83 (1.19\%), an aggregate MN$-$WHOLE difference of $-0.26\pp$. The paired bootstrap 95\% CI is $[-0.6429,+0.1286]\pp$. Six of seven individual budget levels favor MN; the per-budget differences in MN$-$WHOLE break rate are $+0.60,\allowbreak-0.10,\allowbreak-0.40,\allowbreak-0.60,\allowbreak-0.20,\allowbreak-0.40,\allowbreak-0.70\pp$ from $\lambda=0.5$ through 2.0. A later 1,000,000-draw paired sign-flip analysis gives two-sided $p=0.2023$, and repeated bootstrap seeds leave the confidence interval essentially unchanged. Both geometries therefore preserve the large majority of baseline-correct non-counting behavior, with a stable directional trend toward lower MN collateral burden.

This stress test complements the target-task specificity controls. The Bottom-4/random-subspace experiments ask whether the learned low-rank directions are behaviorally privileged for counting; GQA asks whether a matched finite edit perturbs already-correct behavior outside the discovery task. Together they separate \emph{target efficacy} from \emph{off-target preservation}.

\ifaclstyle\begin{table*}[!tp]\else\begin{table}[H]\fi
\centering\fontsize{8.5}{10.5}\selectfont
\caption{GQA off-target break-burden summary. Difference is MN$-$WHOLE; negative values favor MN.}
\label{tab:gqa_offtarget_budget}
\begin{tabular}{@{}lrrrrrrrr@{}}
\toprule
$\lambda$ & 0.50 & 0.75 & 1.00 & 1.25 & 1.50 & 1.75 & 2.00 & Aggregate \\
\midrule
MN$-$WHOLE break rate (pp) & +0.60 & -0.10 & -0.40 & -0.60 & -0.20 & -0.40 & -0.70 & -0.257 \\
\bottomrule
\end{tabular}
\ifaclstyle\end{table*}\else\end{table}\fi

MN has 18 fewer breaks than WHOLE. A 200,000-repetition bootstrap gives $P(\Delta<0)=0.903$; five additional 100,000-repetition seeds reproduce essentially the same interval and $P(\Delta<0)\approx0.903$. The interval is therefore stable across Monte Carlo seeds.

Each stage of this test (baseline population, matched-intervention protocol, raw outcomes, endpoint analysis, and robustness analysis) is fixed before the next stage begins.

\subsection{Finite-Step Response on Natural Data}
\label{app:natural_response}
We next ask a post hoc question: \emph{what distinguishes useful finite movement from destructive finite movement?} These analyses reuse existing outcomes and do not change any policy.

\ifaclstyle\begin{figure*}[!tp]\else\begin{figure}[!htbp]\fi
\centering

\ifaclstyle\resizebox{\linewidth}{!}{\fi
\begin{tikzpicture}

\begin{groupplot}[
group style={
group size=3 by 1,
horizontal sep=1.50cm
},
scale only axis,width=0.25\textwidth,
height=3.0cm,
tick label style={font=\rmfamily\fontsize{7.5}{9}\selectfont},
label style={font=\rmfamily\fontsize{8}{9.5}\selectfont},
title style={font=\rmfamily\fontsize{8.5}{10.5}\selectfont\bfseries,at={(0,1)},anchor=south west,yshift=23pt},
axis x line*=bottom,axis y line*=left,tick align=outside,
axis line style={draw=aromaInk!60,line width=.45pt},
tick style={draw=aromaInk!60,line width=.4pt},
ymin=0,enlarge x limits=0.18,
legend image code/.code={\path[#1] (0pt,-2pt) rectangle (8pt,3pt);},
legend style={font=\rmfamily\fontsize{7.5}{9}\selectfont,draw=none,fill=none,at={(.5,1.025)},anchor=south,legend columns=-1},
ymajorgrids=true,
grid style={draw=aromaRule!30,line width=.35pt}
]
\nextgroupplot[
title={(a) Selective intervention},
ybar,
legend image code/.code={\path[#1] (0pt,-2pt) rectangle (8pt,3pt);},
bar width=8pt,
symbolic x coords={TallyQA,HoloCount},
xtick=data,
ylabel={Rate},
ymax=0.4,enlarge x limits=0.35
]
\addplot[draw=none,fill=aromaBlue] coordinates {
(TallyQA,0.20)
(HoloCount,0.3548)
};
\addplot[draw=none,fill=aromaGreen] coordinates {
(TallyQA,0.1625)
(HoloCount,0.0818)
};
\addplot[draw=none,fill=aromaOrange] coordinates {
(TallyQA,0.0563)
(HoloCount,0.0830)
};
\legend{
Active,
Repair,
Break
}
\nextgroupplot[
title={(b) Movement magnitude trade-off},
ybar,
legend image code/.code={\path[#1] (0pt,-2pt) rectangle (8pt,3pt);},
bar width=7pt,
symbolic x coords={
T-small,
T-medium,
T-large,
H-small,
H-medium,
H-large
},
xtick=data,
xticklabel style={
rotate=45,
anchor=east
},
ylabel={Rate},
ymax=0.30
]
\addplot[draw=none,fill=aromaGreen] coordinates {
(T-small,0.0524)
(T-medium,0.1955)
(T-large,0.2397)
(H-small,0.0136)
(H-medium,0.1229)
(H-large,0.1092)
};
\addplot[draw=none,fill=aromaOrange] coordinates {
(T-small,0.0075)
(T-medium,0.0451)
(T-large,0.1161)
(H-small,0.0170)
(H-medium,0.1024)
(H-large,0.1297)
};
\legend{
Repair,
Break
}
\nextgroupplot[
title={(c) Directional alignment},
ybar,
legend image code/.code={\path[#1] (0pt,-2pt) rectangle (8pt,3pt);},
bar width=10pt,
symbolic x coords={
T-away,
T-toward,
H-away,
H-toward
},
xtick=data,
xticklabel style={
rotate=45,
anchor=east
},
ylabel={Repair rate},
ymax=0.35
]
\addplot[draw=none,fill=aromaGreen] coordinates {
(T-away,0.0049)
(T-toward,0.3217)
(H-away,0.0099)
(H-toward,0.2262)
};
\end{groupplot}

\end{tikzpicture}%
\ifaclstyle}\fi

\caption{
\textbf{Finite-step response geometry explains selective repair and failure.}
(a) AROMA intervenes selectively rather than globally.
(b) Larger expected-numeral movement creates both repair opportunity and break risk.
(c) Repairs are strongly concentrated in target-aligned movement directions.
Target alignment is a post hoc diagnostic using ground truth and is not available at inference time.
}

\label{fig:finite_step_geometry}

\par\vspace{6pt}
\begin{minipage}{\textwidth}
\makeatletter\def\@captype{table}\makeatother
\centering
\small
\caption{\textbf{Post hoc finite-step response diagnostics on natural data.} ``Active'' means nonzero observed expected-numeral movement; on the final TallyQA controller this coincides with non-identity selection. Magnitude bins are tertiles within active examples. Target alignment is defined using ground truth and is therefore diagnostic only, not deployable.}
\label{tab:natural_response_diagnostics}
\setlength{\tabcolsep}{3.4pt}
\begin{tabularx}{\textwidth}{@{}lrrrYYY@{}}
\toprule
\bfseries Dataset & \bfseries Active / total & \bfseries Active repair & \bfseries Active break & \bfseries Repair: small/\allowbreak med./\allowbreak large & \bfseries Break: small/\allowbreak med./\allowbreak large & \bfseries Target-aligned / away repair \\
\midrule
TallyQA & 800 / 4000 (20.00\%) & 16.25\% & 5.63\% & 5.24 / 19.55 / 23.97\% & 0.75 / 4.51 / 11.61\% & 32.17\% / 0.49\% \\
HoloCount & 880 / 2480 (35.48\%) & 8.18\% & 8.30\% & 1.36 / 12.29 / 10.92\% & 1.70 / 10.24 / 12.97\% & 22.62\% / 0.99\% \\
\bottomrule
\end{tabularx}
\end{minipage}
\ifaclstyle\end{figure*}\else\end{figure}\fi

\paragraph{Selectivity and utility under transfer.}
On the final TallyQA confirmation, the 3,200 inactive identity examples are substantially easier: baseline and post-intervention accuracy are both $2402/3200=75.06\%$. The 800 active examples are much harder, with baseline accuracy $195/800=24.38\%$; after natural-domain recalibration their accuracy rises to $280/800=35.00\%$, reflecting 130 repairs against 45 breaks.
On HoloCount, 1,600 zero-movement examples likewise have higher baseline accuracy ($885/1600=55.31\%$) than the 880 active examples ($273/880=31.02\%$). However, active accuracy changes only to $272/880=30.91\%$ because repairs and breaks nearly cancel (72 versus 73). Thus the transferred policy still concentrates intervention on harder cases, but its action utility no longer transfers.
The mean active $|\Delta\mu|$ is 0.282 on TallyQA and 0.685 on HoloCount.

\paragraph{Movement magnitude.}
Among active natural examples, we partition $|\Delta\mu|$ into within-dataset tertiles. On TallyQA, repair rates rise from 5.24\% to 19.55\% to 23.97\% across small/medium/large movement, while break rates simultaneously rise from 0.75\% to 4.51\% to 11.61\%. On HoloCount, repair rises from 1.36\% to 12.29\% from small to medium movement but then falls to 10.92\% in the largest bin, while break rate continues to increase from 1.70\% to 10.24\% to 12.97\%. The external dataset therefore exposes an overshoot/instability boundary: stronger movement creates more opportunity for state change, not guaranteed improvement.

\paragraph{Where repairs occur.}
For analysis only, we call an active baseline error \emph{target-aligned} when the intervention reduces expected-numeral distance to the ground truth, $|\mu_{\rm post}-y|<|\mu_{\rm base}-y|$, and \emph{away} when it increases that distance. On TallyQA, 129/401 target-aligned active baseline errors are repaired (32.17\%) versus 1/204 away cases (0.49\%). On HoloCount, the corresponding repair rates are 69/305=22.62\% and 3/302=0.99\%. This diagnostic is intentionally not treated as a deployable routing feature: it uses $y$ and therefore cannot be available at inference time. Its value is explanatory---the discrete repairs produced by AROMA are strongly concentrated among interventions whose continuous expected-numeral movement is directionally consistent with the required correction.

\ifaclstyle\begin{table*}[!tp]\else\begin{table}[H]\fi
\centering\fontsize{8.5}{10.5}\selectfont
\caption{Natural active-response magnitude tertiles. Repair and break rates are fractions of all active examples in each magnitude bin, not conditional on baseline error/correctness.}
\label{tab:natural_magnitude_bins}
\begin{tabular}{@{}llrrrr@{}}
\toprule
\bfseries Dataset & \bfseries Bin & \bfseries $N$ & \bfseries Mean $|\Delta\mu|$ & \bfseries Repair & \bfseries Break \\
\midrule
TallyQA & small & 267 & 0.04150 & 5.24\% & 0.75\% \\
TallyQA & medium & 266 & 0.16862 & 19.55\% & 4.51\% \\
TallyQA & large & 267 & 0.63573 & 23.97\% & 11.61\% \\
\midrule
HoloCount & small & 294 & 0.06950 & 1.36\% & 1.70\% \\
HoloCount & medium & 293 & 0.31308 & 12.29\% & 10.24\% \\
HoloCount & large & 293 & 1.67425 & 10.92\% & 12.97\% \\
\bottomrule
\end{tabular}
\ifaclstyle\end{table*}\else\end{table}\fi

\subsection{Finite-Gain Response Curves on the Direction-Development Cohort}
\label{app:synthetic_response}
\paragraph{Synthetic response curves.}
On the 500-example direction-development cohort, increasing gain causes repairs and breaks to accumulate together rather than producing a single monotonic ``more is better'' regime. For example, MN accuracy increases from 50.8\% at identity to 55.6\% at gain 8, but the same sweep grows from 0/0 repairs/breaks at identity to 107/83 at gain 8; SA reaches 56.6\% with 99/70 repairs/breaks, while WHOLE reaches 59.6\% with 91/47. These development diagnostics help explain why the final system routes gain and geometry instead of selecting a universally maximal displacement.

\ifaclstyle\begin{table*}[!tp]\else\begin{table}[H]\fi
\centering\fontsize{8.5}{10.5}\selectfont
\caption{direction-development finite-gain response curve ($N=500$). $\Delta\mu$ is mean post-minus-baseline expected numeral. Repairs and breaks are counted relative to identity.}
\label{tab:sa_dev_response_curve}
\setlength{\tabcolsep}{2.4pt}
\begin{tabular}{@{}r*{3}{rrrr}@{}}
\toprule
& \multicolumn{4}{c}{\textbf{MN}} & \multicolumn{4}{c}{\textbf{SA}} & \multicolumn{4}{c}{\textbf{WHOLE}} \\
\cmidrule(lr){2-5}\cmidrule(lr){6-9}\cmidrule(l){10-13}
\textbf{Gain} & \textbf{Accuracy} & \textbf{Mean $\Delta\mu$} & \textbf{Repairs} & \textbf{Breaks} & \textbf{Accuracy} & \textbf{Mean $\Delta\mu$} & \textbf{Repairs} & \textbf{Breaks} & \textbf{Accuracy} & \textbf{Mean $\Delta\mu$} & \textbf{Repairs} & \textbf{Breaks} \\
\midrule
1.0 & 50.8\% & 0 & 0 & 0 & 50.8\% & 0 & 0 & 0 & 50.8\% & 0 & 0 & 0 \\
1.5 & 51.6\% & -0.0471 & 12 & 8 & 51.4\% & -0.107 & 11 & 8 & 51.4\% & -0.0249 & 13 & 10 \\
2.0 & 52.0\% & -0.0728 & 21 & 15 & 52.2\% & -0.195 & 20 & 13 & 50.8\% & -0.0633 & 20 & 20 \\
3.0 & 51.4\% & -0.0880 & 40 & 37 & 51.2\% & -0.337 & 33 & 31 & 49.6\% & -0.198 & 32 & 38 \\
4.0 & 54.2\% & -0.0684 & 64 & 47 & 54.4\% & -0.442 & 57 & 39 & 49.4\% & -0.395 & 36 & 43 \\
6.0 & 54.6\% & +0.169 & 83 & 64 & 55.2\% & -0.489 & 79 & 57 & 55.0\% & -0.718 & 67 & 46 \\
8.0 & 55.6\% & +0.709 & 107 & 83 & 56.6\% & -0.137 & 99 & 70 & 59.6\% & -0.847 & 91 & 47 \\
\bottomrule
\end{tabular}
\ifaclstyle\end{table*}\else\end{table}\fi

\subsection{Finite-Step Gradient-Drift Diagnostic}
\label{app:gradient_drift}
The gradient-drift diagnostic recomputes the gradient of correct-answer negative log-likelihood at the input to the L18H13 cross-attention output projection. The loss used by the diagnostic is
\begin{equation}
\mathcal L_{\rm correct}(x,y)=-\log p(y\mid x),
\end{equation}
and the head-slice gradient is pooled across token positions. A fixed supplied direction is then applied with displacement proportional to $(\alpha-1)\|h\|_2$, and the gradient is recomputed after the intervention. We report cosine with the identity-state gradient. This design probes drift of local objective geometry under a finite step.

\ifaclstyle\begin{table*}[!tp]\else\begin{table}[H]\fi
\centering\fontsize{8.5}{10.5}\selectfont
\caption{Finite-step gradient drift on the direction-development cohort. Values are cosine similarity between the post-intervention correct-answer-loss gradient and the identity-state gradient.}
\label{tab:gradient_drift}
\begin{tabular}{@{}lrrrrrrr@{}}
\toprule
\bfseries Direction label & \bfseries 1.0 & \bfseries 1.5 & \bfseries 2.0 & \bfseries 3.0 & \bfseries 4.0 & \bfseries 6.0 & \bfseries 8.0 \\
\midrule
MN mean & 1.000 & 0.918 & 0.866 & 0.734 & 0.629 & 0.418 & 0.136 \\
MN median & 1.000 & 0.995 & 0.981 & 0.930 & 0.857 & 0.621 & 0.129 \\
SA mean & 1.000 & 0.910 & 0.855 & 0.737 & 0.621 & 0.407 & 0.146 \\
SA median & 1.000 & 0.995 & 0.981 & 0.944 & 0.892 & 0.716 & 0.296 \\
\bottomrule
\end{tabular}
\ifaclstyle\end{table*}\else\end{table}\fi

For gains above identity, mean gradient norms remain on the order of $10^{-4}$ rather than collapsing to zero: approximately 0.000385--0.000438 for the MN-labeled runs and 0.000277--0.000406 for the SA-labeled runs over the recorded gain range. The central observation is therefore rotation of the local gradient direction, not disappearance of local sensitivity.
Local correct-answer-loss geometry changes substantially along large finite interventions, so a direction that is locally informative evolves after a large step. This provides mechanistic support for the local-versus-finite distinction of Gap~2.
\FloatBarrier
\Needspace{6\baselineskip}
\section{Extended Discussion, Scope of Claims, and Related Work}
\label{app:discussion}

\subsection{Extended Discussion}
\paragraph{Coverage, magnitude, and scientific interpretation.}
The held-out Stage-2 improvement is substantial within its fixed selection regime, and it remains positive and statistically supported over the complete natural cohort even though only 0.788\% of inputs are edited. In contrast, high oracle reachability and matched-budget re-linearization benefits isolate an actuator-control capability that aggregate accuracy alone cannot resolve. Together with the global GrAInS-inspired and gate-conditional-zero controls, this argues for reporting \emph{mechanistic reachability}, \emph{policy-level deployment utility}, and \emph{population-level gain} as separate endpoints.

\paragraph{Local sensitivity and finite control.}
Across Llama and IDEFICS, cardinality sensitivity is strongly concentrated in a low-dimensional subspace, yet that subspace contains comparatively little activation energy.
This mismatch is operational rather than merely spectral: preserving most local sensitivity does not by itself yield a finite edit of useful magnitude.
Magnitude normalization repairs one failure mode, but the IDEFICS rank screen further shows that capturing more sensitivity variance does not necessarily improve behavioral steering.
The finite-step diagnostics point in the same direction: as gain increases, the correct-answer-loss gradient rotates away from its identity-state orientation even though its norm remains nonzero.
Taken together, these results distinguish a locally informative sensitivity geometry from a behaviorally validated intervention geometry.
AROMA therefore treats spectral structure as a proposal for control, not as control evidence by itself.

\paragraph{Controllability and utility.}
Causal screening establishes that an internal component can move the target behavior; it does not determine whether a particular finite intervention will improve correctness on a particular example.
This distinction appears repeatedly.
Direction--gain orderings cross, a prospective sensitivity-only gain rule fails to outperform the learned candidates, and larger finite movement creates both repair opportunities and break risk.
The natural-domain diagnostics sharpen the same point: intervention is concentrated on harder examples, yet selective targeting alone is insufficient when action utility shifts, as on HoloCount.
Likewise, target-aligned expected-numeral movement strongly concentrates repairs, but this quantity depends on the ground-truth answer and is therefore explanatory rather than deployable.
These results motivate routing candidate correctness rather than responsiveness: a useful actuator still requires a policy that decides whether, how, and how strongly to intervene.

\paragraph{Geometry as a per-example decision.}
The low-rank Utility Router first shows that direction and gain should be selected conditionally even within a fixed intervention family.
Matched-budget MN, SA, and WHOLE trajectories then show no stable intrinsic geometry winner once finite gain is calibrated.
The later UGR experiments turn this ambiguity into a prospective policy question.
On procedural counting, fixed Unified routing improves the strongest development-selected restricted policy by $+2.21\pp$ on held-out data.
Under natural-domain shift, the same principle survives only after policy refactorization: the Unified action space exposes an additional $+6.181\pp$ of oracle capacity over Whole-only control, and the hierarchical router converts part of that capacity into a $+1.902\pp$ paired final gain.
The important conclusion is therefore not that MN, SA, or WHOLE is globally superior, but that intervention geometry itself is an example-conditional control variable.

\paragraph{Domain shift.}
Natural-domain transfer reveals that mechanism, geometry, and policy do not generalize as a single unit.
Fixed low-rank geometry retains positive utility on TallyQA, whereas the synthetic Proposal Controller becomes harmful until its utility mapping is recalibrated.
At Stage 2, a flat unified classifier and pairwise ranking alone fail to exploit the enlarged natural action space reliably; successful transfer requires a hierarchical factorization that first decides whether to intervene and only then ranks non-identity candidates.
After this architecture is selected on development data and fixed, it produces a statistically supported gain on the held-out natural Stage-2 population.
IDEFICS provides a complementary form of generalization: the actuator--geometry--gain hierarchy independently reappears without transferring Llama coordinates.
HoloCount, however, remains a boundary in which intervention headroom persists while the tested policy/readout does not recover positive utility.
Generalization is therefore factorized rather than binary: causal capacity can persist even when gain, abstention calibration, action ranking, or readout requires new domain-specific learning.

\paragraph{Separate decisions.}
The routing and targeted-control experiments form one decomposition. Causal localization supplies \emph{where}; MN, SA, and WHOLE routing select \emph{geometry and gain}; the hierarchical gate supplies \emph{when}; runner-up targeting supplies \emph{which target}; and \mea{} supplies \emph{how to reach} it. Keeping these roles separate clarifies why apparently contradictory results coexist: no geometry is globally best after gain calibration, the same actuator can be useful or destructive under different targets, and a strong target controller can coexist with a sparse deployment gate.

\paragraph{Residual bottleneck.}
The prospective gated set sharpens the diagnosis: target selection is correct on 80.95\% of examples, assigned-target reachability is 96.83\%, and all 51/51 correct targets are reached. The execution controller therefore leaves almost no observed residual error once given a correct target on this cohort. Improving target construction is consequently higher-leverage than making MEA more aggressive under the current protocol.

\subsection{Scope of Claims}
The evidence supports four findings: the actuator--low-rank-geometry--finite-control pattern independently replicates in two VLMs; procedural UGR and domain-developed hierarchical Natural UGR each receive held-out prospective confirmation on upward proposals; ground-truth-free runner-up targeting plus fixed-gate MEA receives a separate held-out natural deployment confirmation in Llama, and its edits change the model's own generated answers; and model/domain-specific gain, target policy, and routing calibration are separate problems. At equal displacement, the sensitivity subspace is additionally the most effective of the tested control geometries. The findings are established at three levels of specificity. (i) Model level: the exact actuator identity, basis vectors, and $r=4,\beta=2$ configuration are model-specific; subspaces are found independently in each model, and causal intervention identifies the mechanism. (ii) Domain and policy level: policies and procedural UGR parameters are recalibrated per natural dataset, Llama \mncsa{} transfer is established on the reported fixed confirmations, and second-stage routing is evaluated on upward proposals in Llama. (iii) Task level: the results concern visual counting and the current linear/hierarchical routing representations.

\subsection{Closest VLM Intervention Work}
\label{app:closest_work}
Table~\ref{tab:closest_intervention} summarizes the \emph{control object emphasized} by nearby VLM intervention work rather than scoring methods by checkmarks. The comparison is deliberately descriptive: different papers optimize different behaviors, models, and evaluation regimes.

\ifaclstyle\begin{table*}[!tp]\else\begin{table}[H]\fi
\centering
\fontsize{8}{9.5}\selectfont
\caption{Descriptive positioning against closely related VLM intervention work. The table records each method's main intervention/control emphasis; it is not a quality ranking.}
\label{tab:closest_intervention}
\setlength{\tabcolsep}{3pt}
\begin{tabularx}{\linewidth}{@{}P{3.0cm}P{3.0cm}P{3.7cm}Y@{}}
\toprule
\bfseries Method & \bfseries Localization / intervention object & \bfseries Conditional control emphasis & \bfseries Evaluation emphasis \\
\midrule
NOTICE \citep{golovanevsky2025notice} & causally localized multimodal components & diagnostic localization & corruption-based mechanism analysis \\
V-SEAM \citep{wang2025vseam} & causal visual-semantic components & intervention after attribution & editing / attention modulation \\
SteerVLM \citep{sivakumar2025steervlm} & hidden activation dimensions & lightweight adaptive steering & robustness / off-target behavior \\
CAST \citep{lee2025cast} & steering vectors & conditional activation steering & condition-dependent deployment \\
ITI \citep{li2023iti} & probe directions at selected attention heads & fixed-strength head steering & truthfulness under steering \\
CAA \citep{rimsky2024caa} & contrastive residual-stream directions & signed steering multiplier & behavior shift in multiple-choice and open generation \\
ReFT / DAS \citep{wu2024reft,geiger2024das} & learned low-rank subspace & supervised representation intervention & task performance / causal alignment \\
Rudman et al. \citep{rudman2026prompt} & hallucination-relevant mechanisms & causal intervention & prompt-induced hallucination analysis \\
VIB-Probe \citep{zhang2026vib} & probe-derived hallucination signal & mitigation after detection & hallucination correction \\
HONES \citep{wang2026hones} & head/neuron causal structure & neuron-level steering & multi-task causal attribution \\
\method{} & validated causal actuator $\rightarrow$ low-rank control subspace & geometry/gain routing plus explicit target selection and iterative target execution & held-out mechanism, routing, targeted-deployment, transfer, and off-target boundaries \\
\bottomrule
\end{tabularx}
\ifaclstyle\end{table*}\else\end{table}\fi

\paragraph{From steering directions to closed-loop control.}
Most activation-steering pipelines couple a direction and magnitude into one intervention object. \method{} instead treats intervention geometry, finite gain, target selection, target execution, and deployment gating as separately falsifiable stages. The distinction matters empirically: a target can be highly reachable and still destroy task utility, while a fixed gate plus a prospectively confirmed target policy can convert the same actuator into a useful controller. This mechanism--target--policy decomposition is the main conceptual difference from treating steering as a fixed vector edit.
\FloatBarrier
\Needspace{6\baselineskip}
\section{Reproducibility}
\label{app:reproducibility}
Each evaluation stage is tied to a fixed protocol and stored results, including the exact model revision, controller and Top-4 basis hashes, null and random-subspace seeds, and dataset manifests.
Experiments use Python 3.11, PyTorch 2.11 with CUDA 12.8, Transformers 5.17, Accelerate 1.15, NumPy 2.4, Pandas 3.0, SciPy 1.17, scikit-learn 1.9, Pillow 12.3, and Matplotlib 3.11 on NVIDIA Blackwell GPUs (B200 and RTX PRO 6000 Server Edition).
The cross-model IDEFICS runs use the pinned revision \path{7065afb8a5bb03015700b315f6751dc3dd4e16eb}; the IDEFICS Top-4 basis, its source gradient matrix, and the split manifests are stored with SHA256 checksums.

Before each evaluation run, scripted checks verify the controller, model revision, manifests, image hashes, the absence of ground-truth information in inference features, and the identity intervention.
The v4 set was generated only after the CSA algorithm was fixed; a pre-inference check verified 2,000 unique sample IDs, seeds, and image SHA256 hashes, exact count/condition balance, and zero seed or image-hash overlap with v1--v3.
Seed or pixel collisions during generation were resolved by regeneration under a fixed uniqueness rule; no sample was filtered or replaced using model behavior.
Where per-example action matrices are available, the reported tables and plots can be regenerated without rerunning the VLM.
Recorded compute totals at least 41.5 GPU-hours, of which the additional analyses in Appendix~\ref{app:additional} account for 19.1.


\begin{thebibliography}{99}

\bibitem[Acharya et~al.(2019)Acharya, Kafle, and Kanan]{acharya2019tallyqa}
Manoj Acharya, Kushal Kafle, and Christopher Kanan.
\newblock TallyQA: Answering Complex Counting Questions.
\newblock In \emph{AAAI}, pages 8076--8084, 2019.

\bibitem[Alayrac et~al.(2022)Alayrac et al.]{alayrac2022flamingo}
Jean-Baptiste Alayrac, Jeff Donahue, Pauline Luc, Antoine Miech, Iain Barr, Yana Hasson, Karel Lenc, Arthur Mensch, Katherine Millican, Malcolm Reynolds, Roman Ring, Eliza Rutherford, Serkan Cabi, Tengda Han, Zhitao Gong, Sina Samangooei, Marianne Monteiro, Jacob Menick, Sebastian Borgeaud, Andrew Brock, Aida Nematzadeh, Sahand Sharifzadeh, Mikolaj Binkowski, Ricardo Barreira, Oriol Vinyals, Andrew Zisserman, and Karen Simonyan.
\newblock Flamingo: a Visual Language Model for Few-Shot Learning.
\newblock In \emph{NeurIPS}, 2022.

\bibitem[Alzahrani et~al.(2026)Alzahrani et al.]{alzahrani2026countercount}
Reem Alzahrani, Hassan Alshanqiti, Bushra Bin Hemid, Zaid Alyafeai, Abdelrahman Eldesokey, and Bernard Ghanem.
\newblock CounterCount: A Diagnostic Framework for Counting Bias in Vision Language Models.
\newblock arXiv:2605.17826, 2026.

\bibitem[Che et~al.(2026)Che et al.]{che2026counting}
Liwei Che, Zhiyu Xue, Yihao Quan, Benlin Liu, Zeru Shi, Michelle Hurst, Jacob Feldman, Ruixiang Tang, Ranjay Krishna, and Vladimir Pavlovic.
\newblock Counting Circuits: Mechanistic Interpretability of Visual Reasoning in Large Vision-Language Models.
\newblock arXiv:2603.18523, 2026.

\bibitem[Deng et~al.(2026)Deng, Qiao, and Wan]{deng2026holocount}
Jinhong Deng, Limeng Qiao, and Guanglu Wan.
\newblock HoloCount: A Holistic Visual Counting Benchmark for MLLMs.
\newblock arXiv:2607.06420, 2026.

\bibitem[El-Shangiti et~al.(2026)El-Shangiti et al.]{elshangiti2026count}
Ahmed Oumar El-Shangiti, Abzal Nurgazy, Hilal AlQuabeh, Nikolai Rozanov, and Kentaro Inui.
\newblock The Count Is There, but Misaligned: Understanding and Correcting Counting Failures in VLMs.
\newblock arXiv:2607.09544, 2026.

\bibitem[Geiger et~al.(2024)Geiger, Wu, Potts, Icard, and Goodman]{geiger2024das}
Atticus Geiger, Zhengxuan Wu, Christopher Potts, Thomas Icard, and Noah Goodman.
\newblock Finding Alignments Between Interpretable Causal Variables and Distributed Neural Representations.
\newblock In \emph{Proceedings of the Third Conference on Causal Learning and Reasoning}, PMLR 236, pages 160--187, 2024.

\bibitem[Golovanevsky et~al.(2025)Golovanevsky, Rudman, Palit, Eickhoff, and Singh]{golovanevsky2025notice}
Michal Golovanevsky, William Rudman, Vedant Palit, Carsten Eickhoff, and Ritambhara Singh.
\newblock What Do VLMs NOTICE? A Mechanistic Interpretability Pipeline for Gaussian-Noise-free Text-Image Corruption and Evaluation.
\newblock In \emph{NAACL}, pages 11462--11482, 2025.

\bibitem[Goyal et~al.(2017)Goyal, Khot, Summers-Stay, Batra, and Parikh]{goyal2017vqa}
Yash Goyal, Tejas Khot, Douglas Summers-Stay, Dhruv Batra, and Devi Parikh.
\newblock Making the V in VQA Matter: Elevating the Role of Image Understanding in Visual Question Answering.
\newblock In \emph{CVPR}, 2017.

\bibitem[Hasani et~al.(2026)Hasani et al.]{hasani2026countscope}
Hosein Hasani, Amirmohammad Izadi, Fatemeh Askari, Mobin Bagherian, Sadegh Mohammadian, Mohammad Izadi, and Mahdieh Soleymani Baghshah.
\newblock Understanding Counting Mechanisms in Large Language and Vision-Language Models.
\newblock In \emph{CVPR}, pages 5125--5133, 2026.

\bibitem[Hudson and Manning(2019)]{hudson2019gqa}
Drew A. Hudson and Christopher D. Manning.
\newblock GQA: A New Dataset for Real-World Visual Reasoning and Compositional Question Answering.
\newblock In \emph{CVPR}, 2019.

\bibitem[Hugging Face(2023)]{huggingface2023idefics}
Hugging Face Multimodal Team.
\newblock Introducing IDEFICS: An Open Reproduction of State-of-the-Art Visual Language Model.
\newblock Hugging Face technical blog and \texttt{idefics-9b-instruct} model documentation, 2023.

\bibitem[Krishna et~al.(2017)Krishna et al.]{krishna2017visualgenome}
Ranjay Krishna, Yuke Zhu, Oliver Groth, Justin Johnson, Kenji Hata, Joshua Kravitz, Stephanie Chen, Yannis Kalantidis, Li-Jia Li, David A. Shamma, Michael S. Bernstein, and Li Fei-Fei.
\newblock Visual Genome: Connecting Language and Vision Using Crowdsourced Dense Image Annotations.
\newblock \emph{International Journal of Computer Vision}, 123:32--73, 2017.

\bibitem[Le Dinh Anh et~al.(2026)Le Dinh Anh, Irawan, and Vo]{ledinhanh2026counting}
Duy Le Dinh Anh, Patrick Amadeus Irawan, and Tuan Van Vo.
\newblock Counting to Four is still a Chore for VLMs.
\newblock arXiv:2604.10039, 2026.

\bibitem[Lee et~al.(2025)Lee et al.]{lee2025cast}
Bruce W. Lee, Inkit Padhi, Karthikeyan Natesan Ramamurthy, Erik Miehling, Pierre Dognin, Manish Nagireddy, and Amit Dhurandhar.
\newblock Programming Refusal with Conditional Activation Steering.
\newblock In \emph{ICLR}, 2025.

\bibitem[Li et~al.(2023a)Li, Patel, Vi{\'e}gas, Pfister, and Wattenberg]{li2023iti}
Kenneth Li, Oam Patel, Fernanda Vi{\'e}gas, Hanspeter Pfister, and Martin Wattenberg.
\newblock Inference-Time Intervention: Eliciting Truthful Answers from a Language Model.
\newblock In \emph{NeurIPS}, 2023a.

\bibitem[Li et~al.(2023b)Li, Du, Zhou, Wang, Zhao, and Wen]{li2023pope}
Yifan Li, Yifan Du, Kun Zhou, Jinpeng Wang, Xin Zhao, and Ji-Rong Wen.
\newblock Evaluating Object Hallucination in Large Vision-Language Models.
\newblock In \emph{EMNLP}, pages 292--305, 2023b.

\bibitem[Makelov et~al.(2024)Makelov, Lange, Geiger, and Nanda]{makelov2024subspace}
Aleksandar Makelov, Georg Lange, Atticus Geiger, and Neel Nanda.
\newblock Is This the Subspace You Are Looking for? An Interpretability Illusion for Subspace Activation Patching.
\newblock In \emph{ICLR}, 2024.

\bibitem[Meng et~al.(2022)Meng, Bau, Andonian, and Belinkov]{meng2022rome}
Kevin Meng, David Bau, Alex Andonian, and Yonatan Belinkov.
\newblock Locating and Editing Factual Associations in GPT.
\newblock In \emph{NeurIPS}, 2022.

\bibitem[Meta(2024)]{meta2024llama32}
Meta.
\newblock Llama 3.2 Vision Model Card: Llama-3.2-11B-Vision-Instruct.
\newblock Hugging Face model documentation, 2024.

\bibitem[Moosavi-Dezfooli et~al.(2016)Moosavi-Dezfooli, Fawzi, and Frossard]{moosavi2016deepfool}
Seyed-Mohsen Moosavi-Dezfooli, Alhussein Fawzi, and Pascal Frossard.
\newblock DeepFool: A Simple and Accurate Method to Fool Deep Neural Networks.
\newblock In \emph{CVPR}, pages 2574--2582, 2016.

\bibitem[Nguyen et~al.(2026)Nguyen, Prasad, Stengel-Eskin, and Bansal]{nguyen2025grains}
Duy Nguyen, Archiki Prasad, Elias Stengel-Eskin, and Mohit Bansal.
\newblock GrAInS: Gradient-based Attribution for Inference-Time Steering of LLMs and VLMs.
\newblock In \emph{Proceedings of the 64th Annual Meeting of the Association for Computational Linguistics (Volume 1: Long Papers)}, pages 46523--46543, 2026.

\bibitem[Pearl(2009)]{pearl2009causality}
Judea Pearl.
\newblock \emph{Causality: Models, Reasoning and Inference}.
\newblock Cambridge University Press, second edition, 2009.

\bibitem[Qharabagh et~al.(2026)Qharabagh, Ghofrani, and Fountoulakis]{qharabagh2026lvlmcount}
Muhammad Fetrat Qharabagh, Mohammadreza Ghofrani, and Kimon Fountoulakis.
\newblock LVLM-Count: Enhancing the Counting Ability of Large Vision-Language Models.
\newblock \emph{Transactions on Machine Learning Research}, 2026.

\bibitem[Qian et~al.(2025)Qian et al.]{qian2025t2icount}
Yifei Qian, Zhongliang Guo, Bowen Deng, Chun Tong Lei, Shuai Zhao, Chun Pong Lau, Xiaopeng Hong, and Michael P. Pound.
\newblock T2ICount: Enhancing Cross-modal Understanding for Zero-Shot Counting.
\newblock In \emph{CVPR}, pages 25336--25345, 2025.

\bibitem[Rahmanzadehgervi et~al.(2025)Rahmanzadehgervi, Nguyen, Liu, Mai, and Nguyen]{rahmanzadehgervi2025tab}
Pooyan Rahmanzadehgervi, Hung Huy Nguyen, Rosanne Liu, Long Mai, and Anh Totti Nguyen.
\newblock TAB: Transformer Attention Bottlenecks Enable User Intervention and Debugging in Vision-Language Models.
\newblock In \emph{ICCV}, pages 22551--22562, 2025.

\bibitem[Rimsky et~al.(2024)Rimsky et al.]{rimsky2024caa}
Nina Rimsky, Nick Gabrieli, Julian Schulz, Meg Tong, Evan Hubinger, and Alexander Turner.
\newblock Steering Llama 2 via Contrastive Activation Addition.
\newblock In \emph{ACL}, pages 15504--15522, 2024.

\bibitem[Rong et~al.(2026)Rong et al.]{rong2026unicbench}
Chenggang Rong, Tao Han, Zhiyuan Zhao, Yaowu Fan, Jia Wan, Song Guo, Yuan Yuan, and Junyu Gao.
\newblock UNICBench: UNIfied Counting Benchmark for MLLM.
\newblock In \emph{CVPR}, pages 23730--23740, 2026.

\bibitem[Rudman et~al.(2026)Rudman et al.]{rudman2026prompt}
William Rudman, Michal Golovanevsky, Dana Arad, Yonatan Belinkov, Carsten Eickhoff, Ritambhara Singh, and Kyle Mahowald.
\newblock Mechanisms of Prompt-Induced Hallucination in Vision--Language Models.
\newblock In \emph{ACL}, pages 41894--41912, 2026.

\bibitem[Sengupta et~al.(2026)Sengupta, Moradinasab, Liu, and Brown]{sengupta2025count}
Saurav Sengupta, Nazanin Moradinasab, Jiebei Liu, and Donald E. Brown.
\newblock Can Vision-Language Models Count? A Synthetic Benchmark and Analysis of Attention-Based Interventions.
\newblock In \emph{Proceedings of the IEEE/CVF Conference on Computer Vision and Pattern Recognition (CVPR) Workshops}, 2026.

\bibitem[Sivakumar et~al.(2025)Sivakumar, Zhang, Abdul Hakim, and Thomas]{sivakumar2025steervlm}
Anushka Sivakumar, Andrew Zhang, Zaber Ibn Abdul Hakim, and Chris Thomas.
\newblock SteerVLM: Robust Model Control through Lightweight Activation Steering for Vision Language Models.
\newblock In \emph{Findings of EMNLP}, pages 23640--23665, 2025.

\bibitem[Stoehr et~al.(2024)Stoehr et al.]{stoehr2024scaling}Niklas Stoehr, Kevin Du, V{\'e}steinn Sn{\ae}bjarnarson, Robert West, Ryan Cotterell, and Aaron Schein.
\newblock Activation Scaling for Steering and Interpreting Language Models.
\newblock In \emph{Findings of EMNLP}, pages 8189--8200, 2024.

\bibitem[Tamarapalli et~al.(2025)Tamarapalli, Grover, Pande, and Yerramilli]{tamarapalli2025countqa}
Jayant Sravan Tamarapalli, Rynaa Grover, Nilay Pande, and Sahiti Yerramilli.
\newblock CountQA: How Well Do MLLMs Count in the Wild?
\newblock arXiv:2508.06585, 2025.

\bibitem[Testolin et~al.(2025)Testolin, Hou, and Zorzi]{testolin2025enumeration}
Alberto Testolin, Kuinan Hou, and Marco Zorzi.
\newblock Visual Enumeration Remains Challenging for Multimodal Generative AI.
\newblock \emph{PLOS ONE}, 20(9):e0331566, 2025.

\bibitem[Trott et~al.(2018)Trott, Xiong, and Socher]{trott2018counting}
Alexander Trott, Caiming Xiong, and Richard Socher.
\newblock Interpretable Counting for Visual Question Answering.
\newblock In \emph{ICLR}, 2018.

\bibitem[Turner et~al.(2023)Turner et al.]{turner2023acteng}
Alexander Matt Turner, Lisa Thiergart, Gavin Leech, David Udell, Juan J. Vazquez, Ulisse Mini, and Monte MacDiarmid.
\newblock Steering Language Models With Activation Engineering.
\newblock arXiv:2308.10248, 2023.

\bibitem[Vig et~al.(2020)Vig et al.]{vig2020causal}
Jesse Vig, Sebastian Gehrmann, Yonatan Belinkov, Sharon Qian, Daniel Nevo, Yaron Singer, and Stuart Shieber.
\newblock Investigating Gender Bias in Language Models Using Causal Mediation Analysis.
\newblock In \emph{NeurIPS}, 2020.

\bibitem[Wang et~al.(2025)Wang, Hu, and Jiang]{wang2025vseam}
Qidong Wang, Junjie Hu, and Ming Jiang.
\newblock V-SEAM: Visual Semantic Editing and Attention Modulating for Causal Interpretability of Vision-Language Models.
\newblock In \emph{EMNLP}, pages 17396--17420, 2025.

\bibitem[Wang et~al.(2026)Wang, Hu, and Jiang]{wang2026hones}
Qidong Wang, Junjie Hu, and Ming Jiang.
\newblock From Heads to Neurons: Causal Attribution and Steering in Multi-Task Vision--Language Models.
\newblock In \emph{Findings of ACL}, pages 36151--36175, 2026.

\bibitem[Wu et~al.(2024)Wu, Arora, Wang, Geiger, Jurafsky, Manning, and Potts]{wu2024reft}
Zhengxuan Wu, Aryaman Arora, Zheng Wang, Atticus Geiger, Dan Jurafsky, Christopher D. Manning, and Christopher Potts.
\newblock ReFT: Representation Finetuning for Language Models.
\newblock In \emph{NeurIPS}, 2024.

\bibitem[Wu et~al.(2025)Wu, Arora, Geiger, Wang, Huang, Jurafsky, Manning, and Potts]{wu2025axbench}
Zhengxuan Wu, Aryaman Arora, Atticus Geiger, Zheng Wang, Jing Huang, Dan Jurafsky, Christopher D. Manning, and Christopher Potts.
\newblock AxBench: Steering LLMs? Even Simple Baselines Outperform Sparse Autoencoders.
\newblock In \emph{ICML}, PMLR 267, pages 67035--67080, 2025.

\bibitem[Yang et~al.(2026)Yang, Liu, Yang, and Lee]{yang2026crvlm}
Jiaxi Yang, Shicheng Liu, Yuchen Yang, and Dongwon Lee.
\newblock Steering to Say No: Configurable Refusal via Activation Steering in Vision Language Models.
\newblock arXiv:2602.07013, 2026.

\bibitem[Yin et~al.(2026)Yin, Chen, Chen, Zhou, Wu, and He]{yin2026dynamic}
Jianghao Yin, Qin Chen, Kedi Chen, Jie Zhou, Xingjiao Wu, and Liang He.
\newblock Dynamic Multimodal Activation Steering for Hallucination Mitigation in Large Vision-Language Models.
\newblock In \emph{ICLR}, 2026.

\bibitem[Zhang et~al.(2018)Zhang, Hare, and Pr\"ugel-Bennett]{zhang2018counting}
Yan Zhang, Jonathon Hare, and Adam Pr\"ugel-Bennett.
\newblock Learning to Count Objects in Natural Images for Visual Question Answering.
\newblock In \emph{ICLR}, 2018.

\bibitem[Zhang et~al.(2026)Zhang et al.]{zhang2026vib}
Feiran Zhang, Yixin Wu, Zhenghua Wang, Xiaohua Wang, Changze Lv, Xuanjing Huang, and Xiaoqing Zheng.
\newblock VIB-Probe: Detecting and Mitigating Hallucinations in Vision-Language Models via Variational Information Bottleneck.
\newblock In \emph{ACL}, pages 23509--23521, 2026.

\bibitem[Zou et~al.(2023)Zou et al.]{zou2023repe}
Andy Zou, Long Phan, Sarah Chen, James Campbell, Phillip Guo, Richard Ren, Alexander Pan, Xuwang Yin, Mantas Mazeika, Ann-Kathrin Dombrowski, Shashwat Goel, Nathaniel Li, Michael J. Byun, Zifan Wang, Alex Mallen, Steven Basart, Sanmi Koyejo, Dawn Song, Matt Fredrikson, J. Zico Kolter, and Dan Hendrycks.
\newblock Representation Engineering: A Top-Down Approach to AI Transparency.
\newblock arXiv:2310.01405, 2023.

\end{thebibliography}
\end{document}